\documentclass[acmtog]{acmart}

\AtBeginDocument{%
  }

\setcopyright{cc}
\setcctype{by}
\copyrightyear{2026}
\acmYear{2026}
\acmJournal{TOG}
\acmVolume{45}
\acmNumber{4}
\acmMonth{7}
\acmArticle{}
\acmDOI{10.1145/3811334}
\usepackage{amsmath}
\usepackage[normalem]{ulem}
\usepackage{multirow}
\usepackage{colortbl}
\usepackage[dvipsnames]{xcolor}
\usepackage{makecell}
\usepackage{algorithm}
\usepackage{algorithmic}
\usepackage{float}

\begin{document}
\newcommand{\ourmethod}{MotionBricks}

\title{\ourmethod{}: Scalable Real-Time Motions with Modular Latent Generative Model and Smart Primitives}

\author{Tingwu Wang}
\setcounter{footnote}{1}
\authornote{Joint first authors. Corresponding author: Tingwu Wang, tingwuw@nvidia.com.}
\email{tingwuw@nvidia.com}
\affiliation{%
  \institution{NVIDIA}
  \country{USA}
}

\author{Olivier Dionne}
\authornotemark[2]
\email{olivier.dionne@gmail.com}
\affiliation{%
  \institution{NVIDIA}
  \country{Canada}
}

\author{Michael De Ruyter}
\email{michael.deruyter@gmail.com}
\affiliation{%
  \institution{NVIDIA}
  \country{USA}
}

\author{David Minor}
\email{dminor@nvidia.com}
\affiliation{%
  \institution{NVIDIA}
  \country{Canada}
}

\author{Davis Rempe}
\email{davrempe@gmail.com}
\affiliation{%
  \institution{NVIDIA}
  \country{USA}
}

\author{Kaifeng Zhao}
\email{kaifeng.zhao@inf.ethz.ch}
\affiliation{%
  \institution{NVIDIA}
  \country{Switzerland}
}
\affiliation{%
  \institution{ETH Z\"urich}
  \country{Switzerland}
}

\author{Mathis Petrovich}
\email{mpetrovich@nvidia.com}
\affiliation{%
  \institution{NVIDIA}
  \country{Switzerland}
}

\author{Ye Yuan}
\email{yey@nvidia.com}
\affiliation{%
  \institution{NVIDIA}
  \country{USA}
}

\author{Chenran Li}
\email{chenranl@nvidia.com}
\affiliation{%
  \institution{NVIDIA}
  \country{USA}
}

\author{Zhengyi Luo}
\email{zhengyil@nvidia.com}
\affiliation{%
  \institution{NVIDIA}
  \country{USA}
}

\author{Brian Robison}
\email{brobison@nvidia.com}
\affiliation{%
  \institution{NVIDIA}
  \country{USA}
}

\author{Xavier Blackwell}
\email{xblackwell@nvidia.com}
\affiliation{%
  \institution{NVIDIA}
  \country{USA}
}

\author{Bernardo Antoniazzi}
\email{bernie3d@gmail.com}
\affiliation{%
  \institution{NVIDIA}
  \country{USA}
}

\author{Xue Bin Peng}
\email{xbpeng@sfu.ca}
\affiliation{%
  \institution{NVIDIA}
  \country{Canada}
}
\affiliation{%
  \institution{Simon Fraser University}
  \country{Canada}
}

\author{Yuke Zhu}
\email{yukez@cs.utexas.edu}
\affiliation{%
  \institution{NVIDIA}
  \country{USA}
}
\affiliation{%
  \institution{The University of Texas at Austin}
  \country{USA}
}

\author{Simon Yuen}
\email{siyuen@nvidia.com}
\affiliation{%
  \institution{NVIDIA}
  \country{USA}
}
\received{22 January 2026}
\received[revised]{15 April 2026}
\received[accepted]{20 May 2026}

\authorsaddresses{}
\renewcommand{\shortauthors}{T. Wang, O. Dionne, M. De Ruyter, D. Minor, et al.}



\begin{teaserfigure}
  \includegraphics[width=\textwidth]{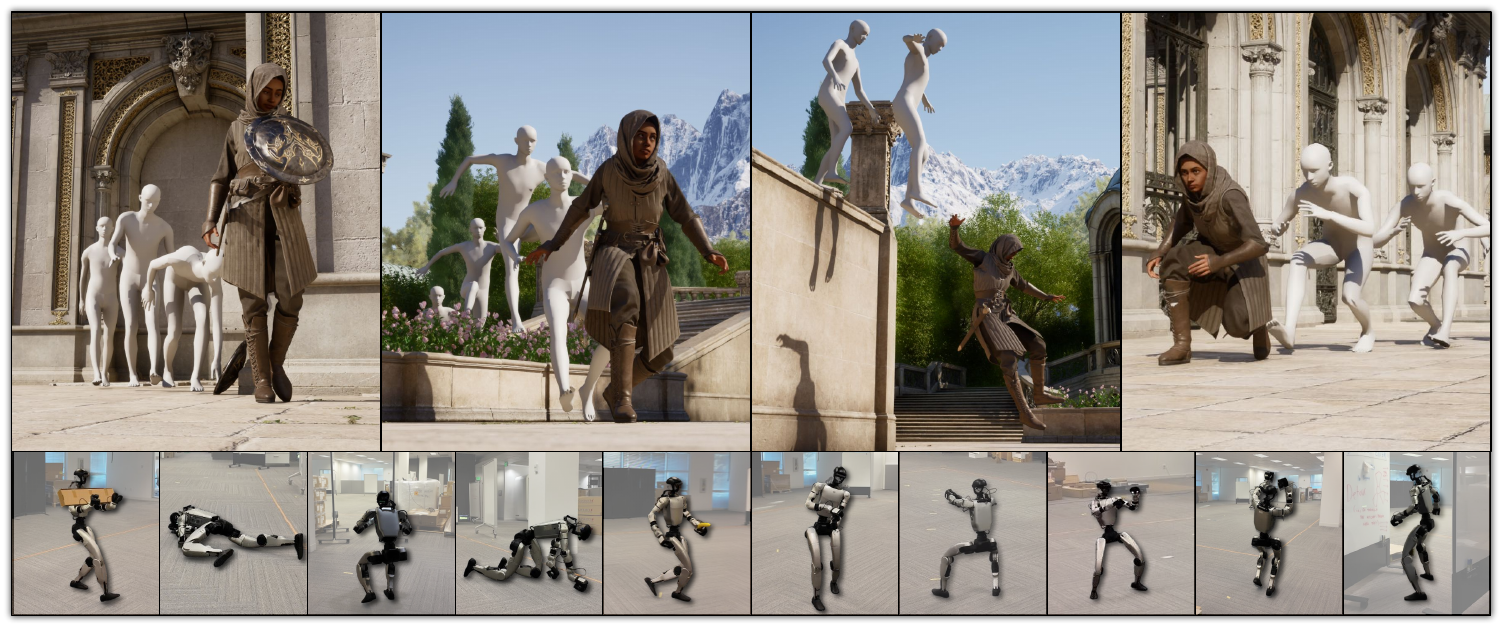}
  \caption{
  \ourmethod{} enables real-time motion control across animation and robotics.
  All motions are generated by our unified latent neural backbone using the smart primitive interface.
  Top: We showcase {\ourmethod{}} in a UE5 demo covering diverse locomotion styles, acrobatics, and object-scene interactions.
  Bottom: We deploy {\ourmethod{}} on the Unitree G1 robot for real-world whole-body control.
  The links to the dataset, code, and videos are available on the project webpage: \url{https://nvlabs.github.io/motionbricks}.
  }\label{fig:teaser}
\end{teaserfigure}

\begin{abstract}
Despite transformative advances in generative motion synthesis,
real-time interactive motion control remains dominated by traditional techniques.
In this work, we identify two key challenges in bridging research and production:
1) \emph{Real-time scalability}: 
Industry applications demand real-time generation of a vast repertoire of motion skills,
while generative methods exhibit significant degradation in quality and scalability under real-time computation constraints, and
2) \emph{Integration}:
Industry applications demand fine-grained multi-modal control involving velocity commands, style selection, and precise keyframes, a need largely unmet by existing text- or tag-driven models.
Moreover, a systematic motion design interface for generative models remains absent.
To overcome these limitations, we introduce {\ourmethod}: a large-scale, real-time generative framework with a two-fold solution.
First, we propose a large-scale modular latent generative backbone tailored for robust real-time motion generation, 
effectively modeling a dataset of over 350,000 motion clips with a single model.
Second, we introduce \textit{smart primitives} that provide a unified, robust, and intuitive interface for authoring both navigation and object interaction.
Notably, {\ourmethod} applies to new downstream tasks in a zero-shot manner, where no fine-tuning or task-specific tagging is required.
Applications can be designed in a plug-and-play manner like assembling bricks without expert animation knowledge, enabling an accessible interface for applications in animation and robotics.
Quantitatively, we show that {\ourmethod} produces state-of-the-art motion quality on open-source and proprietary datasets of various scales,
while also achieving a real-time throughput of 15,000 FPS with 2ms latency.
We demonstrate the flexibility and robustness of {\ourmethod} in a complete production-level animation demo, covering navigation and object-scene interaction across various styles with a unified model.
To showcase our framework's application beyond animation, we deploy {\ourmethod} on the Unitree G1 humanoid robot to demonstrate its flexibility and generalization for real-time robotic control.
\end{abstract}
\maketitle

\section{Introduction}
Runtime interactive motion control has been one of the foundational pillars supporting the expanding array of animation and robotics applications.
However, the typical process of designing runtime behavior systems is highly labor-intensive.
Animation or behavior graphs, and their variants, have been widely adopted in the animation and robotics industries~\cite{kovar2004automated,safonova2007construction,min2012motion,blendtrees,boston_dynamics_choreographer,unitree_boxing}, 
largely due to their ability to construct complex, hierarchical state machines with granular quality control.
In a conventional animation graph, pre-recorded motion clips are organized into discrete states,
with transitions between them governed by game events and user input. 
Consequently, animation graphs suffer from severe scalability limitations during the process of asset and motion graph design. 
For example, a modern AAA animation state machine, such as that used in \textit{Assassin's Creed},
may require managing over 15,000 animations, 5,000 states, and nested graphs up to 12 levels deep~\cite{holden2018character}, making such systems nearly impossible to maintain or extend.
This overwhelming complexity has effectively made high-quality runtime behavior systems the exclusive domain of large, well-resourced studios, highlighting an urgent need for new, more accessible approaches for the broader community.

Recent advances in generative models have led to remarkable breakthroughs across a wide range of content creation tasks, including motion synthesis~\cite{holden2017phase, peng2018deepmimic, tevet2022human}. 
Notably, two formulations have been extensively studied for interactive runtime motion control:
(1) Models that directly receive control commands as input.
This includes works spanning from non-parametric methods such as motion matching~\cite{buttner2015motion,clavet2016motion,buttner2019},
to deterministic models~\cite{holden2017phase, zhang2018mode,starke2019neural,starke2020local,starke2021neural,starke2023motion,lu2024choice},
to generative methods~\cite{gou2025control, shi2024interactive}.
Control commands include, for example, tags, root trajectories, velocities, and object attributes.
(2) Text-to-motion models~\cite{tevet2022human,zhang2024motiondiffuse,zhou2024emdm,alexanderson2023listen,cohan2024flexible,guo2024momask,pinyoanuntapong2024mmm} use text prompts as the main control modality.
The former requires manual tagging or attribute preparation during training,
and it is quite common to use a predefined ``one-hot'' vector to represent the control commands,
which ultimately limits scalability as the types of control commands vary significantly across different motion categories.
The latter, however, treats animation as a monolithic generation task from text prompts,
which has yet to provide the fine-grained control needed for reliable industrial-level applications and often struggles to balance scalability, quality, and speed.
Ultimately, runtime motion control is a complex task that bridges low-level motion synthesis and high-level behavior design, where a successful system needs to excel in both requirements.

To achieve fine-grained control and scalability in real-time,
we present \ourmethod{}, a framework designed for production use that seamlessly combines a powerful low-level latent neural backbone with a flexible, universal high-level behavior system built upon our proposed \emph{smart primitives}.
For the low-level neural backbone, we adopt motion in-betweening as the foundational paradigm.
We address real-time scalability challenges with a novel structured latent design that enables our backbone to significantly improve model capacity and spatial precision.
Our model achieves state-of-the-art motion quality while maintaining throughput far exceeding real-time requirements.
We further organize the model into a modular setup that supports progressive, coarse-to-fine motion generation, improving robustness and quality, while allowing for a transparent workflow for inspecting and refining intermediate results.
Throughout the pipeline, we support flexible combinations of constraint specifications,
enabling maximum adaptability for high-level smart primitive designs.
For the high-level behavior system, we introduce two smart primitives: \emph{smart locomotion} and \emph{smart object}.
Smart locomotion provides a robust mechanism for generating proxy keyframes across arbitrary navigation styles and velocity-heading commands, while safeguarding against command dead zones and producing natural movement details via neural root trajectory refinement.
Smart objects offer an intuitive interface for proxy object interactions with keyframes, allowing precise and natural control over interaction with the neural backbone.
This enables building highly complex scenes in a plug-and-play manner without the need for animation graphs.
Both primitives expose a user-friendly, scalable interface that unifies control into keyframe commands directed to the low-level backbone.
Notably, all applications and demos shown in this paper are built using a fixed pre-trained neural backbone, without requiring further fine-tuning or additional tagging.

The core contributions of \ourmethod{} are summarized as follows:
1) We propose a modular generative neural backbone, featuring a novel multi-head tokenizer and a progressive coarse-to-fine generation pipeline. Our design achieves state-of-the-art motion quality on various internal and open-source datasets, while maintaining 2ms latency and 15,000 FPS throughput, far exceeding real-time requirements;
2) We introduce smart primitives, a flexible high-level behavior system built on top of our neural backbone. We implement smart locomotion and smart object primitives in Unreal Engine, and demonstrate their effectiveness through a complete production-level demo, showcasing a wide variety of navigation and interaction skills with a unified interface, all without fine-tuning or additional tagging; and
3) We deploy \ourmethod{} on the Unitree G1 humanoid robot, demonstrating that our framework bridges the gap between virtual character animation and physical robot control, enabling a unified approach to motion synthesis across both domains.
The open-source project for \ourmethod{} can be found on our webpage.

\section{Related Work}
In this section, we provide an overview of relevant literature, covering both established traditional methods widely adopted in real-world applications and the latest advancements in data-driven neural network techniques.

\paragraph{Traditional Methods}
Motion graphs, or animation behavior graphs, are the prevalent approach for runtime animation and certain robotics applicability in the industry~\cite{lee2002interactive,arikan2002interactive,kovar2002motion,animgraph,unitree_boxing,boston_dynamics_choreographer}.
Animation graphs are state machines consisting of motion states such as walking, running, idle, etc.
Motion is generated by replaying clips within the state or blending between states during transitions~\cite{blendtrees} based on user controls or game events.
Academic research has further advanced the animation behavior graph, as seen in works such as~\citet{kovar2004automated,safonova2007construction,min2012motion}.
A particularly notable development is presented in~\citet{motionfield2010},
which enables motion transitions at a fine-grained, frame-by-frame level rather than the traditional node-by-node approach.
These academic innovations,~\citet{motionfield2010} in particular, ultimately inspired motion matching~\cite{buttner2015motion,clavet2016motion,buttner2019}.
In motion matching, future frames are retrieved from a motion database to match current state and user controls.
This approach eliminates the rigid, node-based constraints of traditional animation graphs,
enabling significantly more flexible and nuanced motion synthesis.
While efficiency improvements have been achieved through neural distillation and better search algorithms~\cite{holden2020learned,yi2019search}, motion matching is typically still used as a subcomponent for navigation within broader animation graphs and does not overcome the fundamental scalability limitations.
Ultimately, animation graphs become increasingly fragile and labor-intensive as the number of states and transitions grows, posing significant challenges for large-scale applications.

\paragraph{Pre-Generative Methods}
Prior to the advent of powerful generative models~\cite{ho2020denoising, achiam2023gpt},
researchers applied feedforward deterministic neural networks to motion synthesis~\cite{fragkiadaki2015recurrent,holden2016deep}.
Phase-Functioned Neural Networks (PFNN)~\cite{holden2017phase} is a seminal work in this direction,
which predicts navigation motion on different terrains autoregressively.
This approach inspired numerous models for runtime animation methods~\cite{zhang2018mode,starke2019neural,starke2020local,starke2021neural,starke2023motion,lu2024choice,starke2024categorical}.
However, deploying these models in real-world settings presents persistent challenges,
such as foot sliding, object penetration and excessive smoothing. 
These models are also often specialized, tailored for a small set of tasks.
As a result, their applicability in production remains restricted.

Another line of research formulates motion synthesis as an in-betweening or inpainting problem given the starting and ending keyframe constraints~\cite{harvey2020robust, oreshkin2023motion, qin2022motion}.
However, since these methods are typically tailored for a narrow set of skills, they inherit the same scalability challenges as the aforementioned autoregressive approaches.
Moreover, as target keyframes are usually unavailable during runtime animation,
these methods are predominantly utilized as offline tools for animation authoring.
\begin{figure*}[!tp]
    \centering
    \includegraphics[width=1.0\linewidth]{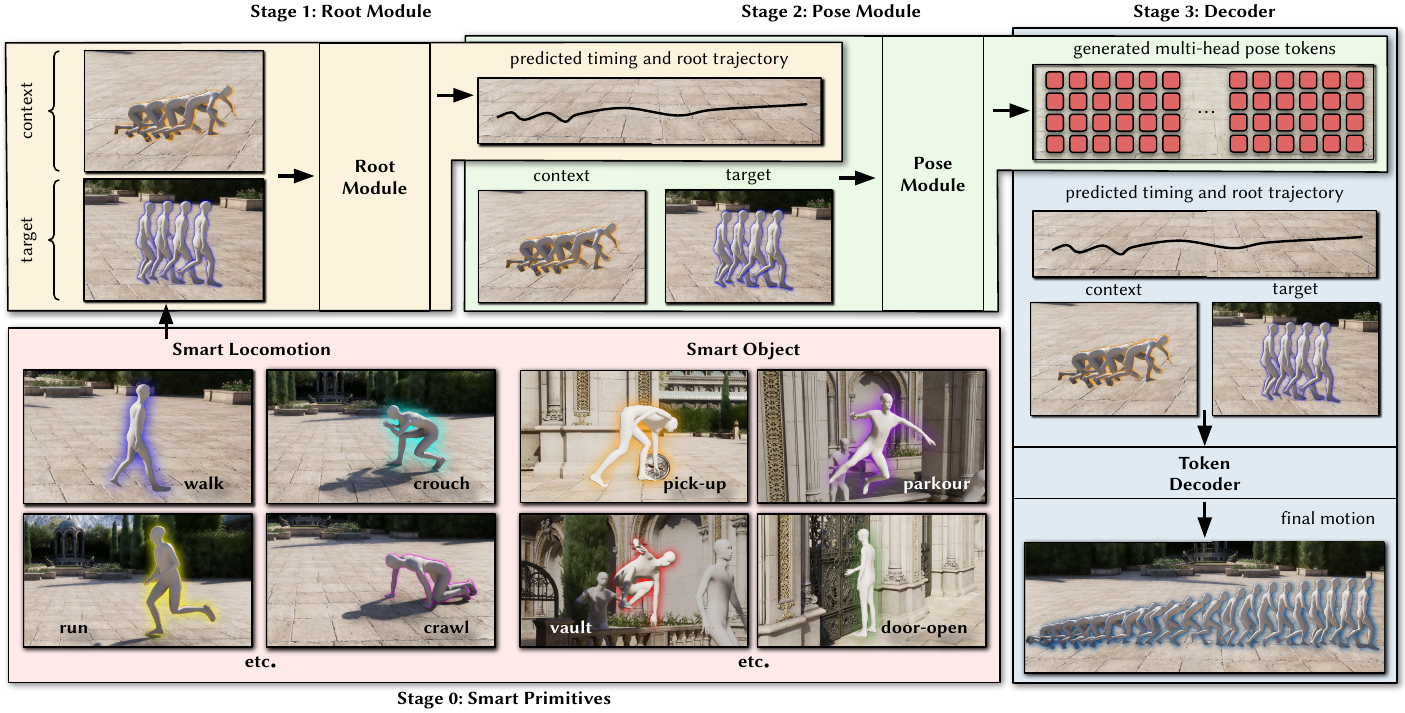}
    \vspace{-15pt}
    \caption{
    \ourmethod{}'s inference pipeline consists of four stages.
    Given user commands or game events, smart primitives generate target keyframes.
    The root module first predicts timing and root trajectory,
    followed by the pose module that models the distribution of multi-head latent pose tokens.
    Finally, the decoder produces continuous motion conditioned on pose tokens, root trajectories, and keyframes.
    }\label{fig:diagram}
\end{figure*}
\paragraph{Generative Methods}
Recently, diffusion models~\cite{ho2016generative, song2020denoising} have achieved remarkable success, particularly in image~\cite{rombach2022high, esser2024scaling} and video generation~\cite{openai_sora, google_veo3},
inspiring widespread adoption across various domains.
Of particular relevance are motion diffusion models (MDMs)~\cite{tevet2022human,zhang2024motiondiffuse,zhou2024emdm,alexanderson2023listen,cohan2024flexible,tevetclosd,li2025genmo}.
In the pioneering work~\cite{tevet2022human},
the authors train a motion diffusion model on the HumanML3D~\cite{guo2022generating} conditioned on text descriptions.
This work is further extended in~\citet{shafir2023human} to handle longer sequences and multiple interacting humans.
Runtime diffusion guidance~\cite{ho2022classifier} is incorporated in~\citet{shafir2023human},
enabling the enforcement of spatial constraints during inference without retraining.
In~\citet{li2023object,li2024controllable}, MDMs have also been extended to support object interactions. 
While MDMs offer strong scalability and motion quality, a significant challenge remains: inference speed. 
Standard diffusion models typically require several seconds, or even minutes, to generate motion clips, rendering them impractical for real-time applications. 
Although recent research explores auto-regressive diffusion models that amortize computation~\cite{shi2024interactive, han2024amd, chen2024taming, li2024aamdm} or use DDIM~\cite{song2020denoising} or techniques such as flow matching~\cite{lipman2022flow,gou2025control},
these approaches still face a significant trade-off between real-time performance and output quality,
and barely achieve real-time speed,
leaving them impractical for integration into already demanding animation or robotics pipelines.
Another promising line of research centers on token-based generative models,
which discretize continuous representations into discrete tokens~\cite{van2017neural,razavi2019generating,esser2021taming,tian2024visual},
and subsequently employ transformers for prediction.
This approach draws inspiration from the advances seen in discrete language, image, and video generative models~\cite{yu2023magvit,luo2024open,yu2023language,sun2024autoregressive}.
Works such as~\citet{guo2022tm2t,jiang2023motiongpt,zhang2024motiongpt} generate motion from text by treating motion tokens analogously to language tokens.
It is common to use an iteratively masked token prediction strategy,
in which tokens are predicted by progressively finalizing the most confident predictions~\cite{guo2024momask, pinyoanuntapong2024mmm,pinyoanuntapong2024controlmm}.
Concurrently, SONIC~\cite{luo2025sonic} adopts a similar token-based latent generation approach for humanoid robot control, though with a different tokenizer and network architecture, and is limited to locomotion without support for object interaction.
Overall, these methods face a similar trade-off between quality and speed as diffusion models, and have yet to achieve industry-grade interactivity, motion quality, or scalability.

\section{Overview}\label{sections:overview}
We present \ourmethod{}, an intuitive framework for fine-grained, scalable, and robust runtime motion control.
An overview of our framework is shown in Figure~\ref{fig:diagram}. The pipeline consists of four components: the smart primitive module, tokenizer, root module, and pose module.

We first train a conditional structured multi-head tokenizer that encodes $T$ frames of motion into $T/4$ discrete tokens using a temporal encoder and decoder model. 
Then we train a generative backbone comprising a root module and a pose module. The root module predicts timing and initial root trajectories from in-betweening constraints, while the pose module then models encoded pose token distributions conditioned on the in-betweening constraints as well as the root predictions.
Crucially, training is agnostic to downstream control modalities.
No predefined control types or task-specific tags are required, as all smart primitives communicate through a unified keyframe interface.
Once trained, the backbone requires no fine-tuning.
Users configure smart primitives: smart locomotion for navigation motions,
and smart object for scene object interactions.
These smart primitives can be intuitively set up from motion libraries and interactive authoring tools, essentially creating a fully connected motion graph where transitions are generated with the neural backbone.
At runtime, smart primitives generate target keyframes from user commands and game events.
These keyframes, along with the context frames, are fed into the generative backbone to progressively 
generate timing, root trajectories, and pose tokens, which are finally decoded into continuous motion.
The framework operates autoregressively, and replanning is triggered whenever the control signals are changed or the future motion buffer does not have enough frames left.
The runtime inference loop is summarized in Algorithm~\ref{alg:inference}. We use simplified notation for clarity; the formal definitions are provided in Sections~\ref{section:tokenizer} to \ref{section:smart_primitives}.

\begin{algorithm}[!tp]
\caption{\ourmethod{} Inference Loop}\label{alg:inference}
\begin{algorithmic}[1]
\REQUIRE Token decoder, root module $\mathcal{F}$, pose module $\mathcal{P}$, smart primitives $\mathcal{SP}$
\STATE Initialize motion buffer $\mathcal{B}$, character state $\mathcal{S}$
\WHILE{running}
\STATE Read game event or user commands $\mathcal{C}$
\IF{$\mathcal{C}$ changed \textbf{or} $|\mathcal{B}|$ running low}
\STATE {\% Stage 0}: construct keyframe constraints $\mathcal{T}$
\STATE $\mathcal{T} \leftarrow$ $\mathcal{SP}$($\mathcal{S}$, $\mathcal{C}$)
\vspace{4pt}
\STATE {\% Stage 1}: predict frame count $T$, root trajectory $\{r\}$
\STATE $T$, $\{r\} \leftarrow$ $\mathcal{F}$($\mathcal{T}$)
\vspace{4pt}
\STATE {\% Stage 2}: predict pose tokens $\{z_q\}$
\STATE $\{z_q\} \leftarrow$ $\mathcal{P}$($\{r\}$, $\mathcal{T}$, $T$)
\vspace{4pt}
\STATE {\% Stage 3}: decode continuous motion $\{r,\, p,\, q,\, v,\, c\}$
\STATE  $\{r,\, p,\, q,\, v,\, c\}$ $\leftarrow$ $\mbox{decoder}$\,($\{z_q\}$, $\{r\}$, $\mathcal{T}$, $T$)
\vspace{4pt}
\STATE update $\mathcal{B}$ with generated motion
\ENDIF
\STATE Pop next frame from $\mathcal{B}$; update $\mathcal{S}$
\ENDWHILE
\end{algorithmic}
\end{algorithm}

\paragraph{Flexible In-betweening Constraint Handling.}
In \ourmethod{}, we adopt an in-betweening formulation: context keyframes represent the character's recent states, while target keyframes are goals supplied by smart primitives.
As shown in Figure~\ref{fig:input_constraints}, $\mathcal{T}_1$, $\mathcal{T}_2$, and $\mathcal{T}_3$ denote the sets of provided constraints for local root, global root, and pose, respectively, covering both context and target keyframes.
We denote the keyframe constraints as $\mathcal{T} = \{\mathcal{T}_1, \mathcal{T}_2, \mathcal{T}_3\}$.
While context keyframes (indices $0, 1, 2, 3$) are always available, target constraints (indices $4, 5, 6, 7$) vary by task.
Different tasks require different levels of constraint density: navigation typically needs only 1 or 2 keyframes for style and root velocity to avoid over-constraining the motion, while object interaction requires denser specifications such as consecutive hand positions.
To handle this variability, each module in \ourmethod{} accepts an arbitrary subset of constraints. When a constraint is not provided, we substitute it with a learnable mask embedding, allowing the model to gracefully handle partial specifications.
This masking mechanism is applied consistently across the root module, pose module, and decoder.
Throughout all modules, we support in-betweening segments ranging from 12 to 64 frames at 30 FPS.

\section{Structured Multi-headed Tokenizer}\label{section:tokenizer}
In \ourmethod{}, we propose a structured conditional multi-head tokenizer that serves as the cornerstone of our generative neural backbone.
We map raw motion data into compact, abstract, structured latent representations and introduce a root-pose disentanglement strategy alongside a flexible conditioning mechanism tailored for diverse application needs.
Together, these designs greatly improve the scalability, flexibility, and precision of the generative model under runtime constraints.

\paragraph{State Representation.}
We consider motion segments of length $T$, where $T$ is randomly sampled from 12 to 64 frames in increments of 4.
For each frame at time $t$, we define the motion state as a tuple $(r_g, r_l, p, q, v, c)$, comprising:
(1) global root values $r_g$: projected global positions and the cosine and sine of the heading angle of the pelvis joint;
(2) local root values $r_l$: projected positional velocity and angular velocity of the pelvis joint in global coordinates;
(3) joint positions $p$ and rotations $q$ for all joints (excluding the projected root joint);
(4) joint velocities $v$ and contact labels $c$.
Note that local and global root values are interconvertible; we retain both to enable flexible constraint handling in the neural backbone.
To properly process motions such as crawling and flipping, we represent joint rotations in global coordinates to avoid canonicalization with ill-defined root headings.
Consequently, we do not apply heading canonicalization during training, but instead augment samples with random rotations to learn motion skills across all directions.
Similarly, joint positions $p$ are expressed as global coordinates relative to the root position, and velocities $v$ are computed in the global frame, both without heading canonicalization, to maintain a consistent motion representation.

\subsection{Encoder}
Drawing inspiration from foot-scaling and time-warping in traditional animation~\cite{bereznyak2024shadow}, we encode root and pose separately rather than jointly.
Although root and pose are strongly correlated, pose intent can be largely disentangled from root intent: for instance, a person may walk at slightly varying speeds while maintaining the same gait pattern.
Therefore, the encoder only encodes the local pose states without the root information as follows:
\begin{equation}
    \left\{z^{t}_e\right\}_{t=1}^{T/4} = \mbox{enc}\left(\left\{p^t, q^t\right\}_{t=1}^T\right).
\end{equation}
The latent embedding $z_e^t$ is generated by progressively down-sampling the features at rates of $2$ and $4$ using either a 1D convolutional network~\cite{kiranyaz20211d} or a transformer with positional encoding~\cite{vaswani2017attention} to ensure temporal consistency. 
For notational convenience, we overload $t$ to index both tokens (left-hand side, $T/4$ total because of the 4-frame downsampling) and frames (right-hand side, $T$ total); we adopt this convention throughout the paper for cleaner presentation.

Rather than tokenizing or discretizing the feature space with one giant codebook or manually partitioning features by body part as in prior work~\cite{aberman2020skeleton,pinyoanuntapong2024mmm}, we argue that such approaches are insufficient to capture the full diversity of human movement required for real-world applications.
Instead, we rely on the network to learn this latent decomposition in a fully data-driven manner.
The continuous latent embedding $z_e^t$ is quantized into the discrete latent embedding $z_q^t$ separately along the feature dimension using $K$ discrete codebooks $\{e_1, e_2, \ldots, e_K\}$:
\begin{equation}
    z_q^t = \begin{Bmatrix}
        z_{q,1}^t \\
        z_{q,2}^t \\
        \vdots \\
        z_{q,K}^{t}
    \end{Bmatrix}
    =
    \begin{Bmatrix}
        \arg\min_{e_1 \in \mathcal{E}_{1}} \|z_{e,1}^{t} - e_1\|_2^2 \\
        \arg\min_{e_2 \in \mathcal{E}_{2}} \|z_{e,2}^{t} - e_2\|_2^2 \\
        \vdots \\
        \arg\min_{e_K \in \mathcal{E}_{K}} \|z_{e,K}^{t} - e_K\|_2^2
    \end{Bmatrix}.
\end{equation}\label{equation:discretization}
This strategy encourages the encoder to learn the subtle compositionality and disentanglement of human movement, without imposing manual bias on how motion should be partitioned.
Additionally, multi-head tokenization yields a more robust latent manifold,
enabling graceful degradation when individual tokens are mispredicted, as demonstrated later in our experiments.

\begin{figure}[!t]
    \centering
    \includegraphics[width=\linewidth]{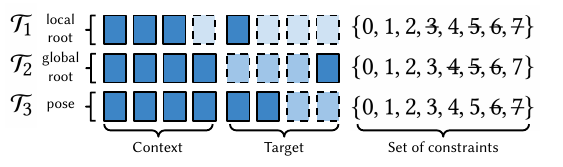}
    \caption{
        In {\ourmethod{}}, the root module, pose module, and decoder accept a flexible number of constraints from smart primitives.
        We denote the sets of constraints for local root, global root, and pose as $\mathcal{T}_1$, $\mathcal{T}_2$, and $\mathcal{T}_3$, respectively.
        Solid boxes indicate provided constraints; dashed boxes indicate optional or missing entries.
    }
    \label{fig:input_constraints}
\end{figure}

\subsection{Decoder}
Mirroring the encoder, the decoder progressively up-samples the discrete pose tokens $z_q^t$ at rates of $2$ and $4$.
Root trajectory conditions are incorporated at each up-sampling layer through skip connections at their corresponding temporal positions, encouraging coherent full-body motion generation.
To support flexible combinations of constraints from our smart primitives as shown in Figure~\ref{fig:input_constraints}, we allow keyframes produced by the primitives to be passed directly to the decoder.
During training, we randomly sample between 0 and 10 keyframes and incorporate them at each layer in the same manner as root conditions.
The decoder is formally defined as:
\begin{equation}\label{equation:token_decoder}
    \{r_l^t,\, p^t,\, q^t,\, v^t,\, c^t\}_{t=1}^{T} = \mbox{dec}\left(\{z_q^t\}_{t=0}^{T/4},\, \{\hat{r}_l^t\}_{t=0}^{T},\, \{\check{p}\},\, \{\check{q}\}\right),
\end{equation}
where the local root trajectory $\hat{r}_l^t$ is sampled from the dataset during training and predicted by the neural backbone at inference.
We denote $\{\check{p}\}$ and $\{\check{q}\}$ as variable-length sets of keyframe constraints on joint positions and rotations, respectively.
The local root trajectory $\hat{r}_l^t$ and keyframe constraints $\{\check{p}\}$, $\{\check{q}\}$ are provided to the decoder as input and additionally injected into hidden states at each up-sampling layer via skip connections.
Since the pose tokens $\{z_q^t\}_{t=0}^{T/4}$ have temporal dimension $T/4$ while the local root trajectory $\hat{r}_l^t$ has temporal dimension $T$, root features are temporally stacked by the downsampling factor $4$ before concatenation with the pose token embeddings along the feature dimension.
This stacking process is applied to the skip connection features of $\hat{r}_l^t$ at each up-sampling layer, where the temporal downsampling factor progressively decreases from $4$ to $2$ to $1$.
Sparse keyframe constraints $\{\check{p}\}$, $\{\check{q}\}$ are zero-padded to length $T$ and processed in the same manner as root trajectory features.
Additionally, a boolean availability mask determines whether keyframe embeddings from the skip connection or the decoder hidden states are selected at each layer.
We refer readers to our open-source repository for implementation details.
Notably, the decoder can also refine the predicted root trajectory, producing cleaner footsteps and smoother transitions.

\subsection{Quantizer and Training}
We train the tokenizer with a standard VQ-VAE quantizer loss, e.g., $L= \sum_{t=0}^{T/4}\sum_{k=1}^K\|sg(z_{q,k}^t) - e_k\|^2_2 + \beta\|z_{q,k}^t - sg(e_k)\|^2_2$, where $sg$ denotes the stop-gradient operator~\cite{van2017neural}.
In practice, we use a running-mean codebook update, which yields more stable training~\cite{razavi2019generating}.
Alternatively, an FSQ tokenizer~\cite{mentzer2023finite} can be used, which generates similar performance.
As discussed in Section~\ref{sections:overview}, during inference the decoder uses the first 4 frames as context keyframes, with 1 to 4 target keyframes provided by the smart primitives; together these form the keyframe constraints $\{\check{p}\}$, $\{\check{q}\}$ in Equation~\ref{equation:token_decoder}.
However, during training, a random number (between $0$ and $10$) of keyframe constraints are sampled at arbitrary temporal positions to encourage the model to learn generalizable and robust features.
The same variable-constraint sampling strategy is applied to the root and pose module training as well.

\section{Neural Backbone}\label{section:backbone}
The neural backbone of \ourmethod{} is a progressive modular architecture that generates in-betweening motions driven by smart primitives in a coarse-to-fine manner.
It comprises two components: the root module and the pose module.
The root module predicts the number of frames between keyframes and produces an initial estimate of the root trajectory.
The pose module then models the distribution of pose tokens conditioned on the root trajectory and keyframes.
This modular design not only iteratively improves motion quality, but also provides a transparent interface for inspecting, modifying, and refining intermediate results.

\subsection{Root Module}
The root module follows a two-step design, taking the constraints shown in Figure~\ref{fig:input_constraints} as input.
The first step predicts the number of in-between frames; the second step estimates the initial root trajectory conditioned on the timing prediction and hidden states from the first step.

\paragraph{Step 1: Hidden State Initialization and Timing Prediction.}
We use a transformer encoder for step 1, whose input consists of three types of embeddings:
(1) a sequence of $16$ learnable frame-slot embeddings $\{h_1^t\}_{t=1}^{16}$, each serving as a query position for predicting the root trajectory at that temporal location, covering a maximum of $64$ frames after $4\times$ downsampling;
(2) a global embedding encoding the keyframe constraints, denoted as $f(\mathcal{T}_1, \mathcal{T}_2, \mathcal{T}_3)$; and
(3) a timing embedding $g(T_1)$, encoding the ground-truth frame count information $T_1$ when available, or a learnable mask embedding otherwise.
Both $f$ and $g$ are learnable 3-layer MLPs that project inputs to the same dimension as the frame-slot embeddings.
The transformer outputs a distribution over in-between frame counts at 4-frame resolution, from which we sample or take the argmax during inference, yielding the predicted frame count $T_2$.
When the ground-truth frame count is provided, the predicted distribution is masked out to only allow ground-truth frame count.
We also support bit-wise masking over the distribution of valid frame counts, enabling flexible timing control at inference.

\paragraph{Step 2: Conditional Root Trajectory Estimation.}
Intermediate frame-slot embeddings from step 1 are passed to step 2, denoted as $\{h_2^t\}_{t=1}^{16}$.
We mask out frame-slot embeddings beyond the predicted frame count $T_2$.
A temporal transformer then processes this sequence to predict the initial root trajectory conditioned on keyframe and timing constraints.

Finally, the root module is formally defined as:
\begin{equation}
\begin{aligned}\label{equation:root_first_stage}
\{h_2\},\, T_2 &= \mathcal{F}_1\left(\{h_1\};\,g(T_1);\, f(\mathcal{T}_1,\, \mathcal{T}_2,\, \mathcal{T}_3)\right), \\
\{r_g\} &= \mathcal{F}_2\left( \{h_2\};\, g(T_2);\, f(\mathcal{T}_1,\, \mathcal{T}_2,\, \mathcal{T}_3)\right). \\
\end{aligned}
\end{equation}
In Equation~\ref{equation:root_first_stage}, $\mathcal{F}_1$ and $\mathcal{F}_2$ denote the transformer encoders for step 1 and step 2, respectively.
In practice, we also augment the frame-slot embeddings $h_1$ and $h_2$ with positional encoding~\cite{vaswani2017attention} for temporal information.
The root module outputs global root values $\{r_g\}_{t=1}^{T_2}$, enabling direct control over root trajectories to satisfy global constraints.
Due to the $4$ times downsampling, each final embedding is decoded into 4 consecutive frames of root trajectory $r_g$.
In subsequent pose and decoder modules, we convert these global root values to local root values for finer-grained motion generation.

\subsection{Pose Module}
The pose module is the largest transformer in our framework, modeling the distribution of pose tokens conditioned on root trajectories and keyframe constraints.
Similar to the root module, we construct frame-slot embeddings, where each encodes:
(1) local and global root values $r_l$, $r_g$, sampled from the dataset during training and predicted by the root module at inference;
(2) local pose keyframe constraints $p$, $q$; and
(3) pose token embeddings, using a cosine scheduled mix of ground truth and masked-token embeddings during training, and all initialized to masked-token embeddings at inference.
For local pose keyframe constraints, similar to the decoder (Section~\ref{section:tokenizer}), we randomly sample 0 to 10 keyframes during training and extract them from $\mathcal{T}_3$ at inference. 
Missing entries are replaced with learnable masked-pose embeddings.
We adopt the masked token modeling strategy~\citep{yu2023magvit,yu2023language,luo2024open,pinyoanuntapong2024mmm} which iteratively predicts the pose tokens based on confidence.
In practice, a single forward pass at inference typically produces high-quality motion. However, we still use cosine scheduling as an effective curriculum learning approach during training.

The pose module is formally defined as follows:
\begin{equation}
\begin{aligned}
    \{h_{p}\} &= \mathcal{P}\left(\{\phi\left(\{r_l\};\, \{r_g\};\, \{z_q\} \right)\}\right),\\
    p(z_{q,k}^t) &= \sigma\left(f_k\left(h_{p,k}^t\right)\right), \,k = 1, \ldots, K
\end{aligned}
\end{equation}
where $z_q$ is the latent token embedding vector in Equation~\ref{equation:discretization} and $\phi$ is the input embedding function that applies separate linear projections to the three input types and combines them via an MLP.
$\mathcal{P}$ denotes the transformer encoder, and $h_{p,k}^t$ is the $k$-th partition of the final hidden state $h_p^t$ along the feature dimension.
Each partition is projected through a linear layer $f_k$ followed by softmax $\sigma$ to produce the token distribution for the $k$-th codebook head.
The finalized pose tokens are then decoded along with the predicted root from the root module, as shown in Equation~\eqref{equation:token_decoder}.

\begin{figure}[!t]
    \centering
    \includegraphics[width=0.48\textwidth]{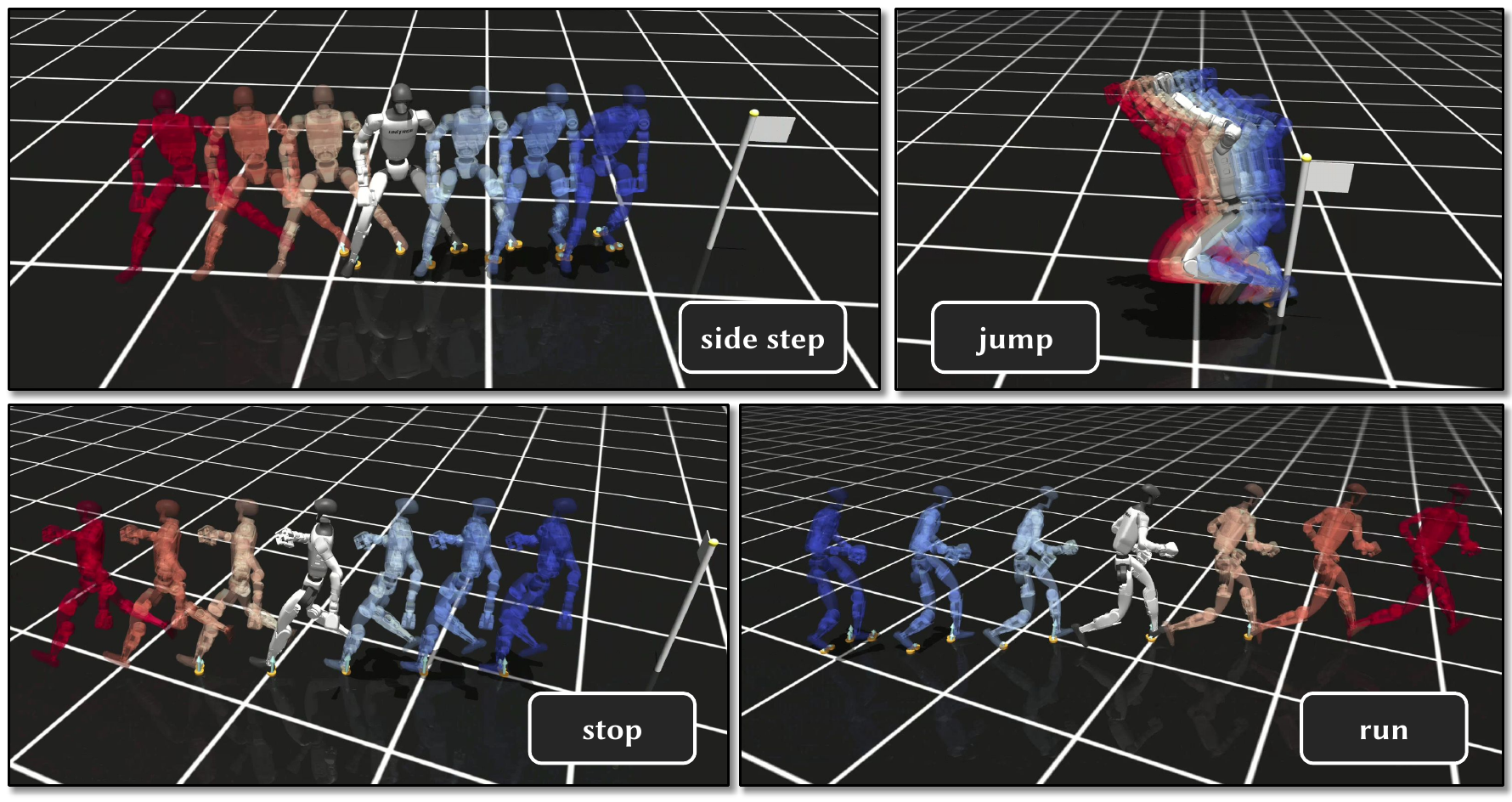}
    \caption{
    Our root-disentangled decoder automatically warps motion to different root trajectories with the same pose tokens,
    enabling both robust and precise control.
    Root trajectories are interpolated linearly from red to blue characters.
    The neutral-colored character shows the original motion.}
    \label{fig:root_interpolation_g1}
    \centering
    \includegraphics[width=1.01\linewidth]{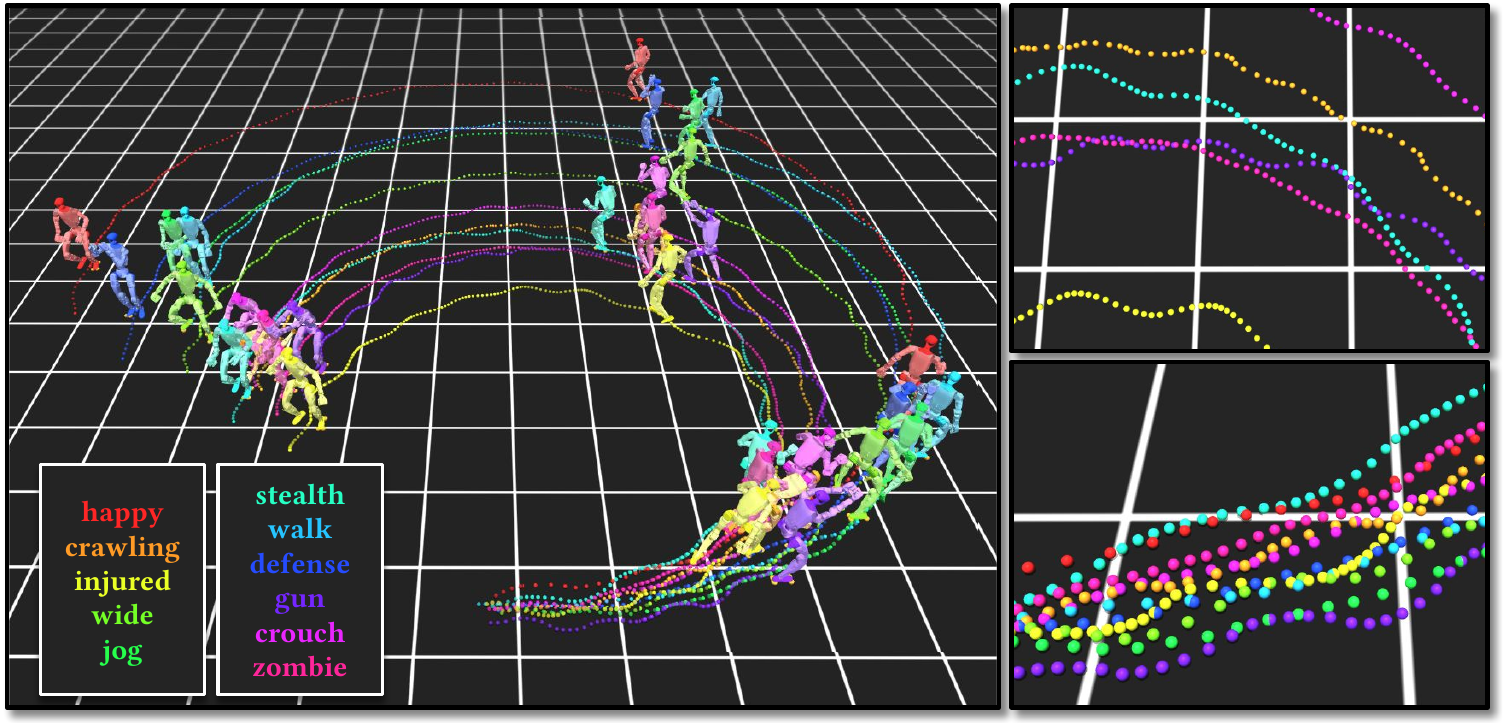}
    \caption{
    Neural root refinement automatically adjusts trajectories based on locomotion style.
    Despite identical initial root states and navigation commands, each style (e.g., happy, crawling, stealth, crouch) produces a distinct root trajectory.
    Right column: zoomed views showing fine-grained trajectory details.
}
    \label{fig:smart_locomotion_root_trajectory}
\end{figure}

\begin{figure*}[!t]
    \centering
    \includegraphics[width=1.0\linewidth]{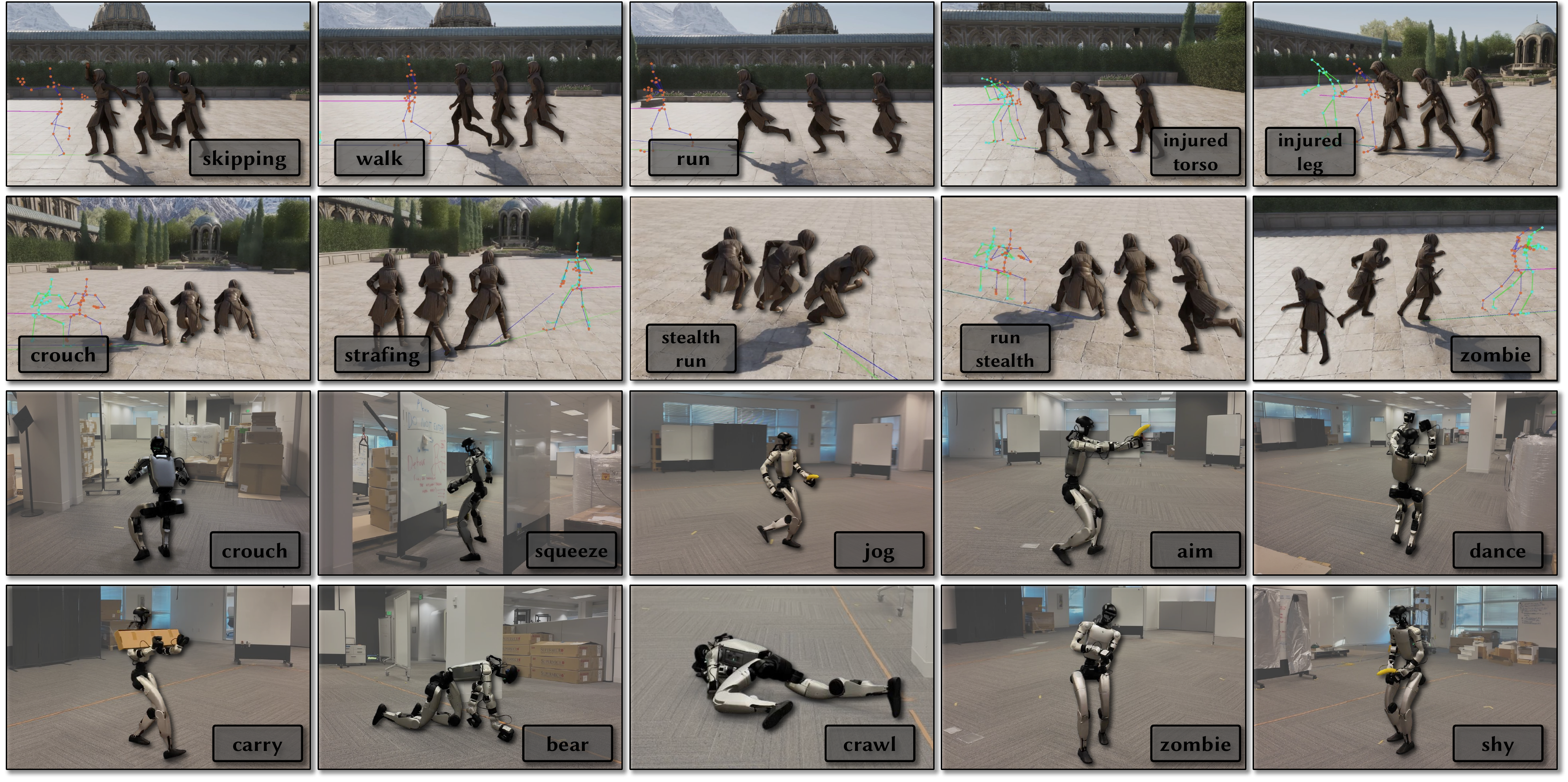}
    \caption{
    Diverse locomotion styles generated by smart locomotion in UE5 (top two rows) and on the Unitree G1 robot (bottom two rows).
    Styles include walking, running, crouching, strafing, stealth, crawling, and expressive movements, all generated from a single neural backbone.
}\label{fig:smart_locomotion_styles}
\end{figure*}
\section{Smart Primitives}\label{section:smart_primitives}
In this section, we introduce two smart primitives: (1) \emph{smart locomotion} for navigation with arbitrary styles and velocity/heading commands, and (2) \emph{smart object} for interacting with static scenes and dynamic objects.
Together, these primitives form a unified and flexible framework powered by a shared neural backbone, providing a complete solution for building rich, interactive scenes.

A key advantage of {\ourmethod} is zero-shot generalization to diverse downstream tasks, enabled by our scalable generative neural backbone.
Unlike prior methods that require task-specific conditioning during training, such as velocity commands, one-hot task labels, or learned task descriptors, our model requires no such supervision.
All task specifications are instead defined at runtime, enabling broad applicability without retraining.

\subsection{Smart Locomotion}\label{section:smart_locomotion}
Controllable locomotion presents two key challenges:
(1) generating natural motions from a wide range of velocity and heading commands,
which is nontrivial when the ``correct'' velocity for a given navigation context is unknown or ambiguous, or when users specify infeasible targets (control dead zone); and
(2) supporting diverse locomotion styles with smooth transitions.
Smart locomotion addresses these with two mechanisms:
\emph{progressive root trajectory refinement}, which reliably and intelligently balances user input with neural correction, and \emph{styled keyframe placement} with repetitive planning for natural style transitions.
We introduce each component in temporal order of execution.

\paragraph{Critically Damped Spring Model for Initial Root Trajectory Estimation.}
As defined earlier, root states $r_g$ comprise the projected global position and heading angle of the pelvis joint.
Inspired by the damped spring model commonly used in motion matching~\citep{clavet2016motion,holden2020learned}, we generate an initial guess of the root trajectory.
We first compute a naive target $r_{g,1}$ by linearly extrapolating the current state using the user's commanded velocity and heading (from gamepad, keyboard, etc.) over a 1.0s horizon.
We then apply a critically damped spring to smooth the trajectory $r(t)$ from the current root $r_0$ toward the target $r_{g,1}$:
\begin{equation}\label{equation:spring_model}
    r(t) = e^{-\gamma t}\left((r_0 - r_{g,1}) + (v_0 + \gamma(r_0 - r_{g,1}))t\right) + r_{g,1},
\end{equation}
where $\gamma$ is the damping coefficient, and $r_0$ and $v_0$ are the current position and velocity. Setting $t=1.0$~s yields the spring-smoothed target $r_{g,2}$.

\begin{figure*}[!t]
    \centering
    \includegraphics[width=1.0\linewidth]{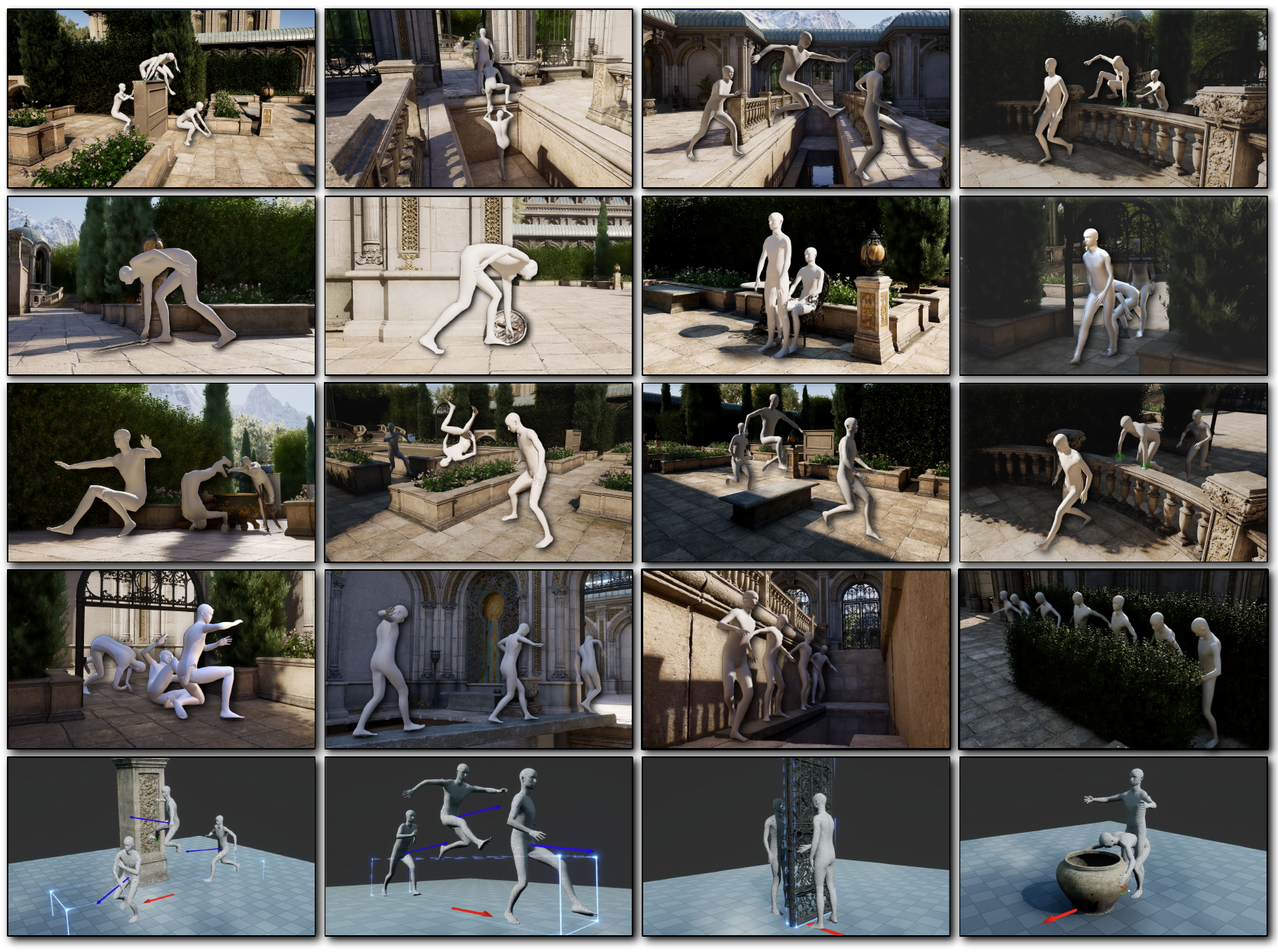}
    \caption{
    Smart objects enable diverse scene and object interactions from a flexible number of keyframes.
    Top rows: Examples including ledge climbing, vaulting, sitting, falling from heights, and object pickup.
    Bottom row: Keyframe authoring interface using standard translation and rotation gizmos in UE5.}
    \label{fig:smart_object_examples}
\end{figure*}

\paragraph{Style Control via Keyframe Selection.}
To control locomotion style, we leverage our in-betweening formulation by placing keyframes, sampled from short reference clips or human-authored poses of the desired style, onto the spring-smoothed trajectory $r_{g,2}$.

Traditional locomotion controllers must carefully align keyframes with footstep phases and explicitly handle style transitions to avoid artifacts.
In contrast, our method automatically generates high-quality, natural motions with smooth style transitions.
We attribute this to two factors:
(1) a powerful generative backbone that has learned diverse motion skills and can adapt keyframes to context, and
(2) a replanning mechanism that re-plans after a fixed interval, preventing the output from being rigidly locked to arbitrarily placed keyframes.
We demonstrate in both our UE5 demo and robotics deployment that even with a single keyframe and artificially inserted velocity/heading values, {\ourmethod} produces natural, coherent motions.
We also support keyframe generation via 2D blendspaces, showcasing the flexibility of our approach for integration with existing industrial animation tools.
We discuss the differences between traditional 2D blendspaces and the keyframe blending approach in \ourmethod{} in Appendix~\ref{section:appendix_2d_blendspace}.

\begin{table}[!h]
\centering
\caption{Progressive root trajectory refinement in smart locomotion.}
\label{tab:root_stages}
\begin{tabular}{l|c|l}
\toprule
stage & symbol & description \\
\midrule
naive & $r_{g,1}$ & direct extrapolation\\
spring model & $r_{g,2}$ & critically damped spring smoothing \\
root module & $r_{g,3}$ & refinement with style and timing \\
decoder & $r_{g,4}$ & refinement with full-body motion \\
\bottomrule
\end{tabular}
\end{table}

\paragraph{Neural Root Refinement.}
While the spring-smoothed root $r_{g,2}$ provides a reasonable initial estimate, it does not account for constraints imposed by context and target keyframes.
We therefore feed $r_{g,2}$ along with keyframe information into our root module to produce the refined root trajectory $r_{g,3}$.
The root module generates trajectories with natural, realistic movement details (as opposed to the mechanistic spring-smoothed output), and leverages timing prediction to avoid artifacts such as invalid or abrupt velocities, effectively safeguarding the output from control dead zones.
As shown in Figure~\ref{fig:smart_locomotion_root_trajectory}, even with identical initial root states and navigation commands from keyboard, the final root trajectories are automatically adjusted to produce natural motions that respect style constraints.
Finally, $r_{g,3}$ is passed to the decoder, which produces the final motion while further refining root values in coordination with full-body details such as footsteps and velocities, yielding the output $r_{g,4}$.
Crucially, the decoder's disentangled handling of root and pose enables robust adaptation to varied inputs.
As shown in Figure~\ref{fig:root_interpolation_g1}, the final motions adapt gracefully to interpolated root trajectories, enabling flexible post-processing and root constraint adjustments in downstream applications.

As summarized in Table~\ref{tab:root_stages}, the root trajectory undergoes progressive refinement from user input to final motion, striking the balance between responsiveness of user input and neural corrections.
In Figure~\ref{fig:smart_locomotion_styles}, we showcase diverse locomotion styles generated by smart locomotion in both UE5 and on the Unitree G1 robot.

\subsection{Smart Object}\label{section:smart_object}
Object interaction in games and simulations traditionally relies on template-based solutions: pre-canned animation clips or pre-defined robotic skill routines tied to specific object positions and orientations.
This approach suffers from two major limitations:
(1) significant authoring burden, as different objects and configurations require dedicated animation clips or task-specific processing, and
(2) rigid, inflexible playback that struggles with varied object placements or smooth transitions into and out of interactions.

Our smart object primitive addresses these limitations by leveraging the in-betweening capability of our neural backbone.
Rather than relying on pre-canned animation clips, we define each interactive object via two lightweight components: \emph{intent keyframes} that guide the character toward key poses (serving as either soft guidance or hard constraints depending on the interaction phase), and an \emph{interaction binding} that anchors the interaction logic to the object.

\begin{figure}[!t]
    \centering
    \includegraphics[width=1\linewidth]{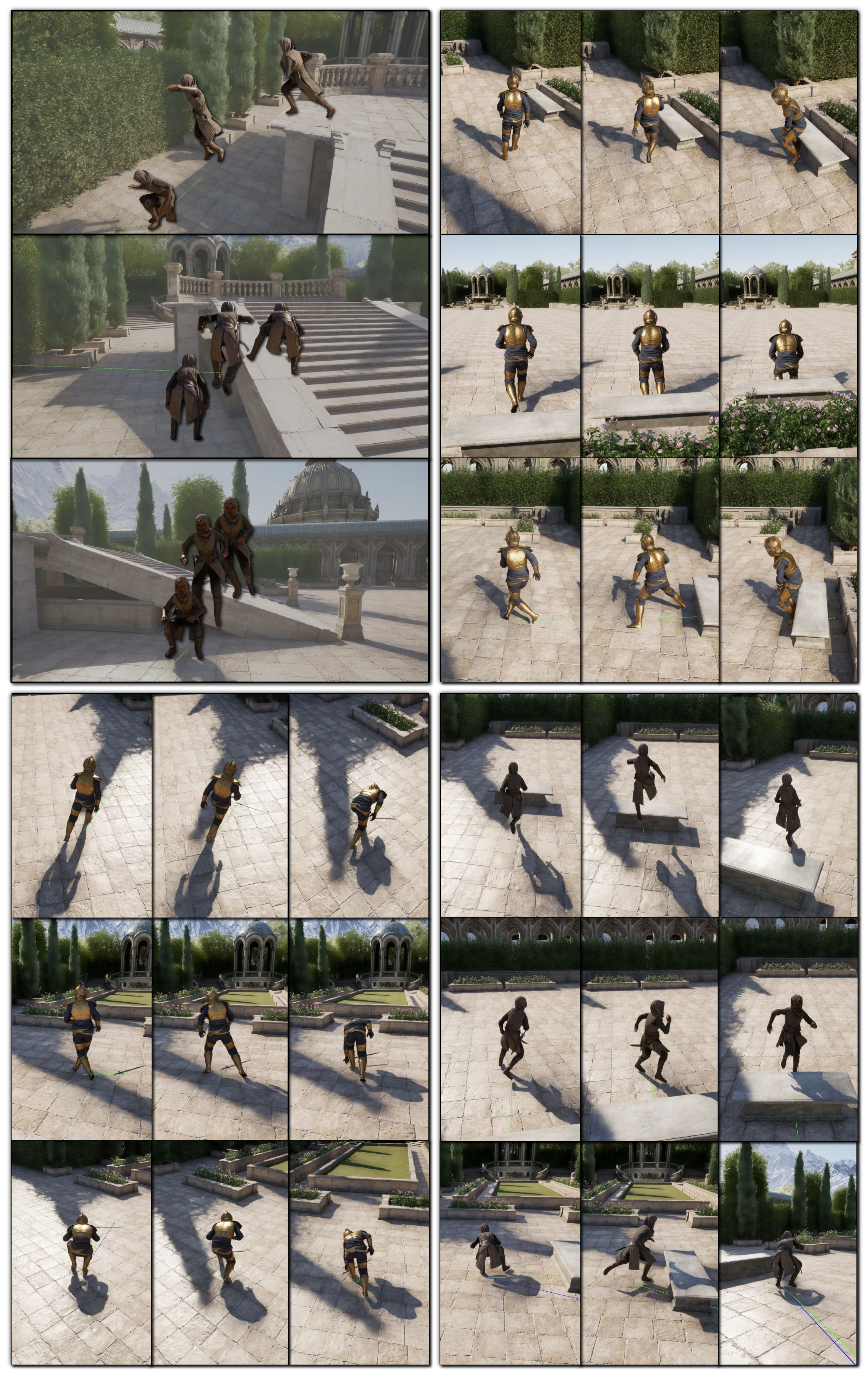}
    \caption{
    Automatic motion variations from smart objects using the same keyframe setup.
    Clockwise from top-left: falling, sitting, vaulting, and sword pickup.
    Each skill shows three variants, with keyframes displayed chronologically from left to right.
}\label{fig:smart_object_auto_styles}
\end{figure}
\paragraph{Smart Object Intent Keyframes.}
Keyframes define the essential poses for an interaction behavior.
For example, a climbing interaction may consist of multiple keyframe sets:
(1) a contact pose where the character's hands reach the wall edge, and
(2) an exit pose where the character stands on the ledge after climbing.
As shown in Figure~\ref{fig:smart_object_examples}, we support a variable number of keyframe sets per behavior, with each set containing one or more keyframes.
Users can obtain keyframes from a motion library or author them manually using standard translation, rotation, and IK gizmo tools.

To control how strictly keyframes are enforced, each keyframe set includes a drop-frame attribute $\tau$.
Setting $\tau=0$ treats the keyframe as a \emph{hard constraint}: the generated motion must fully reach the target pose before proceeding, which is essential for precise contact interactions like hand placement during climbing.
Setting $\tau>0$ treats the keyframe as \emph{soft guidance}: the model uses the keyframe to shape the motion's intent, but may transition to the next phase up to $\tau$ frames early without reaching the exact pose.
This is useful for preparation keyframes (e.g., poses preparing the hands and feet for a vaulting motion) or exit poses, where strict enforcement would produce unnatural, over-constrained motion.
We also note that a single smart object can define multiple keyframe sets for skill variety, and we support neurally generated keyframes on the fly.
Additionally, each keyframe includes a boolean flag $\omega$ indicating whether it should be rotated around the interaction pivot at runtime, as discussed below.
\begin{table}[!t]
    \centering
    \caption{Overview of datasets used in our experiments.}
    \label{tab:dataset_comparison}
    \scalebox{0.9}{%
    \begin{tabular}{l|c|c|c|c}
    \toprule
    Dataset & Hours & Train clips & Test clips & Joints \\
    \midrule
    350k & 700 & 315,162 & 35,018 & 27 \\
    70k & 140 & 62,132 & 35,018 & 27 \\
    HumanML3D & 28.6 & 23,206 & 2,578 & 22 \\
    LaFAN1-G1 & 4.6 & 2,362 & 262 & 34 \\
    \bottomrule
    \end{tabular}
    }
\end{table}

\begin{figure}[!t]
    \centering
    \includegraphics[width=0.48\textwidth, trim={20pt 0pt 0 0pt}]{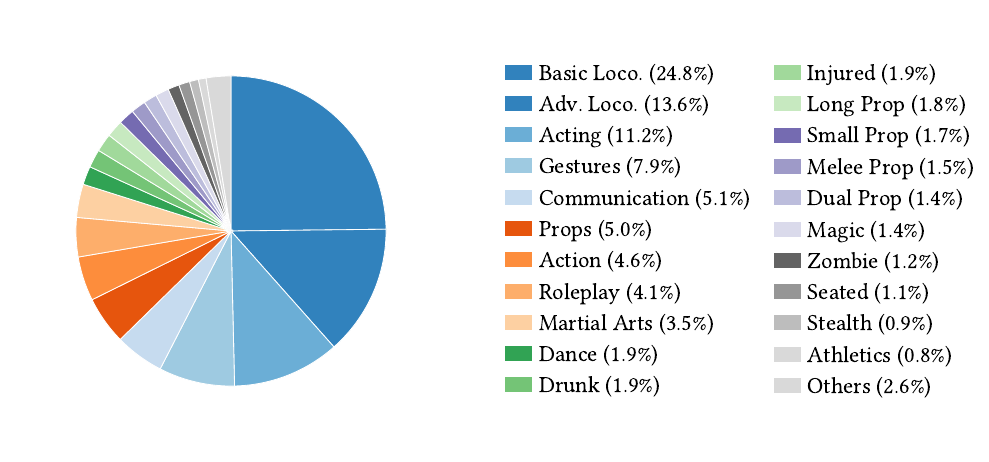}
    \vspace{-30pt}
    \caption{
    Distribution of the 350k dataset across its 36 categories. ``Basic.Loco.'', ``Adv.Loco.'', and ``Ext.Loco.'' denote basic, advanced, and extreme locomotion, respectively. ``Human Inter.'' denotes human interaction.
    Categories comprising fewer than 0.5\% of total clips are grouped under ``Other''.
    }\label{fig:dataset_distribution_by_categories}
\end{figure}

\paragraph{Interaction Binding.}
Each smart object defines an \emph{interaction binding} that manages three aspects of interaction: detection, placement, and keyframe anchoring, as shown below.
(1) Detection:
We derive an interaction mesh from the object's physical geometry, which serves as a trigger volume.
At runtime, box collision tracing determines whether the character is within interaction range and can initiate the interaction.
(2) Object sockets and placement:
For portable objects, the binding defines sockets specifying where the object attaches to the character (e.g., hands) and placement logic determining where the object should be positioned once released, for instance, snapping to a valid surface detected via ray cast.
(3) Keyframe anchoring:
For scene and static object interactions, the mesh's world transform acts as a pivot to anchor and rotate keyframes (for those keyframes with $\omega=1$) relative to the object.
This enables a character to approach the same ledge from different angles, or interact with objects at varying positions, using identical keyframe definitions, a key advantage of our generative approach over template-based playback.
Our implementation leverages UE5's built-in modules (box traces, collision detection, socket systems), demonstrating that our method integrates seamlessly with existing game engine infrastructure.
As shown in Figure~\ref{fig:smart_object_examples}, diverse scene and object interactions can be authored in a scalable way.
Furthermore, Figure~\ref{fig:smart_object_auto_styles} demonstrates automatic motion variations.
Falling motions adapt to different heights, and scene object interactions such as pickup, vaulting, and sitting, adjust to varied angles, velocities, and styles without additional authoring.

\section{Experiments}


\begin{table*}[!t]
\centering
\caption{Quantitative comparison on the \textbf{350k} dataset. 
$\downarrow$ indicates lower is better, $\uparrow$ indicates higher is better.
Best results are in \textbf{bold}, 
while \underline{underlined} indicates the second-best or competitive runner-up results.}
\label{tab:350k_comparison}
\resizebox{\textwidth}{!}{%
\begin{tabular}{l|cc|cc|cc|cc|cc|ccc|ccccc}
\toprule
\multirow{3}{*}{Method} & \multicolumn{2}{c|}{Speed} & \multicolumn{2}{c|}{Distribution} & \multicolumn{2}{c|}{Human Eval} & \multicolumn{2}{c|}{{Diversity}} & \multicolumn{2}{c|}{Smoothness} & \multicolumn{3}{c|}{Physical Plausibility} & \multicolumn{5}{c}{Precision} \\
& FPS $\uparrow$ & Latency $\downarrow$ & MMD $\downarrow$ & FID $\downarrow$ & Win $\uparrow$ & Score $\uparrow$ & {Jnt $\uparrow$} & {Latent $\uparrow$} & \makecell{Jnt \\Jit. $\downarrow$} & \makecell{Root \\ Jit. $\downarrow$} & \makecell{Foot \\Sk. $\downarrow$} & Pene. $\downarrow$ & \makecell{Foot \\Acc. $\uparrow$} & \makecell{Ctx \\KF  $\downarrow$} & \makecell{Tgt \\KF  $\downarrow$} & \makecell{Ctx \\ Root $\downarrow$} & \makecell{Tgt \\Root $\downarrow$} & Reach $\uparrow$ \\
\midrule
Ground Truth & N/A & N/A & 0.0533 & 0.022 & N/A & N/A & {N/A} & {N/A} & 3.47 & 1.96 & 0.0008 & 0.0009 & N/A & N/A & N/A & N/A & N/A & N/A \\
\midrule
Cond. Inbtwn. (2022) & {\textbf{27,000}} & \underline{2.4ms} & 0.1093 & 1.594 & 0.8\% & 1.83 & {0.00} & {0.00} & 16.88 & 12.65 & 0.018 & 0.034 & 51.8\% & 0.094 & \underline{0.078} & 0.050 & \underline{0.051} & \underline{87.7\%} \\
Two-stage (2023) & \underline{24,600} & 2.6ms & 0.1094 & 1.643 & 5.0\% & 2.18 & {0.00} & {0.00} & 38.39 & 33.72 & 0.038 & 0.037 & 43.7\% & 0.293 & 0.102 & 0.215 & 0.061 & 85.7\% \\
Delta-interp. (2024) & 18,800 & 3.4ms & 0.1107 & 1.774 & 1.3\% & 1.92 & {0.00} & {0.00} & 24.92 & 20.15 & 0.028 & 0.032 & 50.6\% & 0.288 & 0.130 & 0.152 & 0.066 & 81.9\% \\
\midrule
CondMDI (2024) & 1,930 & 33.2ms & \underline{0.1080} & 1.213 & \multirow{2}{*}{15.6\%} & \multirow{2}{*}{3.13} & {2.03} & {16.42} & 16.19 & 14.62 & 0.012 & \underline{0.009} & 62.7\% & \textbf{0.045} & 0.143 & 0.022 & 0.118 & 61.3\% \\
CondMDI + CFG (2024) & 1,050 & 60.5ms & 0.1082 & \underline{1.201} & & & {2.05} & {16.72} & 17.03 & 15.39 & 0.012 & 0.009 & 63.2\% & 0.049 & 0.121 & 0.024 & 0.097 & 65.9\% \\
\midrule
MMM (2024) & 3,600 & 18.1ms & 0.1176 & 1.544 &  \multirow{3}{*}{\underline{19.9\%}} &  \multirow{3}{*}{\underline{3.19}} & {\textbf{4.38}} & {\textbf{26.51}} & 5.40 & 3.11 & 0.005 & 0.011 & 86.5\% & 0.311 & 0.377 & 0.006 & 0.071 & 34.8\% \\
MMM + 5 steps & 2,700 & 24.5ms & 0.1225 & 1.562 & & & {3.21} & {\underline{20.97}} & 5.36 & 3.10 & \underline{0.004} & 0.011 & \underline{86.8\%} & 0.300 & 0.365 & 0.006 & 0.071 & 36.8\% \\
MMM + 10 steps & 1,400 & 46.2ms & 0.1234 & 1.564 & & & {2.96} & {19.62} & \underline{5.35} & \underline{3.09} & \underline{0.004} & 0.011 & \underline{86.8\%} & 0.300 & 0.364 & \underline{0.005} & 0.070 & 36.6\% \\
\midrule
Closd-DiP (2025) & 4,200 & 15.3ms & 0.1076 & 1.292 & \multirow{4}{*}{15.1\%} & \multirow{4}{*}{2.90} & {1.84} & {13.40} & 14.03 & 11.94 & 0.015 & \underline{0.010} & 52.7\% & 0.071 & 0.129 & 0.042 & 0.091 & 75.7\% \\
Closd-DiP + CFG & 2,800 & 23ms & 0.1084 & 1.288 & & & {1.70} & {12.45} & 18.33 & 15.98 & 0.022 & 0.012 & 50.5\% & 0.139 & 0.163 & 0.089 & 0.113 & 54.9\% \\
Closd-DiP + 5 steps & 14,500 & 4.4ms & 0.1081 & 1.338 & & & {1.51} & {10.86} & 17.56 & 15.54 & 0.017 & 0.012 & 47.4\% & \underline{0.069} & 0.122 & 0.040 & 0.082 & 78.2\% \\
Closd-DiP + CFG + 5 steps & 10,500 & 6.1ms & \underline{0.1080} & 1.319 & & & {1.40} & {10.08} & 20.85 & 18.60 & 0.023 & 0.013 & 47.7\% & 0.126 & 0.181 & 0.080 & 0.117 & 52.0\% \\
\midrule
\rowcolor{gray!15}
\textbf{Ours} & {15,000} & \textbf{2ms} & \textbf{0.1056} & \textbf{1.054} & \textbf{86.5\%} & \textbf{4.06} & {\underline{3.75}} & {\underline{20.67}} & \textbf{3.38} & \textbf{1.82} & \textbf{0.003} & \textbf{0.008} & \textbf{92.6\%} & \underline{0.052} & \textbf{0.076} & \textbf{0.001} & \textbf{0.023} & \textbf{99.6\%} \\
\bottomrule
\end{tabular}%
}
\end{table*}

\begin{figure*}[!t]
    \centering
    \includegraphics[width=1.0\textwidth]{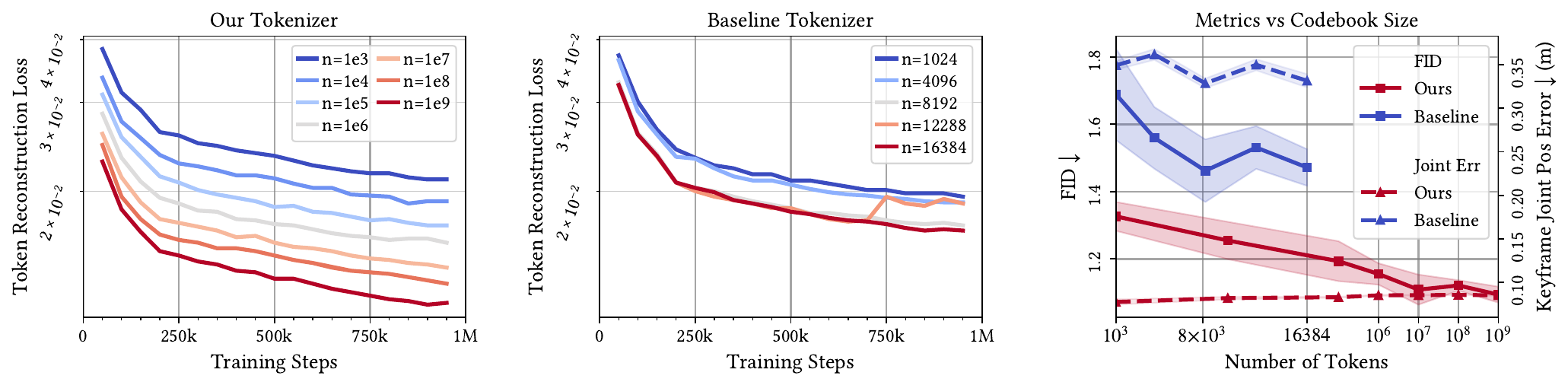}
    \caption{
    Scalability comparison between our multi-head tokenizer and a single-head baseline.
    Left and middle: Token reconstruction loss during training for varying codebook sizes for our method and the baseline.
    Our tokenizer continues to improve with larger codebooks (up to $10^9$ tokens), while the baseline plateaus quickly.
    Right: Trade-off between FID (distribution quality) and keyframe joint position error as codebook size increases.
    Our method achieves better FID with larger codebooks while maintaining low keyframe error.
    }\label{fig:ablation_num_tokens_scaling_metrics}
\end{figure*}


\begin{table*}[!t]
\centering
\caption{Quantitative comparison on \textbf{LaFAN1-G1}, \textbf{HumanML3D}, and \textbf{Bones-70k} datasets. 
$\downarrow$ indicates lower is better, $\uparrow$ indicates higher is better.
Best results are in \textbf{bold}, second best are \underline{underlined}.}
\label{tab:other_datasets_comparison}
\resizebox{\textwidth}{!}{%
\begin{tabular}{c|l|cc|cc|cc|ccc|ccccc}
\toprule
\multirow{2}{*}{Dataset} & \multirow{2}{*}{Method} & \multicolumn{2}{c|}{Distribution} & \multicolumn{2}{c|}{Diversity} & \multicolumn{2}{c|}{Smoothness} & \multicolumn{3}{c|}{Physical Plausibility} & \multicolumn{5}{c}{Precision} \\
& & MMD $\downarrow$ & FID $\downarrow$ & {Jnt $\uparrow$} & {Latent $\uparrow$} & Jnt Jit. $\downarrow$ & Root Jit. $\downarrow$ & Foot Sk. $\downarrow$ & Pene. $\downarrow$ & Foot Acc. $\uparrow$ & Ctx KF $\downarrow$ & Tgt KF $\downarrow$ & Ctx Root $\downarrow$ & Tgt Root $\downarrow$ & Reach $\uparrow$ \\
\midrule
\multirow{7}{*}{\rotatebox{90}{LaFAN1-G1}} 
& Cond. Inbtwn. (2022) & 0.2897 & 1.693 & {0.00} & {0.00} & 40.73 & 38.17 & 0.030 & 0.019 & 83.8\% & 0.374 & 0.453 & 0.264 & 0.369 & 11.5\% \\
& Two-stage (2022) & 0.3083 & 1.985 & {0.00} & {0.00} & 85.06 & 81.31 & 0.036 & 0.021 & 83.4\% & 0.535 & 0.498 & 0.400 & 0.399 & 8.1\% \\
& Delta-interp. (2024) & 0.3060 & 1.846 & {0.00} & {0.00} & 45.27 & 43.19 & 0.028 & 0.020 & 86.1\% & 0.404 & 0.451 & 0.306 & 0.360 & 11.6\% \\
& CondMDI (2024) & 0.2860 & \underline{1.004} & {0.75} & {4.08} & 17.89 & 16.93 & 0.014 & \underline{0.004} & 82.8\% & \underline{0.057} & 0.592 & 0.029 & 0.555 & 6.6\% \\
& MMM (2024) & 0.2920 & 1.189 & {\textbf{1.78}} & {\textbf{17.66}} & \underline{7.00} & \underline{4.05} & \underline{0.006} & \underline{0.004} & \textbf{90.5\%} & 0.289 & 0.331 & \underline{0.008} & \underline{0.103} & 37.3\% \\
& Closd-DiP (2025) & \underline{0.2839} & 1.165 & {\underline{1.72}} & {\underline{12.07}} & 13.96 & 12.41 & 0.013 & 0.005 & 85.5\% & 0.080 & \textbf{0.178} & 0.049 & 0.128 & \underline{50.7\%} \\
& \cellcolor{gray!15}\textbf{Ours} & \cellcolor{gray!15}\textbf{0.2835} & \cellcolor{gray!15}\textbf{0.891} & \cellcolor{gray!15}{\underline{1.76}} & \cellcolor{gray!15}{11.42} & \cellcolor{gray!15}\textbf{4.53} & \cellcolor{gray!15}\textbf{2.53} & \cellcolor{gray!15}\textbf{0.003} & \cellcolor{gray!15}\textbf{0.003} & \cellcolor{gray!15}\underline{88.9\%} & \cellcolor{gray!15}\textbf{0.053} & \cellcolor{gray!15}\underline{0.179} & \cellcolor{gray!15}\textbf{0.006} & \cellcolor{gray!15}\textbf{0.100} & \cellcolor{gray!15}\textbf{61.1\%} \\
\midrule
\multirow{7}{*}{\rotatebox{90}{HumanML3D}} 
& Cond. Inbtwn. (2022) & 0.1107 & 1.497 & {0.00} & {0.00} & 10.07 & 8.04 & 0.019 & 0.016 & 62.7\% & 0.126 & \textbf{0.090} & 0.051 & 0.050 & \underline{92.6\%} \\
& Two-stage (2022) & 0.1131 & 1.709 & {0.00} & {0.00} & 19.78 & 16.51 & 0.030 & 0.017 & 61.6\% & 0.232 & 0.132 & 0.144 & 0.073 & 75.6\% \\
& Delta-interp. (2024) & 0.1133 & 1.585 & {0.00} & {0.00} & 10.86 & 8.74 & 0.031 & 0.015 & 60.1\% & 0.288 & 0.151 & 0.132 & 0.090 & 68.0\% \\
& CondMDI (2024) & \textbf{0.1101} & \underline{1.054} & {1.49} & {\underline{11.09}} &5.82 & 4.87 & 0.009 & \textbf{0.002} & 77.4\% & \textbf{0.037} & 0.147 & 0.016 & 0.121 & 52.3\% \\
& MMM (2024) & 0.1238 & 1.068 & {\underline{1.89}} & {\underline{11.23}} & \underline{2.92} & \underline{1.94} & \textbf{0.005} & \underline{0.002} & \textbf{89.2\%} & 0.242 & 0.293 & \underline{0.003} & \underline{0.037} & 61.2\% \\
& Closd-DiP (2025) & \underline{0.1104} & 1.192 & {1.28} & {9.09} & 6.12 & 5.06 & 0.012 & 0.003 & 69.9\% & 0.062 & \underline{0.105} & 0.035 & 0.070 & 80.8\% \\
& \cellcolor{gray!15}\textbf{Ours} & \cellcolor{gray!15}0.1114 & \cellcolor{gray!15}\textbf{0.914} & \cellcolor{gray!15}{\textbf{2.56}} & \cellcolor{gray!15}{\textbf{13.47}} & \cellcolor{gray!15}\textbf{2.08} & \cellcolor{gray!15}\textbf{1.29} & \cellcolor{gray!15}\underline{0.006} & \cellcolor{gray!15}\underline{0.002} & \cellcolor{gray!15}\underline{83.5\%} & \cellcolor{gray!15}\underline{0.050} & \cellcolor{gray!15}0.141 & \cellcolor{gray!15}\textbf{0.001} & \cellcolor{gray!15}\textbf{0.022} & \cellcolor{gray!15}\textbf{98.9\%} \\
\midrule
\multirow{7}{*}{\rotatebox{90}{Bones-70k}} 
& Cond. Inbtwn. (2022) & 0.1073 & 1.559 & {0.00} & {0.00} & 18.18 & 13.91 & 0.019 & 0.031 & 49.4\% & 0.088 & \underline{0.091} & 0.044 & \underline{0.062} & \underline{83.5\%} \\
& Two-stage (2022) & 0.1081 & 1.604 & {0.00} & {0.00} & 39.32 & 34.11 & 0.039 & 0.037 & 44.3\% & 0.347 & 0.128 & 0.257 & 0.076 & 77.2\% \\
& Delta-interp. (2024) & 0.1183 & 2.024 & {0.00} & {0.00} & 39.32 & 32.79 & 0.048 & 0.042 & 44.8\% & 0.556 & 0.499 & 0.226 & 0.129 & 41.4\% \\
& CondMDI (2024) & \underline{0.1062} & \underline{1.191} & {2.11} & {\underline{16.41}} & 16.62 & 15.10 & 0.012 & \underline{0.009} & 60.7\% & \textbf{0.049} & 0.190 & 0.024 & 0.160 & 48.5\% \\
& MMM (2024) & 0.1227 & 1.494 & {\underline{2.38}} & {\underline{16.57}} & \underline{5.27} & \underline{3.01} & \underline{0.005} & 0.011 & \underline{85.6\%} & 0.306 & 0.375 & \underline{0.006} & 0.074 & 40.0\% \\
& Closd-DiP (2025) & \underline{0.1062} & 1.305 & {1.86} & {13.13} & 14.14 & 12.10 & 0.015 & 0.011 & 50.8\% & 0.065 & 0.136 & 0.036 & 0.102 & 68.6\% \\
& \cellcolor{gray!15}\textbf{Ours} & \cellcolor{gray!15}\textbf{0.1050} & \cellcolor{gray!15}\textbf{1.090} & \cellcolor{gray!15}{\textbf{3.48}} & \cellcolor{gray!15}{\textbf{20.09}} & \cellcolor{gray!15}\textbf{3.24} & \cellcolor{gray!15}\textbf{1.80} & \cellcolor{gray!15}\textbf{0.004} & \cellcolor{gray!15}\textbf{0.009} & \cellcolor{gray!15}\textbf{84.6\%} & \cellcolor{gray!15}\underline{0.054} & \cellcolor{gray!15}\textbf{0.087} & \cellcolor{gray!15}\textbf{0.001} & \cellcolor{gray!15}\textbf{0.023} & \cellcolor{gray!15}\textbf{99.6\%} \\
\bottomrule
\end{tabular}%
}
\vspace{-2mm}
\end{table*}

In this section, we evaluate \ourmethod{} across different datasets in both animation and robotics applications.
Section~\ref{section:exp_dataset} describes the datasets used in our experiments.
Section~\ref{section:exp_numerical_evaluation} presents numerical comparisons with recent state-of-the-art methods.
Section~\ref{section:exp_ablation_studies} provides ablation studies analyzing the contribution of different components.
Section~\ref{section:exp_interactive_animation_and_robotics_applications_and_engineering_details} discusses interactive animation, robotics applications, and engineering details.

\subsection{Dataset}\label{section:exp_dataset}
We evaluate our method on four datasets of varying scales.
Our primary dataset is a proprietary motion capture collection containing approximately 700 hours of high-quality motion data with 350k motion clips, covering diverse actions including locomotion, combat, sports, and object interactions.
Without further specification, we call it \textbf{350k} dataset.
The 350k dataset covers 9,300 unique skills across 36 categories, captured from over 163 performers, as shown in Figure~\ref{fig:dataset_distribution_by_categories}.
The test set comprises 10\% of the full dataset and consists of two 5\% subsets: one split randomly, and the other split by motion skill category to maximize distributional difference from the training set.
The dataset, along with download and usage instructions, is available on our project website.
Additional statistics are provided in Appendix~\ref{section:appendix_dataset_statistics}.

For public benchmark comparison, we use \textbf{HumanML3D}~\cite{guo2022generating},
an open-source dataset with 28.6 hours and 14,616 motion sequences.
We removed broken clips and segment motions that are too long, so the final number of training clips is 23,206 and test clips is 2,578.
Additionally, we evaluate on \textbf{LaFAN1-G1}, a variant of LaFAN1~\cite{harvey2020robust} retargeted to the Unitree G1 humanoid skeleton, which contains 4.6 hours with 2.4k clips and has become a common benchmark for robotics research.
Since most LaFAN1 motion clips are quite long, we cut them into 6s clips.
We also added four end-effector joints to the G1 skeleton, which has 30 joints (29 hinge joints and a free pelvis joint) so that in total we have 34 joints.
We also train on a 140-hour subset of our proprietary dataset (referred to as \textbf{70k}) with 62k clips to study scaling behavior.
For all four datasets, we use 30 FPS and retrain our method and baselines for fair comparison.

In total, as shown in Table~\ref{tab:dataset_comparison}, we have 4 datasets covering different scales of the datasets and skeletons of different number of joints and morphologies across animation and robotics domains.
We use the same motion representation as described in Section~\ref{section:tokenizer}.

\subsection{Quantitative Evaluation}\label{section:exp_numerical_evaluation}
To demonstrate the effectiveness of our method, we compare with the following state-of-the-art methods:
1) Cond. In-betweening~\cite{kim2022conditional},
2) Delta-interpolator~\cite{oreshkin2023motion},
3) Two-stage Trans.~\cite{qin2022motion},
4) CondMDI~\cite{cohan2024flexible},
5) MMM~\cite{pinyoanuntapong2024mmm}, and
6) Closd-DiP~\cite{tevetclosd}.
The first three are deterministic methods using transformers to predict in-between motions directly, with the two-stage transformer method employing a cascade of two transformers for refinement.
CondMDI and Closd-DiP are generative diffusion models.
The former uses a U-Net architecture specifically designed for motion in-betweening,
while the latter employs a closed-loop framework between physical simulation and a diffusion planner.
In our experiments, we use only the diffusion planner of Closd-DiP for a fair comparison.
MMM also uses a token-based approach but employs a single-head tokenizer without root-pose disentanglement or progressive design.
We modified the open-source code repositories to support evaluation on our datasets, using the same motion representation and training settings as the original methods.
All methods are formulated for the in-betweening task for our evaluation.

\paragraph{Evaluation Metrics.}
We randomly sample the context and target keyframes from the test set and generate motions between the two keyframes.
We evaluate methods across seven categories of metrics:
\textbf{(1) Speed}: We report frames per second (FPS) and latency in milliseconds to measure real-time performance.
We take the latency from the original methods' paper directly if the values are provided; otherwise, we use the modified open-source code to estimate the FPS and latency.
\textbf{(2) Distribution}: We use Maximum Mean Discrepancy (MMD)~\cite{jayasumana2024rethinking} and Fréchet Inception Distance (FID)~\cite{heusel2017gans} to measure how well the generated motion distribution matches the ground truth distribution.
To obtain the MMD and FID scores, we retrain a motion autoencoder following TMR~\cite{petrovich2023tmr}, omitting the text branch, and compute both metrics in the learned latent space.
Following~\cite{jayasumana2024rethinking}, we scale the distribution metrics for more human-readable values.
We provide the groundtruth FID and MMD values for reference.
\textbf{(3) Human Evaluation}:
We conduct a user study with 40 participants from diverse backgrounds, including artists, technical specialists, engineers, and scientists from the animation and robotics domain.
For each trial, participants view motions generated by 3 randomly selected methods on the same in-betweening setup and select a single winner.
Win rate is computed as the percentage of trials where a method is selected as the best.
Participants also rate each motion on a 1--5 Likert scale for overall quality.
\textbf{(4) Diversity}: Following~\cite{tevet2022human}, we run each generation 20 times and compute two diversity metrics: joint-level diversity, which measures the variability of joint positions across generated motions, and latent-level diversity, which similarly measures the mean pairwise distance in the learned latent space.
\textbf{(5) Smoothness}: Joint Jitter and Root Jitter (both in m/s$^2$) measure the acceleration magnitude of joint positions and root trajectory, respectively, as was proposed in~\cite{gou2025control}. It aligns well with the human perception of smoothness, and lower values indicate smoother motion.
\textbf{(6) Physical Plausibility}: Foot Skate (m/frame) measures undesirable sliding of feet during ground contact, Penetration (m) measures ground penetration depth, and Foot Contact Accuracy (\%) measures how often foot contact predictions align with ground truth labels. More accurate foot prediction will be helpful for postprocessing in downstream applications when necessary.
\textbf{(7) Precision}: Context/Target Keyframe errors (m) measure joint position deviation from the input keyframes, Context/Target Root errors (m) measure root position deviation, and Reaching Success (\%) measures whether the generated motion successfully reaches the target keyframe within a threshold.
Note that instead of using mean per joint position error (MJPE or MPJPE), we use the \textbf{max} joint position error to measure the precision of the generated motion.
We argue that max per joint position error is more sensitive and aligned with the human perception.
We define Reaching Success (\%) as the percentage of the generated motion that successfully reaches the target keyframe's root position within 5 cms and 15 degrees of the target keyframe's heading angle.

\begin{figure*}[!t]
    \centering
    \includegraphics[width=1.\textwidth]{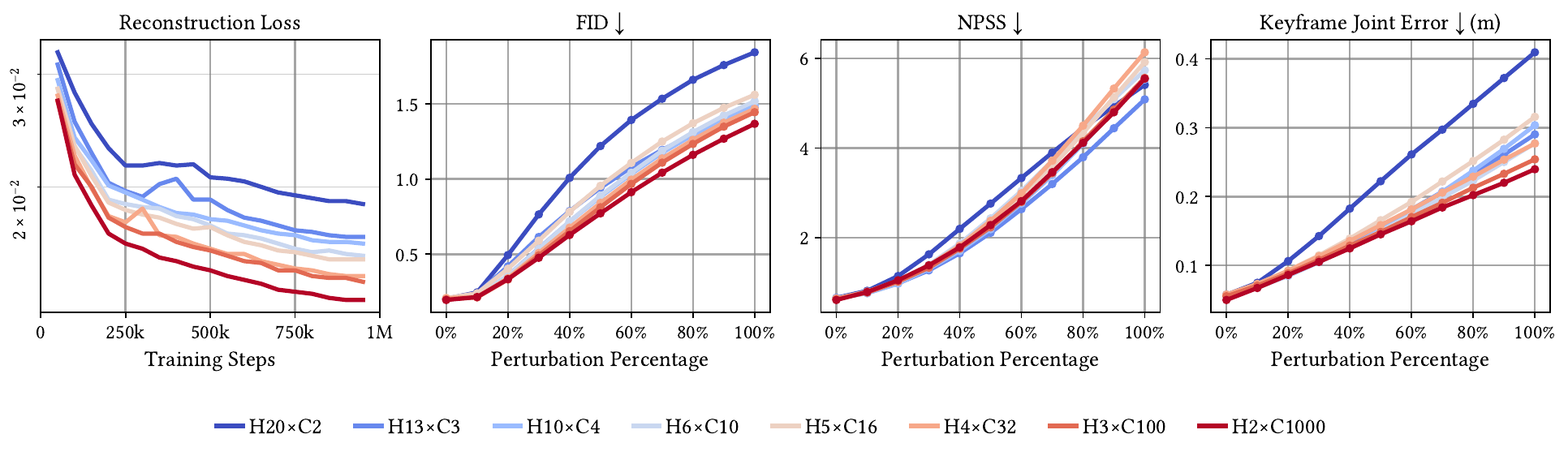}
    \vspace{-20pt}
    \caption{
    Ablation study on multi-head tokenization with fixed total codebook capacity ($\sim10^6$ tokens).
    ``H'' denotes the number of heads, ``C'' denotes the codebook size per head in the experiment name.
    Left: tokenizer's reconstruction loss during training.
    Right three plots: FID, NPSS, and keyframe error under token perturbation, simulating lower-bound motion quality in real applications.
}\label{fig:ablation_tokens_design_choice_metrics}
\end{figure*}
\begin{figure}[!t]
    \centering
    \includegraphics[width=0.48\textwidth]{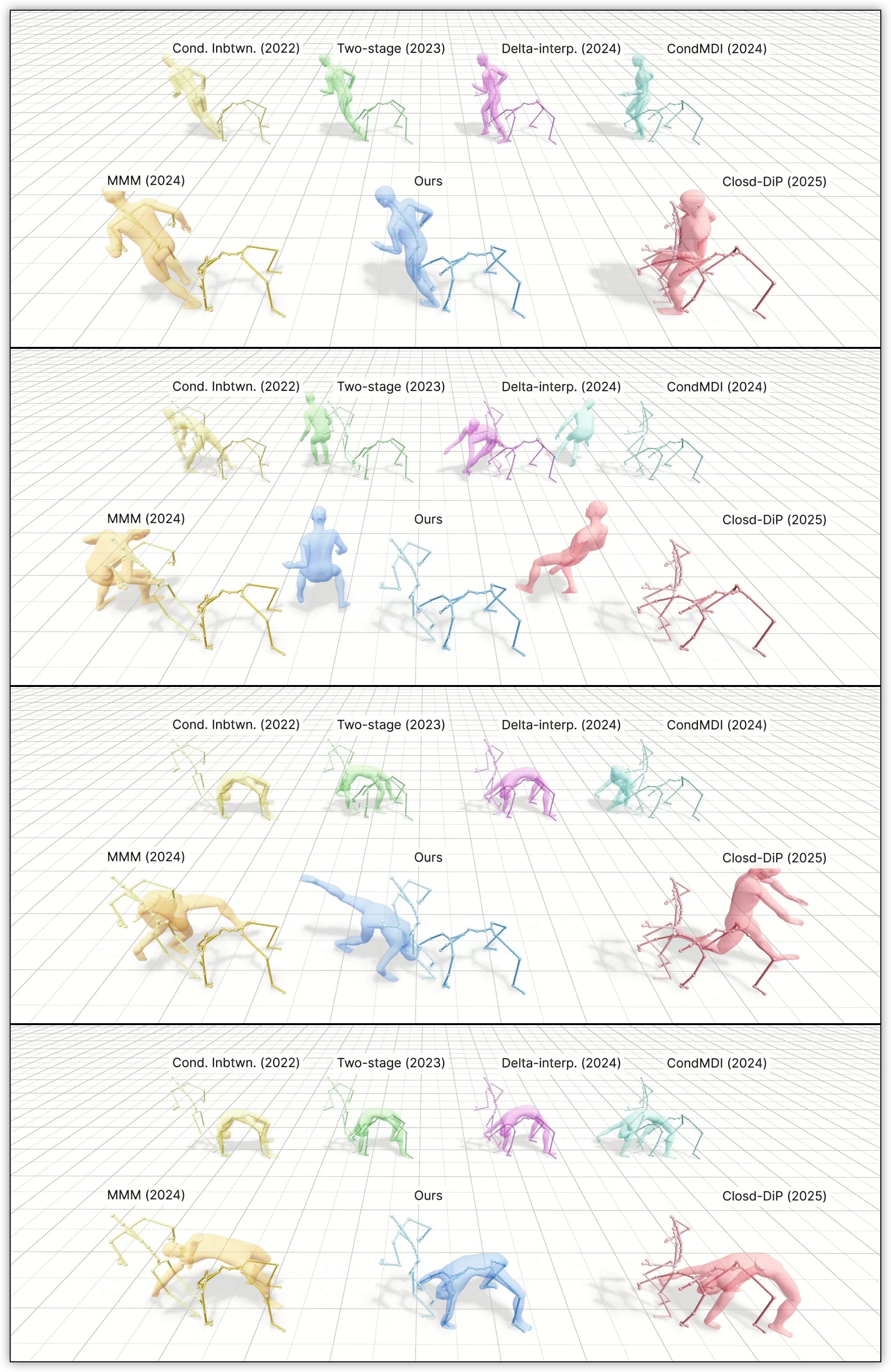}
    \caption{
        An in-betweening test case transitioning from running left to a backbend pose. Time progresses from top to bottom; algorithm names are labeled in each subfigure. \ourmethod{} is shown as the blue character.
    }
    \label{fig:ablation_inbetween_comparison}
\end{figure}

\paragraph{Results Summary.}
As shown in Table~\ref{tab:350k_comparison} and Table~\ref{tab:other_datasets_comparison}, our method achieves consistently better performance than the baseline methods on all datasets. 
Key improvements on the 350k dataset include: better distribution matching including FID and MMD, lower jitter, better keyframe precision and near-perfect reaching success.
Our algorithm also receives the highest score and win rate in human evaluation.
We further provide visual in-betweening comparisons in Figure~\ref{fig:ablation_inbetween_comparison}, where \ourmethod{} exhibits notably better physical plausibility, naturalness, and keyframe adherence than the baselines. More results are provided in Appendix~\ref{section:appendix_inbetweening_visual_comparisons}.
For diversity metrics, \ourmethod{} outperforms baselines on larger datasets, while on smaller datasets such as HumanML3D and LaFAN1, baseline methods perform comparably or marginally better. Overall, there is no significant trade-off between diversity and precision or speed.
We observe that non-generative methods achieve reasonable precision but have bad FID, MMD scores, and poor human evaluation scores,
while diffusion-based methods improve distribution quality but struggle with satisfying spatial constraints.
We also note that having more iterations in MMM gives minor improvements and classifier-free guidance for CondMDI and ClosdDip gives minor improvements to the metrics.
Our method also achieves one of the best runtime efficiency among all methods, with only 2ms latency to generate the motion between two keyframes, achieving 15,000 FPS throughput.

\begin{figure}[!t]
    \centering
    \includegraphics[width=0.50\textwidth]{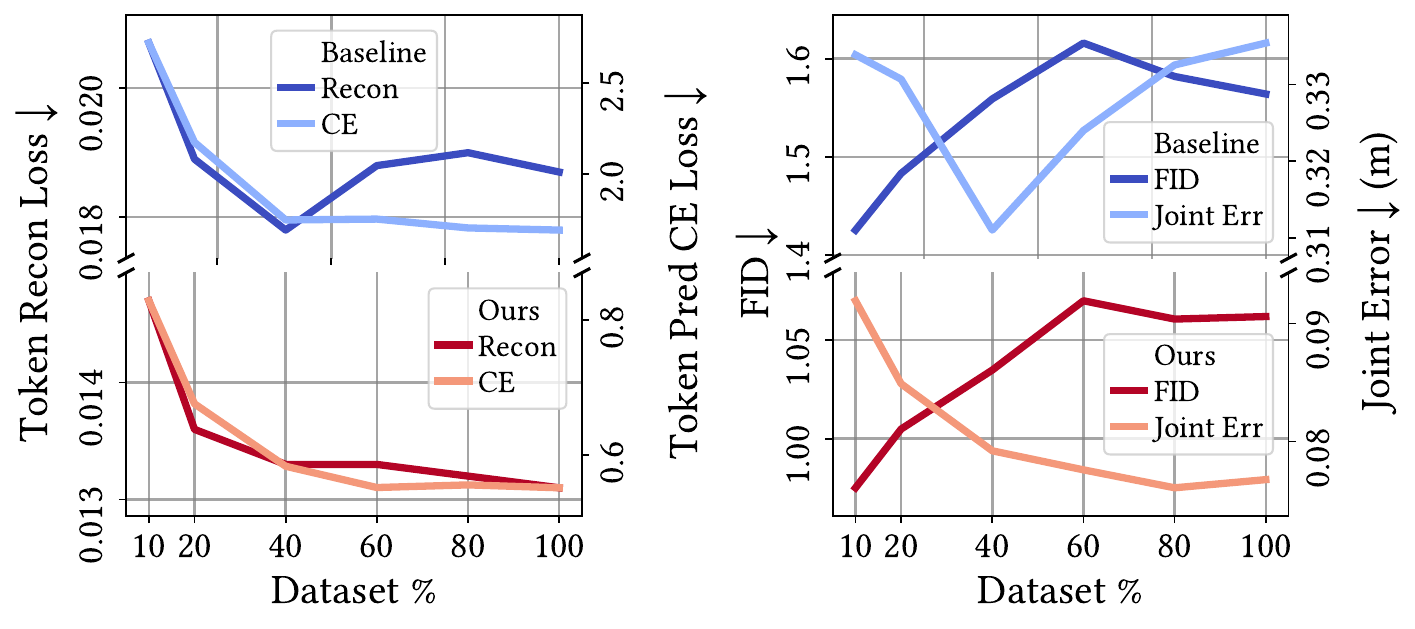}
    \caption{
    Scaling behavior with dataset size from 10\% to 100\% of the 350k dataset.
    Left: Evaluation losses including the reconstruction loss for the tokenizer, and cross-entropy for token prediction.
    Right: FID and keyframe joint position error.
    Our method scales effectively with more data, while the baseline struggles at larger scales.}
    \label{fig:ablation_dataset_scaling_metrics}
\end{figure}
\begin{figure}[!t]
    \centering
    \includegraphics[width=0.48\textwidth]{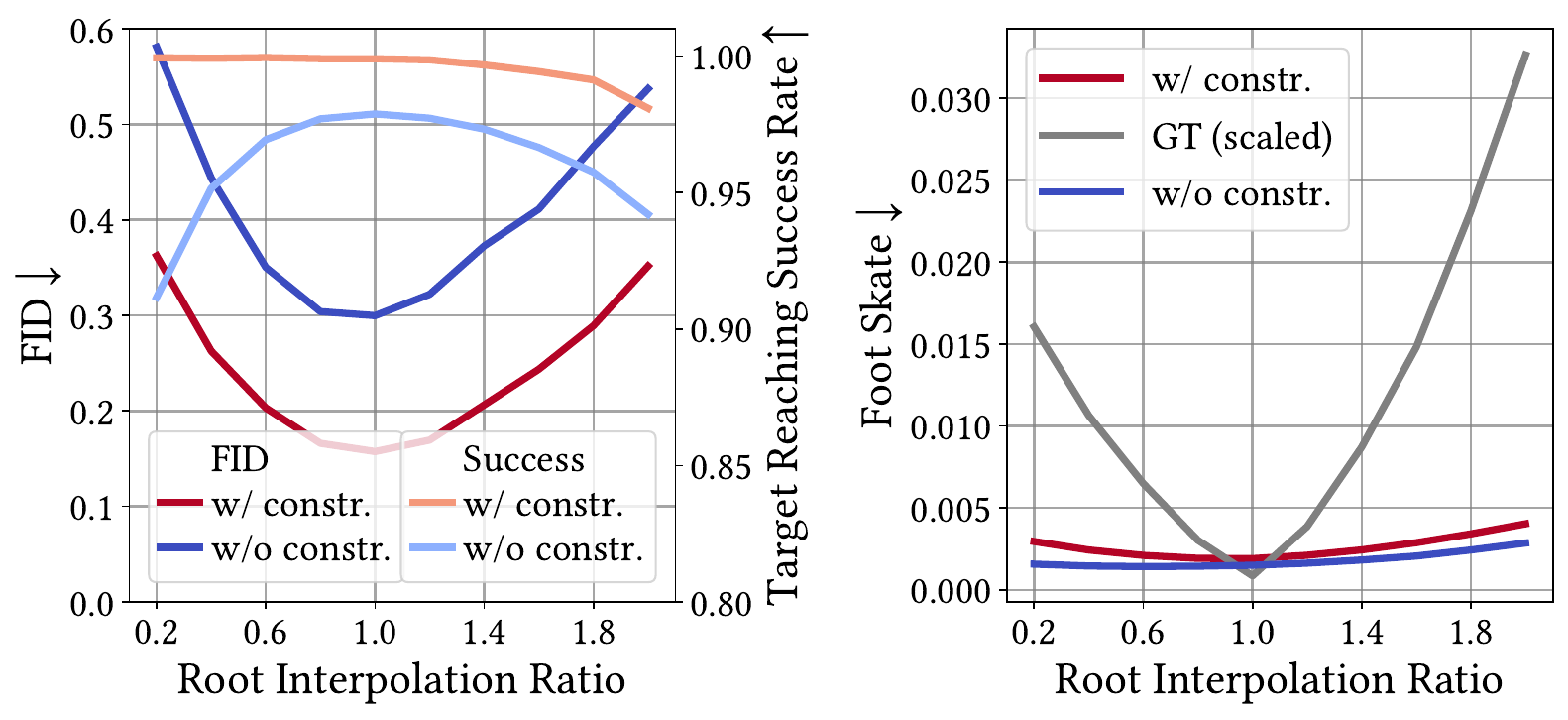}
    \caption{
    Root trajectory interpolation analysis.
    Root interpolation ratio of 1.0 is the original trajectory; $<$1.0 compresses, $>$1.0 stretches.
    Left: FID and target reaching success rate remain stable across interpolation ratios, especially with keyframe constraints.
    Right: Foot skate stays low even under significant root manipulation, demonstrating the decoder's robustness.
}
    \label{fig:ablation_root_interpolation_metrics}
\end{figure}
\subsection{Ablation Studies and Analysis}\label{section:exp_ablation_studies}

\paragraph{Scalability of \ourmethod{} and baseline.}
As shown in Figure~\ref{fig:ablation_num_tokens_scaling_metrics}, we compare the scalability of our structured multi-head tokenizer against a standard single-head VQ-VAE baseline.
Our method (left) demonstrates consistent performance improvements as the number of tokens increases, while the baseline (middle) plateaus quickly regardless of codebook size.
In the right figure, we analyze the trade-off between distribution quality (FID) and keyframe precision as we scale from $10^3$ to $10^9$ tokens.
Our method achieves better FID with more tokens but shows slightly increased keyframe errors at extreme scales, suggesting an optimal operating point around $10^6$ to $10^7$ tokens that balances both objectives.
In contrast, the baseline exhibits no meaningful improvement in either FID or keyframe error with increased capacity, and maintains significantly worse absolute metrics across all scales.
This demonstrates that our structured multi-head design is essential for effectively utilizing model capacities.

\paragraph{Multi-head tokenization.}
In Figure~\ref{fig:ablation_tokens_design_choice_metrics}, we ablate different combinations of the number of heads and tokens per head while keeping the total number of tokens fixed at approximately $10^6$.
We evaluate the tokenizer's reconstruction loss, FID, the normalized power spectrum similarity score (NPSS)~\cite{gopalakrishnan2019neural}, and keyframe errors under token perturbation.
We introduce token perturbation to simulate prediction errors and assess how gracefully the model degrades, i.e., the lower-bound of the generated motion quality in potential real applications.
We observe that increasing tokens per head raises reconstruction error, though the downstream metrics (FID, NPSS, keyframe errors) remain relatively stable.
However, the perturbation analysis reveals a trade-off: too many tokens per head degrades NPSS (temporal coherence), while too few tokens per head hurts FID and keyframe precision.
Based on these results, we recommend 128 to 256 tokens per head as the optimal configuration,
combined with a total codebook capacity of approximately $10^9$ tokens for the best overall performance.

\paragraph{Scaling behavior with the size of the dataset.}
In Figure~\ref{fig:ablation_dataset_scaling_metrics}, we investigate how model performance scales with dataset size by training on progressively larger subsets of our 350k dataset.
This analysis provides guidance for practitioners training on datasets of varying sizes.
Both our method and baselines show improved tokenizer and generative model metrics with increasing data.
However, the baseline struggles significantly on larger datasets due to its limited capacity and scalability, which becomes overwhelmed and exhibits degraded performance.
Interestingly, FID scores worsen with larger datasets for both methods. We hypothesize that models trained on smaller datasets may overfit and produce less diverse outputs that happen to match the training distribution more closely, resulting in artificially lower FID scores.
In contrast, keyframe precision consistently improves with more data, and our method maintains this improvement while the baseline's keyframe errors increase as the dataset size grows.
This demonstrates that our architecture is better suited to leveraging large-scale motion data while still working well with small datasets.

\paragraph{Root interpolation designs.}
Our decoder supports root trajectory interpolation, where we keep the predicted pose tokens fixed but override the root values with interpolated trajectories.
This is a practical utility for runtime animation, enabling precise global keyframe matching without introducing foot sliding artifacts as mentioned earlier in Section~\ref{section:smart_locomotion}.
As shown in numerical results in Figure~\ref{fig:ablation_root_interpolation_metrics} and visual results in Figure~\ref{fig:root_interpolation_g1},
our design successfully hits target keyframes while maintaining FID quality and avoiding foot skating degradation when interpolating root trajectories.
In Figure~\ref{fig:ablation_root_interpolation_metrics}, we also compare decoding with and without pose keyframe constraints (as shown in Figure~\ref{fig:input_constraints}).
Providing pose constraints improves both FID and target reaching success, demonstrating the value of our flexible constraint handling.
This robustness to root manipulation stems from our modular architecture, where the pose decoder is trained to be agnostic to the source of root information.

\paragraph{Additional Ablation Studies}
We provide additional ablation studies in the appendix: (1) the impact of varying the number of GPUs during training (Appendix~\ref{section:appendix_gpu_scaling}), (2) the choice of tokenizer design between FSQ and VQ-VAE (Appendix~\ref{section:appendix_fsq_vqvae}), and (3) the effect of replanning frequency during deployment (Appendix~\ref{section:replanning_frequency}).

\subsection{Interactive Animation, Robotics Applications and Engineering Details}\label{section:exp_interactive_animation_and_robotics_applications_and_engineering_details}
As shown in Figure~\ref{fig:teaser} and the demo video, we showcase our method in UE5 and deploy it on the Unitree G1 humanoid robot.

We train the tokenizer, root module, and pose module on 4 nodes with 8 GPUs each (32 GPUs total); training with 16 GPUs yields similar visual quality.
Training runs for 2 million updates with a batch size of 256 per GPU, using Adam optimizer with learning rate $5 \times 10^{-5}$ and a cosine schedule with 10k warmup steps decaying to $2 \times 10^{-6}$.
The tokenizer uses a U-Net architecture with a downsampling rate of 4 (2 downsampling layers), where each layer consists of three residual 1D convolutional layers with 1024 channels and progressively larger kernel sizes (23.5M parameters total).
The encoder and decoder have a symmetric architecture.
We also experimented with transformer-based encoder and decoder designs, which achieve similar performance but are significantly slower during inference.
FSQ~\cite{mentzer2023finite} and VQ-VAE yield similar performance (see Appendix~\ref{section:appendix_fsq_vqvae} for a detailed comparison).
We also include standard foot sliding and velocity losses during tokenizer training to further improve motion quality,
particularly important for retargeted datasets where artifacts like foot skating are more common.
The pose module is a transformer with embedding dimension 1024, 16 attention heads, and 16 layers (150M parameters).
The root module uses a similar architecture with fewer layers and 512 embedding dimension and 12 attention heads (50M parameters).
Training takes approximately 7 days for the tokenizer, 3 days for the root module, and 7 days for the pose module on H100 GPUs, though Figure~\ref{fig:ablation_dataset_scaling_metrics} suggests performance plateaus earlier.
Training on L40S GPUs yields similar performance but is approximately 2$\times$ slower.
During inference, we achieve 2ms latency and 15,000 FPS throughput on a desktop with an RTX 5090 GPU.
For both the UE5 and G1 applications, we use the same model architecture and training settings.

For the UE5 application, we utilize UE5's plugin system to implement both our neural backbone and smart primitives.
Models are exported via ONNX and loaded in a native C++ plugin using TensorRT.
Since the training skeleton differs from the demo characters, we use UE5's built-in real-time retargeter~\cite{epicgames2025ikriganimationretargetinginunrealengine} to retarget motions on the fly to the ``Messenger'' and ``Guard'' characters.
The authoring time for smart navigation and smart object is similar, each taking less than 10 minutes for non-expert users.
More details about the UE5 plugin are provided in Appendix~\ref{section:appendix_ue5_demo}.

By combining our method with a physical tracking controller~\cite{wang2020unicon,chen2025gmt,luo2025sonic,zhang2025track,zeng2025behavior,yin2025unitrackerlearninguniversalwholebody}, we drive real robots by supplying the generated kinematic motions as tracking targets.
In our experiments, we use the tracking controller from~\citet{luo2025sonic} to control the G1 humanoid robot.
For G1 applications, we retarget our proprietary dataset to the G1 skeleton using a modified version of GMR~\cite{araujo2025retargeting,ze2025gmr}.
Automatic retargeting introduces some artifacts, but the results remain acceptable due to the scale of the dataset.
For deployment, we use the Jetson Orin platform and Unitree's official SDK, with models exported via ONNX and loaded using TensorRT similar to the UE5 application.
Due to Orin's limited compute, latency increases to 5ms per inference.
Replanning is triggered at 10 Hz or whenever commands change, also similar to the strategy used in the UE5 demo.
This lazy replanning strategy is more important for robotics deployment, as it avoids interference and competition of resources with the low-level control policy, which is sensitive to latency.
\section{Discussion}
In this work, we introduced \ourmethod{}, a real-time motion synthesis framework designed to bridge the gap between recent advances in generative models and the practical demands of animation and robotics production.
By combining a scalable latent neural backbone with intuitive smart primitive interfaces,
our approach enables diverse task-agnostic motion generation without fine-tuning or tagging.
Through experiments on datasets of varying scales and demonstrations in both UE5 and on the Unitree G1 robot, we showed that a unified framework can address motion control across virtual characters and physical robots.

\subsection{Limitations and Future Work}\label{section:limitations}

Dataset Scale:
Our dataset of 350,000 motions, while substantial, remains small with limited coverage of rare motion types (e.g., only one clip for vaulting at 0.5m). Additionally, object interactions are not explicitly modeled. Continued dataset expansion~\cite{hymotion2025} and leveraging video foundation models~\cite{ni2025generated, wen2025efficient} offer promising paths forward.

Visual Planning:
Real-world deployment lacks access to privileged simulation information such as ground-truth object poses and terrain geometry, disabling our smart object primitive. Robots must instead rely on noisy sensor input (RGB-D, SLAM). Developing kinematic planners driven by visual input~\cite{he2025viral,zhu2026hiking,chen2025hand} would benefit both robotics and animation domains.

Physical Awareness:
Kinematic planners like \ourmethod{} may generate physically implausible motions (self-collisions) or motions exceeding hardware constraints (e.g., dangerous flips). Co-training the kinematic planner and tracking policy could address this, enabling the planner to produce feasible motions while the policy adapts to the planner's output distribution.

Retargeting:
Adapting motions across diverse morphologies remains challenging. Runtime retargeting (used in our UE5 demo) is fast but low-quality; offline retargeting (used for G1) is accurate but requires months of iteration. Recent work~\cite{araujo2025retargeting,ze2025gmr,yang2025omniretarget,kim2025pyroki} shows promise in bridging this gap.
\section*{Acknowledgments}
We thank Cyrus Hogg, John Malaska, Will Telford, Jon Shepard, Simon Ouellet, Dmitry Korobchenko, Anna Minx, Edy Lim, Eugene Jeong, Sam Wu, Ehsan Hassani, Charles Zhou, Freya Li, Ling Li, Qiao Wang, and Lina Song for their support and guidance throughout the project.
We thank Alberto Guerra, Boon Cotter, Byeong Gyu Park, Craig Christian, Gabriele Leone, Yenal Kal, Pierre Fleau, Miguel Guerrero, and Marc-Andre Carbonneau from the art team for their help in creating the Unreal Engine~5 demo and its assets.
We also thank Hung Yu Ling, Yue Zhao, Danila Krivenkov, Evgenii Tumanov, Morteza Ramezanali, Winston Chen, Yu-Shiang Wong, Jiefeng Li, Yeongho Seol, Jun Saito, Michael Buttner, Haotian Zhang, Yifeng Jiang, Chen Tessler, Sanja Fidler, Umar Iqbal, Jan Kautz, Yan Chang, and Jim Fan for their helpful discussions and feedback.
\bibliographystyle{ACM-Reference-Format}
\bibliography{sample-base}


\begin{thebibliography}{100}


\ifx \showCODEN    \undefined \def \showCODEN     #1{\unskip}     \fi
\ifx \showISBNx    \undefined \def \showISBNx     #1{\unskip}     \fi
\ifx \showISBNxiii \undefined \def \showISBNxiii  #1{\unskip}     \fi
\ifx \showISSN     \undefined \def \showISSN      #1{\unskip}     \fi
\ifx \showLCCN     \undefined \def \showLCCN      #1{\unskip}     \fi
\ifx \shownote     \undefined \def \shownote      #1{#1}          \fi
\ifx \showarticletitle \undefined \def \showarticletitle #1{#1}   \fi
\ifx \showURL      \undefined \def \showURL       {\relax}        \fi
\providecommand\bibfield[2]{#2}
\providecommand\bibinfo[2]{#2}
\providecommand\natexlab[1]{#1}
\providecommand\showeprint[2][]{arXiv:#2}

\bibitem[Aberman et~al\mbox{.}(2020)]%
        {aberman2020skeleton}
\bibfield{author}{\bibinfo{person}{Kfir Aberman}, \bibinfo{person}{Peizhuo Li}, \bibinfo{person}{Dani Lischinski}, \bibinfo{person}{Olga Sorkine-Hornung}, \bibinfo{person}{Daniel Cohen-Or}, {and} \bibinfo{person}{Baoquan Chen}.} \bibinfo{year}{2020}\natexlab{}.
\newblock \showarticletitle{Skeleton-aware networks for deep motion retargeting}.
\newblock \bibinfo{journal}{\emph{ACM Transactions on Graphics (TOG)}} \bibinfo{volume}{39}, \bibinfo{number}{4} (\bibinfo{year}{2020}), \bibinfo{pages}{62--1}.
\newblock


\bibitem[Achiam et~al\mbox{.}(2023)]%
        {achiam2023gpt}
\bibfield{author}{\bibinfo{person}{Josh Achiam}, \bibinfo{person}{Steven Adler}, \bibinfo{person}{Sandhini Agarwal}, \bibinfo{person}{Lama Ahmad}, \bibinfo{person}{Ilge Akkaya}, \bibinfo{person}{Florencia~Leoni Aleman}, \bibinfo{person}{Diogo Almeida}, \bibinfo{person}{Janko Altenschmidt}, \bibinfo{person}{Sam Altman}, \bibinfo{person}{Shyamal Anadkat}, {et~al\mbox{.}}} \bibinfo{year}{2023}\natexlab{}.
\newblock \showarticletitle{Gpt-4 technical report}.
\newblock \bibinfo{journal}{\emph{arXiv preprint arXiv:2303.08774}} (\bibinfo{year}{2023}).
\newblock


\bibitem[Alexanderson et~al\mbox{.}(2023)]%
        {alexanderson2023listen}
\bibfield{author}{\bibinfo{person}{Simon Alexanderson}, \bibinfo{person}{Rajmund Nagy}, \bibinfo{person}{Jonas Beskow}, {and} \bibinfo{person}{Gustav~Eje Henter}.} \bibinfo{year}{2023}\natexlab{}.
\newblock \showarticletitle{Listen, denoise, action! audio-driven motion synthesis with diffusion models}.
\newblock \bibinfo{journal}{\emph{ACM Transactions on Graphics (TOG)}} \bibinfo{volume}{42}, \bibinfo{number}{4} (\bibinfo{year}{2023}), \bibinfo{pages}{1--20}.
\newblock


\bibitem[Araujo et~al\mbox{.}(2025)]%
        {araujo2025retargeting}
\bibfield{author}{\bibinfo{person}{Joao~Pedro Araujo}, \bibinfo{person}{Yanjie Ze}, \bibinfo{person}{Pei Xu}, \bibinfo{person}{Jiajun Wu}, {and} \bibinfo{person}{C~Karen Liu}.} \bibinfo{year}{2025}\natexlab{}.
\newblock \showarticletitle{Retargeting matters: General motion retargeting for humanoid motion tracking}.
\newblock \bibinfo{journal}{\emph{arXiv preprint arXiv:2510.02252}} (\bibinfo{year}{2025}).
\newblock


\bibitem[Arikan and Forsyth(2002)]%
        {arikan2002interactive}
\bibfield{author}{\bibinfo{person}{Okan Arikan} {and} \bibinfo{person}{David~A Forsyth}.} \bibinfo{year}{2002}\natexlab{}.
\newblock \showarticletitle{Interactive motion generation from examples}.
\newblock \bibinfo{journal}{\emph{ACM Transactions on Graphics (TOG)}} \bibinfo{volume}{21}, \bibinfo{number}{3} (\bibinfo{year}{2002}), \bibinfo{pages}{483--490}.
\newblock


\bibitem[Bereznyak(2024)]%
        {bereznyak2024shadow}
\bibfield{author}{\bibinfo{person}{Alex Bereznyak}.} \bibinfo{year}{2024}\natexlab{}.
\newblock \showarticletitle{Shadow of HyperPose: New Animation System}.
\newblock In \bibinfo{booktitle}{\emph{ACM SIGGRAPH 2024 Talks}}. \bibinfo{pages}{1--2}.
\newblock


\bibitem[{Bones Studio}(2026)]%
        {bones_seed}
\bibfield{author}{\bibinfo{person}{{Bones Studio}}.} \bibinfo{year}{2026}\natexlab{}.
\newblock \bibinfo{title}{{BONES-SEED}: {S}keletal {E}veryday {E}mbodied {D}ataset}.
\newblock \bibinfo{howpublished}{\url{https://bones.studio/datasets}}.
\newblock
\newblock
\shownote{Open-source motion dataset, approximately 140k motion clips.}.


\bibitem[Buttner(2015)]%
        {buttner2015motion}
\bibfield{author}{\bibinfo{person}{Michael Buttner}.} \bibinfo{year}{2015}\natexlab{}.
\newblock \showarticletitle{Motion Matching-The Road to Next-Gen Animation}. In \bibinfo{booktitle}{\emph{Nucl. ai Conference}}.
\newblock


\bibitem[Buttner(2019)]%
        {buttner2019}
\bibfield{author}{\bibinfo{person}{Michael Buttner}.} \bibinfo{year}{2019}\natexlab{}.
\newblock \showarticletitle{Machine Learning for Motion Synthesis and Character Control}. In \bibinfo{booktitle}{\emph{Interactive 3D Graphics and Games (I3D) 2019}}.
\newblock
\urldef\tempurl%
\url{https://www.youtube.com/watch?v=zuvmQxcCOM4}
\showURL{%
\tempurl}


\bibitem[Chen et~al\mbox{.}(2024)]%
        {chen2024taming}
\bibfield{author}{\bibinfo{person}{Rui Chen}, \bibinfo{person}{Mingyi Shi}, \bibinfo{person}{Shaoli Huang}, \bibinfo{person}{Ping Tan}, \bibinfo{person}{Taku Komura}, {and} \bibinfo{person}{Xuelin Chen}.} \bibinfo{year}{2024}\natexlab{}.
\newblock \showarticletitle{Taming diffusion probabilistic models for character control}. In \bibinfo{booktitle}{\emph{ACM SIGGRAPH 2024 Conference Papers}}. \bibinfo{pages}{1--10}.
\newblock


\bibitem[Chen et~al\mbox{.}(2025b)]%
        {chen2025hand}
\bibfield{author}{\bibinfo{person}{Sirui Chen}, \bibinfo{person}{Yufei Ye}, \bibinfo{person}{Zi-ang Cao}, \bibinfo{person}{Jennifer Lew}, \bibinfo{person}{Pei Xu}, {and} \bibinfo{person}{C~Karen Liu}.} \bibinfo{year}{2025}\natexlab{b}.
\newblock \showarticletitle{Hand-eye autonomous delivery: Learning humanoid navigation, locomotion and reaching}.
\newblock \bibinfo{journal}{\emph{arXiv preprint arXiv:2508.03068}} (\bibinfo{year}{2025}).
\newblock


\bibitem[Chen et~al\mbox{.}(2025a)]%
        {chen2025gmt}
\bibfield{author}{\bibinfo{person}{Zixuan Chen}, \bibinfo{person}{Mazeyu Ji}, \bibinfo{person}{Xuxin Cheng}, \bibinfo{person}{Xuanbin Peng}, \bibinfo{person}{Xue~Bin Peng}, {and} \bibinfo{person}{Xiaolong Wang}.} \bibinfo{year}{2025}\natexlab{a}.
\newblock \showarticletitle{GMT: General Motion Tracking for Humanoid Whole-Body Control}.
\newblock \bibinfo{journal}{\emph{arXiv preprint arXiv:2506.14770}} (\bibinfo{year}{2025}).
\newblock


\bibitem[Clavet(2016)]%
        {clavet2016motion}
\bibfield{author}{\bibinfo{person}{Simon Clavet}.} \bibinfo{year}{2016}\natexlab{}.
\newblock \showarticletitle{Motion matching and the road to next-gen animation}.
\newblock \bibinfo{journal}{\emph{Proc. of GDC 2016}} (\bibinfo{year}{2016}).
\newblock


\bibitem[Cohan et~al\mbox{.}(2024)]%
        {cohan2024flexible}
\bibfield{author}{\bibinfo{person}{Setareh Cohan}, \bibinfo{person}{Guy Tevet}, \bibinfo{person}{Daniele Reda}, \bibinfo{person}{Xue~Bin Peng}, {and} \bibinfo{person}{Michiel van~de Panne}.} \bibinfo{year}{2024}\natexlab{}.
\newblock \showarticletitle{Flexible motion in-betweening with diffusion models}. In \bibinfo{booktitle}{\emph{ACM SIGGRAPH 2024 Conference Papers}}. \bibinfo{pages}{1--9}.
\newblock


\bibitem[Dynamics(2025)]%
        {boston_dynamics_choreographer}
\bibfield{author}{\bibinfo{person}{Boston Dynamics}.} \bibinfo{year}{2025}\natexlab{}.
\newblock \bibinfo{title}{Use Choreographer with Spot}.
\newblock \bibinfo{howpublished}{\url{https://support.bostondynamics.com/s/article/Use-Choreographer-with-Spot-72036}}.
\newblock


\bibitem[Engine(2025)]%
        {animgraph}
\bibfield{author}{\bibinfo{person}{Unreal Engine}.} \bibinfo{year}{2025}\natexlab{}.
\newblock \bibinfo{title}{Graphing in Animation Blueprints}.
\newblock \bibinfo{howpublished}{\url{https://dev.epicgames.com/documentation/en-us/unreal-engine/graphing-in-animation-blueprints-in-unreal-engine}}.
\newblock


\bibitem[{Epic Games}(2025)]%
        {epicgames2025blendspaces}
\bibfield{author}{\bibinfo{person}{{Epic Games}}.} \bibinfo{year}{2025}\natexlab{}.
\newblock \bibinfo{title}{Blend Spaces in {Unreal Engine}}.
\newblock \bibinfo{howpublished}{\url{https://dev.epicgames.com/documentation/en-us/unreal-engine/blend-spaces-in-unreal-engine}}.
\newblock
\newblock
\shownote{Accessed: 2026-01-22}.


\bibitem[Esser et~al\mbox{.}(2024)]%
        {esser2024scaling}
\bibfield{author}{\bibinfo{person}{Patrick Esser}, \bibinfo{person}{Sumith Kulal}, \bibinfo{person}{Andreas Blattmann}, \bibinfo{person}{Rahim Entezari}, \bibinfo{person}{Jonas M{\"u}ller}, \bibinfo{person}{Harry Saini}, \bibinfo{person}{Yam Levi}, \bibinfo{person}{Dominik Lorenz}, \bibinfo{person}{Axel Sauer}, \bibinfo{person}{Frederic Boesel}, {et~al\mbox{.}}} \bibinfo{year}{2024}\natexlab{}.
\newblock \showarticletitle{Scaling rectified flow transformers for high-resolution image synthesis}. In \bibinfo{booktitle}{\emph{Forty-first international conference on machine learning}}.
\newblock


\bibitem[Esser et~al\mbox{.}(2021)]%
        {esser2021taming}
\bibfield{author}{\bibinfo{person}{Patrick Esser}, \bibinfo{person}{Robin Rombach}, {and} \bibinfo{person}{Bjorn Ommer}.} \bibinfo{year}{2021}\natexlab{}.
\newblock \showarticletitle{Taming transformers for high-resolution image synthesis}. In \bibinfo{booktitle}{\emph{Proceedings of the IEEE/CVF conference on computer vision and pattern recognition}}. \bibinfo{pages}{12873--12883}.
\newblock


\bibitem[Fragkiadaki et~al\mbox{.}(2015)]%
        {fragkiadaki2015recurrent}
\bibfield{author}{\bibinfo{person}{Katerina Fragkiadaki}, \bibinfo{person}{Sergey Levine}, \bibinfo{person}{Panna Felsen}, {and} \bibinfo{person}{Jitendra Malik}.} \bibinfo{year}{2015}\natexlab{}.
\newblock \showarticletitle{Recurrent network models for human dynamics}. In \bibinfo{booktitle}{\emph{Proceedings of the IEEE international conference on computer vision}}. \bibinfo{pages}{4346--4354}.
\newblock


\bibitem[Games(2025)]%
        {epicgames2025ikriganimationretargetinginunrealengine}
\bibfield{author}{\bibinfo{person}{Epic Games}.} \bibinfo{year}{2025}\natexlab{}.
\newblock \bibinfo{title}{IK Rig Animation Retargeting in Unreal Engine}.
\newblock \bibinfo{howpublished}{\url{https://dev.epicgames.com/documentation/en-us/unreal-engine/ik-rig-animation-retargeting-in-unreal-engine}}.
\newblock


\bibitem[{Google}(2025)]%
        {google_veo3}
\bibfield{author}{\bibinfo{person}{{Google}}.} \bibinfo{year}{2025}\natexlab{}.
\newblock \bibinfo{title}{Veo 3}.
\newblock
\urldef\tempurl%
\url{https://deepmind.google/technologies/veo/}
\showURL{%
\tempurl}


\bibitem[Gopalakrishnan et~al\mbox{.}(2019)]%
        {gopalakrishnan2019neural}
\bibfield{author}{\bibinfo{person}{Anand Gopalakrishnan}, \bibinfo{person}{Ankur Mali}, \bibinfo{person}{Dan Kifer}, \bibinfo{person}{Lee Giles}, {and} \bibinfo{person}{Alexander~G Ororbia}.} \bibinfo{year}{2019}\natexlab{}.
\newblock \showarticletitle{A neural temporal model for human motion prediction}. In \bibinfo{booktitle}{\emph{Proceedings of the IEEE/CVF Conference on Computer Vision and Pattern Recognition}}. \bibinfo{pages}{12116--12125}.
\newblock


\bibitem[Gou et~al\mbox{.}(2025)]%
        {gou2025control}
\bibfield{author}{\bibinfo{person}{Ruiyu Gou}, \bibinfo{person}{Michiel van~de Panne}, {and} \bibinfo{person}{Daniel Holden}.} \bibinfo{year}{2025}\natexlab{}.
\newblock \showarticletitle{Control Operators for Interactive Character Animation}.
\newblock \bibinfo{journal}{\emph{ACM Transactions on Graphics (TOG)}} \bibinfo{volume}{44}, \bibinfo{number}{6} (\bibinfo{year}{2025}), \bibinfo{pages}{1--20}.
\newblock


\bibitem[Guo et~al\mbox{.}(2024)]%
        {guo2024momask}
\bibfield{author}{\bibinfo{person}{Chuan Guo}, \bibinfo{person}{Yuxuan Mu}, \bibinfo{person}{Muhammad~Gohar Javed}, \bibinfo{person}{Sen Wang}, {and} \bibinfo{person}{Li Cheng}.} \bibinfo{year}{2024}\natexlab{}.
\newblock \showarticletitle{Momask: Generative masked modeling of 3d human motions}. In \bibinfo{booktitle}{\emph{Proceedings of the IEEE/CVF Conference on Computer Vision and Pattern Recognition}}. \bibinfo{pages}{1900--1910}.
\newblock


\bibitem[Guo et~al\mbox{.}(2022a)]%
        {guo2022generating}
\bibfield{author}{\bibinfo{person}{Chuan Guo}, \bibinfo{person}{Shihao Zou}, \bibinfo{person}{Xinxin Zuo}, \bibinfo{person}{Sen Wang}, \bibinfo{person}{Wei Ji}, \bibinfo{person}{Xingyu Li}, {and} \bibinfo{person}{Li Cheng}.} \bibinfo{year}{2022}\natexlab{a}.
\newblock \showarticletitle{Generating diverse and natural 3d human motions from text}. In \bibinfo{booktitle}{\emph{Proceedings of the IEEE/CVF conference on computer vision and pattern recognition}}. \bibinfo{pages}{5152--5161}.
\newblock


\bibitem[Guo et~al\mbox{.}(2022b)]%
        {guo2022tm2t}
\bibfield{author}{\bibinfo{person}{Chuan Guo}, \bibinfo{person}{Xinxin Zuo}, \bibinfo{person}{Sen Wang}, {and} \bibinfo{person}{Li Cheng}.} \bibinfo{year}{2022}\natexlab{b}.
\newblock \showarticletitle{Tm2t: Stochastic and tokenized modeling for the reciprocal generation of 3d human motions and texts}. In \bibinfo{booktitle}{\emph{European Conference on Computer Vision}}. Springer, \bibinfo{pages}{580--597}.
\newblock


\bibitem[Han et~al\mbox{.}(2024)]%
        {han2024amd}
\bibfield{author}{\bibinfo{person}{Bo Han}, \bibinfo{person}{Hao Peng}, \bibinfo{person}{Minjing Dong}, \bibinfo{person}{Yi Ren}, \bibinfo{person}{Yixuan Shen}, {and} \bibinfo{person}{Chang Xu}.} \bibinfo{year}{2024}\natexlab{}.
\newblock \showarticletitle{Amd: Autoregressive motion diffusion}. In \bibinfo{booktitle}{\emph{Proceedings of the AAAI Conference on Artificial Intelligence}}, Vol.~\bibinfo{volume}{38}. \bibinfo{pages}{2022--2030}.
\newblock


\bibitem[Harvey et~al\mbox{.}(2020)]%
        {harvey2020robust}
\bibfield{author}{\bibinfo{person}{F{\'e}lix~G Harvey}, \bibinfo{person}{Mike Yurick}, \bibinfo{person}{Derek Nowrouzezahrai}, {and} \bibinfo{person}{Christopher Pal}.} \bibinfo{year}{2020}\natexlab{}.
\newblock \showarticletitle{Robust motion in-betweening}.
\newblock \bibinfo{journal}{\emph{ACM Transactions on Graphics (TOG)}} \bibinfo{volume}{39}, \bibinfo{number}{4} (\bibinfo{year}{2020}), \bibinfo{pages}{60--1}.
\newblock


\bibitem[He et~al\mbox{.}(2025)]%
        {he2025viral}
\bibfield{author}{\bibinfo{person}{Tairan He}, \bibinfo{person}{Zi Wang}, \bibinfo{person}{Haoru Xue}, \bibinfo{person}{Qingwei Ben}, \bibinfo{person}{Zhengyi Luo}, \bibinfo{person}{Wenli Xiao}, \bibinfo{person}{Ye Yuan}, \bibinfo{person}{Xingye Da}, \bibinfo{person}{Fernando Casta{\~n}eda}, \bibinfo{person}{Shankar Sastry}, {et~al\mbox{.}}} \bibinfo{year}{2025}\natexlab{}.
\newblock \showarticletitle{VIRAL: Visual Sim-to-Real at Scale for Humanoid Loco-Manipulation}.
\newblock \bibinfo{journal}{\emph{arXiv preprint arXiv:2511.15200}} (\bibinfo{year}{2025}).
\newblock


\bibitem[Heusel et~al\mbox{.}(2017)]%
        {heusel2017gans}
\bibfield{author}{\bibinfo{person}{Martin Heusel}, \bibinfo{person}{Hubert Ramsauer}, \bibinfo{person}{Thomas Unterthiner}, \bibinfo{person}{Bernhard Nessler}, {and} \bibinfo{person}{Sepp Hochreiter}.} \bibinfo{year}{2017}\natexlab{}.
\newblock \showarticletitle{Gans trained by a two time-scale update rule converge to a local nash equilibrium}.
\newblock \bibinfo{journal}{\emph{Advances in neural information processing systems}}  \bibinfo{volume}{30} (\bibinfo{year}{2017}).
\newblock


\bibitem[Ho and Ermon(2016)]%
        {ho2016generative}
\bibfield{author}{\bibinfo{person}{Jonathan Ho} {and} \bibinfo{person}{Stefano Ermon}.} \bibinfo{year}{2016}\natexlab{}.
\newblock \showarticletitle{Generative adversarial imitation learning}. In \bibinfo{booktitle}{\emph{Advances in Neural Information Processing Systems}}. \bibinfo{pages}{4565--4573}.
\newblock


\bibitem[Ho et~al\mbox{.}(2020)]%
        {ho2020denoising}
\bibfield{author}{\bibinfo{person}{Jonathan Ho}, \bibinfo{person}{Ajay Jain}, {and} \bibinfo{person}{Pieter Abbeel}.} \bibinfo{year}{2020}\natexlab{}.
\newblock \showarticletitle{Denoising diffusion probabilistic models}.
\newblock \bibinfo{journal}{\emph{Advances in neural information processing systems}}  \bibinfo{volume}{33} (\bibinfo{year}{2020}), \bibinfo{pages}{6840--6851}.
\newblock


\bibitem[Ho and Salimans(2022)]%
        {ho2022classifier}
\bibfield{author}{\bibinfo{person}{Jonathan Ho} {and} \bibinfo{person}{Tim Salimans}.} \bibinfo{year}{2022}\natexlab{}.
\newblock \showarticletitle{Classifier-free diffusion guidance}.
\newblock \bibinfo{journal}{\emph{arXiv preprint arXiv:2207.12598}} (\bibinfo{year}{2022}).
\newblock


\bibitem[Holden(2018)]%
        {holden2018character}
\bibfield{author}{\bibinfo{person}{Daniel Holden}.} \bibinfo{year}{2018}\natexlab{}.
\newblock \showarticletitle{Character control with neural networks and machine learning}. In \bibinfo{booktitle}{\emph{Proc. of GDC}}, Vol.~\bibinfo{volume}{1}. \bibinfo{pages}{2}.
\newblock


\bibitem[Holden et~al\mbox{.}(2020)]%
        {holden2020learned}
\bibfield{author}{\bibinfo{person}{Daniel Holden}, \bibinfo{person}{Oussama Kanoun}, \bibinfo{person}{Maksym Perepichka}, {and} \bibinfo{person}{Tiberiu Popa}.} \bibinfo{year}{2020}\natexlab{}.
\newblock \showarticletitle{Learned motion matching}.
\newblock \bibinfo{journal}{\emph{ACM Transactions on Graphics (TOG)}} \bibinfo{volume}{39}, \bibinfo{number}{4} (\bibinfo{year}{2020}), \bibinfo{pages}{53--1}.
\newblock


\bibitem[Holden et~al\mbox{.}(2017)]%
        {holden2017phase}
\bibfield{author}{\bibinfo{person}{Daniel Holden}, \bibinfo{person}{Taku Komura}, {and} \bibinfo{person}{Jun Saito}.} \bibinfo{year}{2017}\natexlab{}.
\newblock \showarticletitle{Phase-functioned neural networks for character control}.
\newblock \bibinfo{journal}{\emph{ACM Transactions on Graphics (TOG)}} \bibinfo{volume}{36}, \bibinfo{number}{4} (\bibinfo{year}{2017}), \bibinfo{pages}{42}.
\newblock


\bibitem[Holden et~al\mbox{.}(2016)]%
        {holden2016deep}
\bibfield{author}{\bibinfo{person}{Daniel Holden}, \bibinfo{person}{Jun Saito}, {and} \bibinfo{person}{Taku Komura}.} \bibinfo{year}{2016}\natexlab{}.
\newblock \showarticletitle{A deep learning framework for character motion synthesis and editing}.
\newblock \bibinfo{journal}{\emph{ACM Transactions on Graphics (ToG)}} \bibinfo{volume}{35}, \bibinfo{number}{4} (\bibinfo{year}{2016}), \bibinfo{pages}{1--11}.
\newblock


\bibitem[Jayasumana et~al\mbox{.}(2024)]%
        {jayasumana2024rethinking}
\bibfield{author}{\bibinfo{person}{Sadeep Jayasumana}, \bibinfo{person}{Srikumar Ramalingam}, \bibinfo{person}{Andreas Veit}, \bibinfo{person}{Daniel Glasner}, \bibinfo{person}{Ayan Chakrabarti}, {and} \bibinfo{person}{Sanjiv Kumar}.} \bibinfo{year}{2024}\natexlab{}.
\newblock \showarticletitle{Rethinking fid: Towards a better evaluation metric for image generation}. In \bibinfo{booktitle}{\emph{Proceedings of the IEEE/CVF Conference on Computer Vision and Pattern Recognition}}. \bibinfo{pages}{9307--9315}.
\newblock


\bibitem[Jiang et~al\mbox{.}(2023)]%
        {jiang2023motiongpt}
\bibfield{author}{\bibinfo{person}{Biao Jiang}, \bibinfo{person}{Xin Chen}, \bibinfo{person}{Wen Liu}, \bibinfo{person}{Jingyi Yu}, \bibinfo{person}{Gang Yu}, {and} \bibinfo{person}{Tao Chen}.} \bibinfo{year}{2023}\natexlab{}.
\newblock \showarticletitle{Motiongpt: Human motion as a foreign language}.
\newblock \bibinfo{journal}{\emph{Advances in Neural Information Processing Systems}}  \bibinfo{volume}{36} (\bibinfo{year}{2023}), \bibinfo{pages}{20067--20079}.
\newblock


\bibitem[Kim et~al\mbox{.}(2025)]%
        {kim2025pyroki}
\bibfield{author}{\bibinfo{person}{Chung~Min Kim}, \bibinfo{person}{Brent Yi}, \bibinfo{person}{Hongsuk Choi}, \bibinfo{person}{Yi Ma}, \bibinfo{person}{Ken Goldberg}, {and} \bibinfo{person}{Angjoo Kanazawa}.} \bibinfo{year}{2025}\natexlab{}.
\newblock \showarticletitle{PyRoki: A Modular Toolkit for Robot Kinematic Optimization}.
\newblock \bibinfo{journal}{\emph{arXiv preprint arXiv:2505.03728}} (\bibinfo{year}{2025}).
\newblock


\bibitem[Kim et~al\mbox{.}(2022)]%
        {kim2022conditional}
\bibfield{author}{\bibinfo{person}{Jihoon Kim}, \bibinfo{person}{Taehyun Byun}, \bibinfo{person}{Seungyoun Shin}, \bibinfo{person}{Jungdam Won}, {and} \bibinfo{person}{Sungjoon Choi}.} \bibinfo{year}{2022}\natexlab{}.
\newblock \showarticletitle{Conditional motion in-betweening}.
\newblock \bibinfo{journal}{\emph{Pattern Recognition}}  \bibinfo{volume}{132} (\bibinfo{year}{2022}), \bibinfo{pages}{108894}.
\newblock


\bibitem[Kiranyaz et~al\mbox{.}(2021)]%
        {kiranyaz20211d}
\bibfield{author}{\bibinfo{person}{Serkan Kiranyaz}, \bibinfo{person}{Onur Avci}, \bibinfo{person}{Osama Abdeljaber}, \bibinfo{person}{Turker Ince}, \bibinfo{person}{Moncef Gabbouj}, {and} \bibinfo{person}{Daniel~J Inman}.} \bibinfo{year}{2021}\natexlab{}.
\newblock \showarticletitle{1D convolutional neural networks and applications: A survey}.
\newblock \bibinfo{journal}{\emph{Mechanical systems and signal processing}}  \bibinfo{volume}{151} (\bibinfo{year}{2021}), \bibinfo{pages}{107398}.
\newblock


\bibitem[Kovar(2002)]%
        {kovar2002motion}
\bibfield{author}{\bibinfo{person}{L Kovar}.} \bibinfo{year}{2002}\natexlab{}.
\newblock \showarticletitle{Motion graphs}.
\newblock \bibinfo{journal}{\emph{ACM Trans. Graph.}} \bibinfo{volume}{21}, \bibinfo{number}{3} (\bibinfo{year}{2002}), \bibinfo{pages}{473--482}.
\newblock


\bibitem[Kovar and Gleicher(2004)]%
        {kovar2004automated}
\bibfield{author}{\bibinfo{person}{Lucas Kovar} {and} \bibinfo{person}{Michael Gleicher}.} \bibinfo{year}{2004}\natexlab{}.
\newblock \showarticletitle{Automated extraction and parameterization of motions in large data sets}.
\newblock \bibinfo{journal}{\emph{ACM Transactions on Graphics (ToG)}} \bibinfo{volume}{23}, \bibinfo{number}{3} (\bibinfo{year}{2004}), \bibinfo{pages}{559--568}.
\newblock


\bibitem[Lee et~al\mbox{.}(2002)]%
        {lee2002interactive}
\bibfield{author}{\bibinfo{person}{Jehee Lee}, \bibinfo{person}{Jinxiang Chai}, \bibinfo{person}{Paul~SA Reitsma}, \bibinfo{person}{Jessica~K Hodgins}, {and} \bibinfo{person}{Nancy~S Pollard}.} \bibinfo{year}{2002}\natexlab{}.
\newblock \showarticletitle{Interactive control of avatars animated with human motion data}. In \bibinfo{booktitle}{\emph{Proceedings of the 29th annual conference on Computer graphics and interactive techniques}}. \bibinfo{pages}{491--500}.
\newblock


\bibitem[Lee et~al\mbox{.}(2010)]%
        {motionfield2010}
\bibfield{author}{\bibinfo{person}{Yongjoon Lee}, \bibinfo{person}{Kevin Wampler}, \bibinfo{person}{Gilbert Bernstein}, \bibinfo{person}{Jovan Popovi\'{c}}, {and} \bibinfo{person}{Zoran Popovi\'{c}}.} \bibinfo{year}{2010}\natexlab{}.
\newblock \showarticletitle{Motion Fields for Interactive Character Locomotion}. In \bibinfo{booktitle}{\emph{ACM SIGGRAPH Asia 2010 Papers}} (Seoul, South Korea) \emph{(\bibinfo{series}{SIGGRAPH ASIA '10})}. \bibinfo{publisher}{Association for Computing Machinery}, \bibinfo{address}{New York, NY, USA}, Article \bibinfo{articleno}{138}, \bibinfo{numpages}{8}~pages.
\newblock
\showISBNx{9781450304399}
\href{https://doi.org/10.1145/1866158.1866160}{doi:\nolinkurl{10.1145/1866158.1866160}}


\bibitem[Li et~al\mbox{.}(2025)]%
        {li2025genmo}
\bibfield{author}{\bibinfo{person}{Jiefeng Li}, \bibinfo{person}{Jinkun Cao}, \bibinfo{person}{Haotian Zhang}, \bibinfo{person}{Davis Rempe}, \bibinfo{person}{Jan Kautz}, \bibinfo{person}{Umar Iqbal}, {and} \bibinfo{person}{Ye Yuan}.} \bibinfo{year}{2025}\natexlab{}.
\newblock \showarticletitle{GENMO: A GENeralist Model for Human MOtion}.
\newblock \bibinfo{journal}{\emph{arXiv preprint arXiv:2505.01425}} (\bibinfo{year}{2025}).
\newblock


\bibitem[Li et~al\mbox{.}(2024a)]%
        {li2024controllable}
\bibfield{author}{\bibinfo{person}{Jiaman Li}, \bibinfo{person}{Alexander Clegg}, \bibinfo{person}{Roozbeh Mottaghi}, \bibinfo{person}{Jiajun Wu}, \bibinfo{person}{Xavier Puig}, {and} \bibinfo{person}{C~Karen Liu}.} \bibinfo{year}{2024}\natexlab{a}.
\newblock \showarticletitle{Controllable human-object interaction synthesis}. In \bibinfo{booktitle}{\emph{European Conference on Computer Vision}}. Springer, \bibinfo{pages}{54--72}.
\newblock


\bibitem[Li et~al\mbox{.}(2023)]%
        {li2023object}
\bibfield{author}{\bibinfo{person}{Jiaman Li}, \bibinfo{person}{Jiajun Wu}, {and} \bibinfo{person}{C~Karen Liu}.} \bibinfo{year}{2023}\natexlab{}.
\newblock \showarticletitle{Object motion guided human motion synthesis}.
\newblock \bibinfo{journal}{\emph{ACM Transactions on Graphics (TOG)}} \bibinfo{volume}{42}, \bibinfo{number}{6} (\bibinfo{year}{2023}), \bibinfo{pages}{1--11}.
\newblock


\bibitem[Li et~al\mbox{.}(2024b)]%
        {li2024aamdm}
\bibfield{author}{\bibinfo{person}{Tianyu Li}, \bibinfo{person}{Calvin Qiao}, \bibinfo{person}{Guanqiao Ren}, \bibinfo{person}{KangKang Yin}, {and} \bibinfo{person}{Sehoon Ha}.} \bibinfo{year}{2024}\natexlab{b}.
\newblock \showarticletitle{AAMDM: accelerated auto-regressive motion diffusion model}. In \bibinfo{booktitle}{\emph{Proceedings of the IEEE/CVF Conference on Computer Vision and Pattern Recognition}}. \bibinfo{pages}{1813--1823}.
\newblock


\bibitem[Lipman et~al\mbox{.}(2022)]%
        {lipman2022flow}
\bibfield{author}{\bibinfo{person}{Yaron Lipman}, \bibinfo{person}{Ricky~TQ Chen}, \bibinfo{person}{Heli Ben-Hamu}, \bibinfo{person}{Maximilian Nickel}, {and} \bibinfo{person}{Matt Le}.} \bibinfo{year}{2022}\natexlab{}.
\newblock \showarticletitle{Flow matching for generative modeling}.
\newblock \bibinfo{journal}{\emph{arXiv preprint arXiv:2210.02747}} (\bibinfo{year}{2022}).
\newblock


\bibitem[Lu et~al\mbox{.}(2024)]%
        {lu2024choice}
\bibfield{author}{\bibinfo{person}{Jintao Lu}, \bibinfo{person}{He Zhang}, \bibinfo{person}{Yuting Ye}, \bibinfo{person}{Takaaki Shiratori}, \bibinfo{person}{Sebastian Starke}, {and} \bibinfo{person}{Taku Komura}.} \bibinfo{year}{2024}\natexlab{}.
\newblock \showarticletitle{CHOICE: Coordinated human-object interaction in cluttered environments for pick-and-place actions}.
\newblock \bibinfo{journal}{\emph{arXiv preprint arXiv:2412.06702}} (\bibinfo{year}{2024}).
\newblock


\bibitem[Luo et~al\mbox{.}(2024)]%
        {luo2024open}
\bibfield{author}{\bibinfo{person}{Zhuoyan Luo}, \bibinfo{person}{Fengyuan Shi}, \bibinfo{person}{Yixiao Ge}, \bibinfo{person}{Yujiu Yang}, \bibinfo{person}{Limin Wang}, {and} \bibinfo{person}{Ying Shan}.} \bibinfo{year}{2024}\natexlab{}.
\newblock \showarticletitle{Open-magvit2: An open-source project toward democratizing auto-regressive visual generation}.
\newblock \bibinfo{journal}{\emph{arXiv preprint arXiv:2409.04410}} (\bibinfo{year}{2024}).
\newblock


\bibitem[Luo et~al\mbox{.}(2025)]%
        {luo2025sonic}
\bibfield{author}{\bibinfo{person}{Zhengyi Luo}, \bibinfo{person}{Ye Yuan}, \bibinfo{person}{Tingwu Wang}, \bibinfo{person}{Chenran Li}, \bibinfo{person}{Sirui Chen}, \bibinfo{person}{Fernando Casta{\~n}eda}, \bibinfo{person}{Zi-Ang Cao}, \bibinfo{person}{Jiefeng Li}, \bibinfo{person}{David Minor}, \bibinfo{person}{Qingwei Ben}, {et~al\mbox{.}}} \bibinfo{year}{2025}\natexlab{}.
\newblock \showarticletitle{Sonic: Supersizing motion tracking for natural humanoid whole-body control}.
\newblock \bibinfo{journal}{\emph{arXiv preprint arXiv:2511.07820}} (\bibinfo{year}{2025}).
\newblock


\bibitem[Mentzer et~al\mbox{.}(2023)]%
        {mentzer2023finite}
\bibfield{author}{\bibinfo{person}{Fabian Mentzer}, \bibinfo{person}{David Minnen}, \bibinfo{person}{Eirikur Agustsson}, {and} \bibinfo{person}{Michael Tschannen}.} \bibinfo{year}{2023}\natexlab{}.
\newblock \showarticletitle{Finite scalar quantization: Vq-vae made simple}.
\newblock \bibinfo{journal}{\emph{arXiv preprint arXiv:2309.15505}} (\bibinfo{year}{2023}).
\newblock


\bibitem[Min and Chai(2012)]%
        {min2012motion}
\bibfield{author}{\bibinfo{person}{Jianyuan Min} {and} \bibinfo{person}{Jinxiang Chai}.} \bibinfo{year}{2012}\natexlab{}.
\newblock \showarticletitle{Motion graphs++ a compact generative model for semantic motion analysis and synthesis}.
\newblock \bibinfo{journal}{\emph{ACM Transactions on Graphics (TOG)}} \bibinfo{volume}{31}, \bibinfo{number}{6} (\bibinfo{year}{2012}), \bibinfo{pages}{1--12}.
\newblock


\bibitem[Ni et~al\mbox{.}(2025)]%
        {ni2025generated}
\bibfield{author}{\bibinfo{person}{James Ni}, \bibinfo{person}{Zekai Wang}, \bibinfo{person}{Wei Lin}, \bibinfo{person}{Amir Bar}, \bibinfo{person}{Yann LeCun}, \bibinfo{person}{Trevor Darrell}, \bibinfo{person}{Jitendra Malik}, {and} \bibinfo{person}{Roei Herzig}.} \bibinfo{year}{2025}\natexlab{}.
\newblock \showarticletitle{From Generated Human Videos to Physically Plausible Robot Trajectories}.
\newblock \bibinfo{journal}{\emph{arXiv preprint arXiv:2512.05094}} (\bibinfo{year}{2025}).
\newblock


\bibitem[{OpenAI}(2024)]%
        {openai_sora}
\bibfield{author}{\bibinfo{person}{{OpenAI}}.} \bibinfo{year}{2024}\natexlab{}.
\newblock \bibinfo{title}{Sora: Creating Video From Text}.
\newblock
\urldef\tempurl%
\url{https://openai.com/sora}
\showURL{%
\tempurl}


\bibitem[Oreshkin et~al\mbox{.}(2024)]%
        {oreshkin2023motion}
\bibfield{author}{\bibinfo{person}{Boris~N Oreshkin}, \bibinfo{person}{Antonios Valkanas}, \bibinfo{person}{F{\'e}lix~G Harvey}, \bibinfo{person}{Louis-Simon M{\'e}nard}, \bibinfo{person}{Florent Bocquelet}, {and} \bibinfo{person}{Mark~J Coates}.} \bibinfo{year}{2024}\natexlab{}.
\newblock \showarticletitle{Motion In-Betweening via Deep Delta-Interpolator}.
\newblock \bibinfo{journal}{\emph{IEEE Transactions on Visualization and Computer Graphics}} \bibinfo{volume}{30}, \bibinfo{number}{8} (\bibinfo{year}{2024}), \bibinfo{pages}{5693--5704}.
\newblock


\bibitem[Peng et~al\mbox{.}(2018)]%
        {peng2018deepmimic}
\bibfield{author}{\bibinfo{person}{Xue~Bin Peng}, \bibinfo{person}{Pieter Abbeel}, \bibinfo{person}{Sergey Levine}, {and} \bibinfo{person}{Michiel van~de Panne}.} \bibinfo{year}{2018}\natexlab{}.
\newblock \showarticletitle{Deepmimic: Example-guided deep reinforcement learning of physics-based character skills}.
\newblock \bibinfo{journal}{\emph{ACM Transactions on Graphics (TOG)}} \bibinfo{volume}{37}, \bibinfo{number}{4} (\bibinfo{year}{2018}), \bibinfo{pages}{1--14}.
\newblock


\bibitem[Petrovich et~al\mbox{.}(2023)]%
        {petrovich2023tmr}
\bibfield{author}{\bibinfo{person}{Mathis Petrovich}, \bibinfo{person}{Michael~J Black}, {and} \bibinfo{person}{G{\"u}l Varol}.} \bibinfo{year}{2023}\natexlab{}.
\newblock \showarticletitle{Tmr: Text-to-motion retrieval using contrastive 3d human motion synthesis}. In \bibinfo{booktitle}{\emph{Proceedings of the IEEE/CVF International Conference on Computer Vision}}. \bibinfo{pages}{9488--9497}.
\newblock


\bibitem[Pinyoanuntapong et~al\mbox{.}(2024a)]%
        {pinyoanuntapong2024controlmm}
\bibfield{author}{\bibinfo{person}{Ekkasit Pinyoanuntapong}, \bibinfo{person}{Muhammad~Usama Saleem}, \bibinfo{person}{Korrawe Karunratanakul}, \bibinfo{person}{Pu Wang}, \bibinfo{person}{Hongfei Xue}, \bibinfo{person}{Chen Chen}, \bibinfo{person}{Chuan Guo}, \bibinfo{person}{Junli Cao}, \bibinfo{person}{Jian Ren}, {and} \bibinfo{person}{Sergey Tulyakov}.} \bibinfo{year}{2024}\natexlab{a}.
\newblock \showarticletitle{Controlmm: Controllable masked motion generation}.
\newblock \bibinfo{journal}{\emph{arXiv preprint arXiv:2410.10780}} (\bibinfo{year}{2024}).
\newblock


\bibitem[Pinyoanuntapong et~al\mbox{.}(2024b)]%
        {pinyoanuntapong2024mmm}
\bibfield{author}{\bibinfo{person}{Ekkasit Pinyoanuntapong}, \bibinfo{person}{Pu Wang}, \bibinfo{person}{Minwoo Lee}, {and} \bibinfo{person}{Chen Chen}.} \bibinfo{year}{2024}\natexlab{b}.
\newblock \showarticletitle{Mmm: Generative masked motion model}. In \bibinfo{booktitle}{\emph{Proceedings of the IEEE/CVF Conference on Computer Vision and Pattern Recognition}}. \bibinfo{pages}{1546--1555}.
\newblock


\bibitem[Qin et~al\mbox{.}(2022)]%
        {qin2022motion}
\bibfield{author}{\bibinfo{person}{Jia Qin}, \bibinfo{person}{Youyi Zheng}, {and} \bibinfo{person}{Kun Zhou}.} \bibinfo{year}{2022}\natexlab{}.
\newblock \showarticletitle{Motion In-Betweening via Two-Stage Transformers.}
\newblock \bibinfo{journal}{\emph{ACM Trans. Graph.}} \bibinfo{volume}{41}, \bibinfo{number}{6} (\bibinfo{year}{2022}), \bibinfo{pages}{184--1}.
\newblock


\bibitem[Razavi et~al\mbox{.}(2019)]%
        {razavi2019generating}
\bibfield{author}{\bibinfo{person}{Ali Razavi}, \bibinfo{person}{Aaron Van~den Oord}, {and} \bibinfo{person}{Oriol Vinyals}.} \bibinfo{year}{2019}\natexlab{}.
\newblock \showarticletitle{Generating diverse high-fidelity images with vq-vae-2}.
\newblock \bibinfo{journal}{\emph{Advances in neural information processing systems}}  \bibinfo{volume}{32} (\bibinfo{year}{2019}).
\newblock


\bibitem[Rombach et~al\mbox{.}(2022)]%
        {rombach2022high}
\bibfield{author}{\bibinfo{person}{Robin Rombach}, \bibinfo{person}{Andreas Blattmann}, \bibinfo{person}{Dominik Lorenz}, \bibinfo{person}{Patrick Esser}, {and} \bibinfo{person}{Bj{\"o}rn Ommer}.} \bibinfo{year}{2022}\natexlab{}.
\newblock \showarticletitle{High-resolution image synthesis with latent diffusion models}. In \bibinfo{booktitle}{\emph{Proceedings of the IEEE/CVF conference on computer vision and pattern recognition}}. \bibinfo{pages}{10684--10695}.
\newblock


\bibitem[Safonova and Hodgins(2007)]%
        {safonova2007construction}
\bibfield{author}{\bibinfo{person}{Alla Safonova} {and} \bibinfo{person}{Jessica~K Hodgins}.} \bibinfo{year}{2007}\natexlab{}.
\newblock \showarticletitle{Construction and optimal search of interpolated motion graphs}.
\newblock In \bibinfo{booktitle}{\emph{ACM SIGGRAPH 2007 papers}}. \bibinfo{pages}{106--es}.
\newblock


\bibitem[Shafir et~al\mbox{.}(2023)]%
        {shafir2023human}
\bibfield{author}{\bibinfo{person}{Yonatan Shafir}, \bibinfo{person}{Guy Tevet}, \bibinfo{person}{Roy Kapon}, {and} \bibinfo{person}{Amit~H Bermano}.} \bibinfo{year}{2023}\natexlab{}.
\newblock \showarticletitle{Human motion diffusion as a generative prior}.
\newblock \bibinfo{journal}{\emph{arXiv preprint arXiv:2303.01418}} (\bibinfo{year}{2023}).
\newblock


\bibitem[Shi et~al\mbox{.}(2024)]%
        {shi2024interactive}
\bibfield{author}{\bibinfo{person}{Yi Shi}, \bibinfo{person}{Jingbo Wang}, \bibinfo{person}{Xuekun Jiang}, \bibinfo{person}{Bingkun Lin}, \bibinfo{person}{Bo Dai}, {and} \bibinfo{person}{Xue~Bin Peng}.} \bibinfo{year}{2024}\natexlab{}.
\newblock \showarticletitle{Interactive character control with auto-regressive motion diffusion models}.
\newblock \bibinfo{journal}{\emph{ACM Transactions on Graphics (TOG)}} \bibinfo{volume}{43}, \bibinfo{number}{4} (\bibinfo{year}{2024}), \bibinfo{pages}{1--14}.
\newblock


\bibitem[Song et~al\mbox{.}(2020)]%
        {song2020denoising}
\bibfield{author}{\bibinfo{person}{Jiaming Song}, \bibinfo{person}{Chenlin Meng}, {and} \bibinfo{person}{Stefano Ermon}.} \bibinfo{year}{2020}\natexlab{}.
\newblock \showarticletitle{Denoising diffusion implicit models}.
\newblock \bibinfo{journal}{\emph{arXiv preprint arXiv:2010.02502}} (\bibinfo{year}{2020}).
\newblock


\bibitem[Starke et~al\mbox{.}(2023)]%
        {starke2023motion}
\bibfield{author}{\bibinfo{person}{Paul Starke}, \bibinfo{person}{Sebastian Starke}, \bibinfo{person}{Taku Komura}, {and} \bibinfo{person}{Frank Steinicke}.} \bibinfo{year}{2023}\natexlab{}.
\newblock \showarticletitle{Motion in-betweening with phase manifolds}.
\newblock \bibinfo{journal}{\emph{Proceedings of the ACM on Computer Graphics and Interactive Techniques}} \bibinfo{volume}{6}, \bibinfo{number}{3} (\bibinfo{year}{2023}), \bibinfo{pages}{1--17}.
\newblock


\bibitem[Starke et~al\mbox{.}(2024)]%
        {starke2024categorical}
\bibfield{author}{\bibinfo{person}{Sebastian Starke}, \bibinfo{person}{Paul Starke}, \bibinfo{person}{Nicky He}, \bibinfo{person}{Taku Komura}, {and} \bibinfo{person}{Yuting Ye}.} \bibinfo{year}{2024}\natexlab{}.
\newblock \showarticletitle{Categorical codebook matching for embodied character controllers}.
\newblock \bibinfo{journal}{\emph{ACM Transactions on Graphics (TOG)}} \bibinfo{volume}{43}, \bibinfo{number}{4} (\bibinfo{year}{2024}), \bibinfo{pages}{1--14}.
\newblock


\bibitem[Starke et~al\mbox{.}(2019)]%
        {starke2019neural}
\bibfield{author}{\bibinfo{person}{Sebastian Starke}, \bibinfo{person}{He Zhang}, \bibinfo{person}{Taku Komura}, {and} \bibinfo{person}{Jun Saito}.} \bibinfo{year}{2019}\natexlab{}.
\newblock \showarticletitle{Neural state machine for character-scene interactions}.
\newblock \bibinfo{journal}{\emph{ACM Transactions on Graphics (TOG)}} \bibinfo{volume}{38}, \bibinfo{number}{6} (\bibinfo{year}{2019}), \bibinfo{pages}{1--14}.
\newblock


\bibitem[Starke et~al\mbox{.}(2020)]%
        {starke2020local}
\bibfield{author}{\bibinfo{person}{Sebastian Starke}, \bibinfo{person}{Yiwei Zhao}, \bibinfo{person}{Taku Komura}, {and} \bibinfo{person}{Kazi Zaman}.} \bibinfo{year}{2020}\natexlab{}.
\newblock \showarticletitle{Local motion phases for learning multi-contact character movements}.
\newblock \bibinfo{journal}{\emph{ACM Transactions on Graphics (TOG)}} \bibinfo{volume}{39}, \bibinfo{number}{4} (\bibinfo{year}{2020}), \bibinfo{pages}{54--1}.
\newblock


\bibitem[Starke et~al\mbox{.}(2021)]%
        {starke2021neural}
\bibfield{author}{\bibinfo{person}{Sebastian Starke}, \bibinfo{person}{Yiwei Zhao}, \bibinfo{person}{Fabio Zinno}, {and} \bibinfo{person}{Taku Komura}.} \bibinfo{year}{2021}\natexlab{}.
\newblock \showarticletitle{Neural animation layering for synthesizing martial arts movements}.
\newblock \bibinfo{journal}{\emph{ACM Transactions on Graphics (TOG)}} \bibinfo{volume}{40}, \bibinfo{number}{4} (\bibinfo{year}{2021}), \bibinfo{pages}{1--16}.
\newblock


\bibitem[Sun et~al\mbox{.}(2024)]%
        {sun2024autoregressive}
\bibfield{author}{\bibinfo{person}{Peize Sun}, \bibinfo{person}{Yi Jiang}, \bibinfo{person}{Shoufa Chen}, \bibinfo{person}{Shilong Zhang}, \bibinfo{person}{Bingyue Peng}, \bibinfo{person}{Ping Luo}, {and} \bibinfo{person}{Zehuan Yuan}.} \bibinfo{year}{2024}\natexlab{}.
\newblock \showarticletitle{Autoregressive model beats diffusion: Llama for scalable image generation}.
\newblock \bibinfo{journal}{\emph{arXiv preprint arXiv:2406.06525}} (\bibinfo{year}{2024}).
\newblock


\bibitem[Team(2025)]%
        {hymotion2025}
\bibfield{author}{\bibinfo{person}{Tencent Hunyuan 3D Digital~Human Team}.} \bibinfo{year}{2025}\natexlab{}.
\newblock \showarticletitle{HY-Motion 1.0: Scaling Flow Matching Models for Text-To-Motion Generation}.
\newblock \bibinfo{journal}{\emph{arXiv preprint arXiv:2512.23464}} (\bibinfo{year}{2025}).
\newblock


\bibitem[Tevet et~al\mbox{.}(2025)]%
        {tevetclosd}
\bibfield{author}{\bibinfo{person}{Guy Tevet}, \bibinfo{person}{Sigal Raab}, \bibinfo{person}{Setareh Cohan}, \bibinfo{person}{Daniele Reda}, \bibinfo{person}{Zhengyi Luo}, \bibinfo{person}{Xue~Bin Peng}, \bibinfo{person}{Amit~Haim Bermano}, {and} \bibinfo{person}{Michiel van~de Panne}.} \bibinfo{year}{2025}\natexlab{}.
\newblock \showarticletitle{CLoSD: Closing the Loop between Simulation and Diffusion for multi-task character control}. In \bibinfo{booktitle}{\emph{The Thirteenth International Conference on Learning Representations}}.
\newblock


\bibitem[Tevet et~al\mbox{.}(2022)]%
        {tevet2022human}
\bibfield{author}{\bibinfo{person}{Guy Tevet}, \bibinfo{person}{Sigal Raab}, \bibinfo{person}{Brian Gordon}, \bibinfo{person}{Yonatan Shafir}, \bibinfo{person}{Daniel Cohen-Or}, {and} \bibinfo{person}{Amit~H Bermano}.} \bibinfo{year}{2022}\natexlab{}.
\newblock \showarticletitle{Human motion diffusion model}.
\newblock \bibinfo{journal}{\emph{arXiv preprint arXiv:2209.14916}} (\bibinfo{year}{2022}).
\newblock


\bibitem[Tian et~al\mbox{.}(2024)]%
        {tian2024visual}
\bibfield{author}{\bibinfo{person}{Keyu Tian}, \bibinfo{person}{Yi Jiang}, \bibinfo{person}{Zehuan Yuan}, \bibinfo{person}{Bingyue Peng}, {and} \bibinfo{person}{Liwei Wang}.} \bibinfo{year}{2024}\natexlab{}.
\newblock \showarticletitle{Visual autoregressive modeling: Scalable image generation via next-scale prediction}.
\newblock \bibinfo{journal}{\emph{Advances in neural information processing systems}}  \bibinfo{volume}{37} (\bibinfo{year}{2024}), \bibinfo{pages}{84839--84865}.
\newblock


\bibitem[Unitree(2025)]%
        {unitree_boxing}
\bibfield{author}{\bibinfo{person}{Unitree}.} \bibinfo{year}{2025}\natexlab{}.
\newblock \bibinfo{title}{Unitree Boxing}.
\newblock \bibinfo{howpublished}{\url{https://www.unitree.com/boxing}}.
\newblock
\newblock
\shownote{Accessed: 2024-06-30}.


\bibitem[Unity(2025)]%
        {blendtrees}
\bibfield{author}{\bibinfo{person}{Unity}.} \bibinfo{year}{2025}\natexlab{}.
\newblock \bibinfo{title}{Unity User Manual (6.2)}.
\newblock \bibinfo{howpublished}{\url{https://docs.unity3d.com/Manual/class-BlendTree.html}}.
\newblock


\bibitem[Van Den~Oord et~al\mbox{.}(2017)]%
        {van2017neural}
\bibfield{author}{\bibinfo{person}{Aaron Van Den~Oord}, \bibinfo{person}{Oriol Vinyals}, {et~al\mbox{.}}} \bibinfo{year}{2017}\natexlab{}.
\newblock \showarticletitle{Neural discrete representation learning}.
\newblock \bibinfo{journal}{\emph{Advances in neural information processing systems}}  \bibinfo{volume}{30} (\bibinfo{year}{2017}).
\newblock


\bibitem[Vaswani et~al\mbox{.}(2017)]%
        {vaswani2017attention}
\bibfield{author}{\bibinfo{person}{Ashish Vaswani}, \bibinfo{person}{Noam Shazeer}, \bibinfo{person}{Niki Parmar}, \bibinfo{person}{Jakob Uszkoreit}, \bibinfo{person}{Llion Jones}, \bibinfo{person}{Aidan~N Gomez}, \bibinfo{person}{{\L}ukasz Kaiser}, {and} \bibinfo{person}{Illia Polosukhin}.} \bibinfo{year}{2017}\natexlab{}.
\newblock \showarticletitle{Attention is all you need}.
\newblock \bibinfo{journal}{\emph{Advances in neural information processing systems}}  \bibinfo{volume}{30} (\bibinfo{year}{2017}).
\newblock


\bibitem[Wang et~al\mbox{.}(2020)]%
        {wang2020unicon}
\bibfield{author}{\bibinfo{person}{Tingwu Wang}, \bibinfo{person}{Yunrong Guo}, \bibinfo{person}{Maria Shugrina}, {and} \bibinfo{person}{Sanja Fidler}.} \bibinfo{year}{2020}\natexlab{}.
\newblock \showarticletitle{UniCon: Universal Neural Controller For Physics-based Character Motion}.
\newblock \bibinfo{journal}{\emph{arXiv preprint arXiv:2011.15119}} (\bibinfo{year}{2020}).
\newblock


\bibitem[Wen et~al\mbox{.}(2025)]%
        {wen2025efficient}
\bibfield{author}{\bibinfo{person}{Boran Wen}, \bibinfo{person}{Ye Lu}, \bibinfo{person}{Keyan Wan}, \bibinfo{person}{Sirui Wang}, \bibinfo{person}{Jiahong Zhou}, \bibinfo{person}{Junxuan Liang}, \bibinfo{person}{Xinpeng Liu}, \bibinfo{person}{Bang Xiao}, \bibinfo{person}{Dingbang Huang}, \bibinfo{person}{Ruiyang Liu}, {et~al\mbox{.}}} \bibinfo{year}{2025}\natexlab{}.
\newblock \showarticletitle{Efficient and Scalable Monocular Human-Object Interaction Motion Reconstruction}.
\newblock \bibinfo{journal}{\emph{arXiv preprint arXiv:2512.00960}} (\bibinfo{year}{2025}).
\newblock


\bibitem[Yang et~al\mbox{.}(2025)]%
        {yang2025omniretarget}
\bibfield{author}{\bibinfo{person}{Lujie Yang}, \bibinfo{person}{Xiaoyu Huang}, \bibinfo{person}{Zhen Wu}, \bibinfo{person}{Angjoo Kanazawa}, \bibinfo{person}{Pieter Abbeel}, \bibinfo{person}{Carmelo Sferrazza}, \bibinfo{person}{C~Karen Liu}, \bibinfo{person}{Rocky Duan}, {and} \bibinfo{person}{Guanya Shi}.} \bibinfo{year}{2025}\natexlab{}.
\newblock \showarticletitle{Omniretarget: Interaction-preserving data generation for humanoid whole-body loco-manipulation and scene interaction}.
\newblock \bibinfo{journal}{\emph{arXiv preprint arXiv:2509.26633}} (\bibinfo{year}{2025}).
\newblock


\bibitem[Yi and Jee(2019)]%
        {yi2019search}
\bibfield{author}{\bibinfo{person}{Gwonjin Yi} {and} \bibinfo{person}{Junghoon Jee}.} \bibinfo{year}{2019}\natexlab{}.
\newblock \showarticletitle{Search Space Reduction In Motion Matching by Trajectory Clustering}.
\newblock In \bibinfo{booktitle}{\emph{SIGGRAPH Asia 2019 Posters}}. \bibinfo{pages}{1--2}.
\newblock


\bibitem[Yin et~al\mbox{.}(2025)]%
        {yin2025unitrackerlearninguniversalwholebody}
\bibfield{author}{\bibinfo{person}{Kangning Yin}, \bibinfo{person}{Weishuai Zeng}, \bibinfo{person}{Ke Fan}, \bibinfo{person}{Minyue Dai}, \bibinfo{person}{Zirui Wang}, \bibinfo{person}{Qiang Zhang}, \bibinfo{person}{Zheng Tian}, \bibinfo{person}{Jingbo Wang}, \bibinfo{person}{Jiangmiao Pang}, {and} \bibinfo{person}{Weinan Zhang}.} \bibinfo{year}{2025}\natexlab{}.
\newblock \bibinfo{title}{UniTracker: Learning Universal Whole-Body Motion Tracker for Humanoid Robots}.
\newblock
\showeprint[arxiv]{2507.07356}~[cs.RO]
\urldef\tempurl%
\url{https://arxiv.org/abs/2507.07356}
\showURL{%
\tempurl}


\bibitem[Yu et~al\mbox{.}(2023a)]%
        {yu2023magvit}
\bibfield{author}{\bibinfo{person}{Lijun Yu}, \bibinfo{person}{Yong Cheng}, \bibinfo{person}{Kihyuk Sohn}, \bibinfo{person}{Jos{\'e} Lezama}, \bibinfo{person}{Han Zhang}, \bibinfo{person}{Huiwen Chang}, \bibinfo{person}{Alexander~G Hauptmann}, \bibinfo{person}{Ming-Hsuan Yang}, \bibinfo{person}{Yuan Hao}, \bibinfo{person}{Irfan Essa}, {et~al\mbox{.}}} \bibinfo{year}{2023}\natexlab{a}.
\newblock \showarticletitle{Magvit: Masked generative video transformer}. In \bibinfo{booktitle}{\emph{Proceedings of the IEEE/CVF Conference on Computer Vision and Pattern Recognition}}. \bibinfo{pages}{10459--10469}.
\newblock


\bibitem[Yu et~al\mbox{.}(2023b)]%
        {yu2023language}
\bibfield{author}{\bibinfo{person}{Lijun Yu}, \bibinfo{person}{Jos{\'e} Lezama}, \bibinfo{person}{Nitesh~B Gundavarapu}, \bibinfo{person}{Luca Versari}, \bibinfo{person}{Kihyuk Sohn}, \bibinfo{person}{David Minnen}, \bibinfo{person}{Yong Cheng}, \bibinfo{person}{Vighnesh Birodkar}, \bibinfo{person}{Agrim Gupta}, \bibinfo{person}{Xiuye Gu}, {et~al\mbox{.}}} \bibinfo{year}{2023}\natexlab{b}.
\newblock \showarticletitle{Language Model Beats Diffusion--Tokenizer is Key to Visual Generation}.
\newblock \bibinfo{journal}{\emph{arXiv preprint arXiv:2310.05737}} (\bibinfo{year}{2023}).
\newblock


\bibitem[Ze et~al\mbox{.}(2025)]%
        {ze2025gmr}
\bibfield{author}{\bibinfo{person}{Yanjie Ze}, \bibinfo{person}{João~Pedro Araújo}, \bibinfo{person}{Jiajun Wu}, {and} \bibinfo{person}{C.~Karen Liu}.} \bibinfo{year}{2025}\natexlab{}.
\newblock \bibinfo{booktitle}{\emph{GMR: General Motion Retargeting}}.
\newblock
\urldef\tempurl%
\url{https://github.com/YanjieZe/GMR}
\showURL{%
\tempurl}
\newblock
\shownote{GitHub repository}.


\bibitem[Zeng et~al\mbox{.}(2025)]%
        {zeng2025behavior}
\bibfield{author}{\bibinfo{person}{Weishuai Zeng}, \bibinfo{person}{Shunlin Lu}, \bibinfo{person}{Kangning Yin}, \bibinfo{person}{Xiaojie Niu}, \bibinfo{person}{Minyue Dai}, \bibinfo{person}{Jingbo Wang}, {and} \bibinfo{person}{Jiangmiao Pang}.} \bibinfo{year}{2025}\natexlab{}.
\newblock \showarticletitle{Behavior Foundation Model for Humanoid Robots}.
\newblock \bibinfo{journal}{\emph{arXiv preprint arXiv:2509.13780}} (\bibinfo{year}{2025}).
\newblock


\bibitem[Zhang et~al\mbox{.}(2018)]%
        {zhang2018mode}
\bibfield{author}{\bibinfo{person}{He Zhang}, \bibinfo{person}{Sebastian Starke}, \bibinfo{person}{Taku Komura}, {and} \bibinfo{person}{Jun Saito}.} \bibinfo{year}{2018}\natexlab{}.
\newblock \showarticletitle{Mode-adaptive neural networks for quadruped motion control}.
\newblock \bibinfo{journal}{\emph{ACM Transactions on Graphics (TOG)}} \bibinfo{volume}{37}, \bibinfo{number}{4} (\bibinfo{year}{2018}), \bibinfo{pages}{1--11}.
\newblock


\bibitem[Zhang et~al\mbox{.}(2024a)]%
        {zhang2024motiondiffuse}
\bibfield{author}{\bibinfo{person}{Mingyuan Zhang}, \bibinfo{person}{Zhongang Cai}, \bibinfo{person}{Liang Pan}, \bibinfo{person}{Fangzhou Hong}, \bibinfo{person}{Xinying Guo}, \bibinfo{person}{Lei Yang}, {and} \bibinfo{person}{Ziwei Liu}.} \bibinfo{year}{2024}\natexlab{a}.
\newblock \showarticletitle{Motiondiffuse: Text-driven human motion generation with diffusion model}.
\newblock \bibinfo{journal}{\emph{IEEE transactions on pattern analysis and machine intelligence}} \bibinfo{volume}{46}, \bibinfo{number}{6} (\bibinfo{year}{2024}), \bibinfo{pages}{4115--4128}.
\newblock


\bibitem[Zhang et~al\mbox{.}(2024b)]%
        {zhang2024motiongpt}
\bibfield{author}{\bibinfo{person}{Yaqi Zhang}, \bibinfo{person}{Di Huang}, \bibinfo{person}{Bin Liu}, \bibinfo{person}{Shixiang Tang}, \bibinfo{person}{Yan Lu}, \bibinfo{person}{Lu Chen}, \bibinfo{person}{Lei Bai}, \bibinfo{person}{Qi Chu}, \bibinfo{person}{Nenghai Yu}, {and} \bibinfo{person}{Wanli Ouyang}.} \bibinfo{year}{2024}\natexlab{b}.
\newblock \showarticletitle{Motiongpt: Finetuned llms are general-purpose motion generators}. In \bibinfo{booktitle}{\emph{Proceedings of the AAAI Conference on Artificial Intelligence}}, Vol.~\bibinfo{volume}{38}. \bibinfo{pages}{7368--7376}.
\newblock


\bibitem[Zhang et~al\mbox{.}(2025)]%
        {zhang2025track}
\bibfield{author}{\bibinfo{person}{Zhikai Zhang}, \bibinfo{person}{Jun Guo}, \bibinfo{person}{Chao Chen}, \bibinfo{person}{Jilong Wang}, \bibinfo{person}{Chenghuai Lin}, \bibinfo{person}{Yunrui Lian}, \bibinfo{person}{Han Xue}, \bibinfo{person}{Zhenrong Wang}, \bibinfo{person}{Maoqi Liu}, \bibinfo{person}{Huaping Liu}, {et~al\mbox{.}}} \bibinfo{year}{2025}\natexlab{}.
\newblock \showarticletitle{Track Any Motions under Any Disturbances}.
\newblock \bibinfo{journal}{\emph{arXiv preprint arXiv:2509.13833}} (\bibinfo{year}{2025}).
\newblock


\bibitem[Zhou et~al\mbox{.}(2024)]%
        {zhou2024emdm}
\bibfield{author}{\bibinfo{person}{Wenyang Zhou}, \bibinfo{person}{Zhiyang Dou}, \bibinfo{person}{Zeyu Cao}, \bibinfo{person}{Zhouyingcheng Liao}, \bibinfo{person}{Jingbo Wang}, \bibinfo{person}{Wenjia Wang}, \bibinfo{person}{Yuan Liu}, \bibinfo{person}{Taku Komura}, \bibinfo{person}{Wenping Wang}, {and} \bibinfo{person}{Lingjie Liu}.} \bibinfo{year}{2024}\natexlab{}.
\newblock \showarticletitle{Emdm: Efficient motion diffusion model for fast and high-quality motion generation}. In \bibinfo{booktitle}{\emph{European Conference on Computer Vision}}. Springer, \bibinfo{pages}{18--38}.
\newblock


\bibitem[Zhu et~al\mbox{.}(2026)]%
        {zhu2026hiking}
\bibfield{author}{\bibinfo{person}{Shaoting Zhu}, \bibinfo{person}{Ziwen Zhuang}, \bibinfo{person}{Mengjie Zhao}, \bibinfo{person}{Kun-Ying Lee}, {and} \bibinfo{person}{Hang Zhao}.} \bibinfo{year}{2026}\natexlab{}.
\newblock \showarticletitle{Hiking in the Wild: A Scalable Perceptive Parkour Framework for Humanoids}.
\newblock \bibinfo{journal}{\emph{arXiv preprint arXiv:2601.07718}} (\bibinfo{year}{2026}).
\newblock


\end{thebibliography}
\clearpage
\appendix
\section{Scaling Behavior with Number of GPUs}\label{section:appendix_gpu_scaling}

\begin{figure*}[!t]
    \centering
    \includegraphics[width=1.00\textwidth]{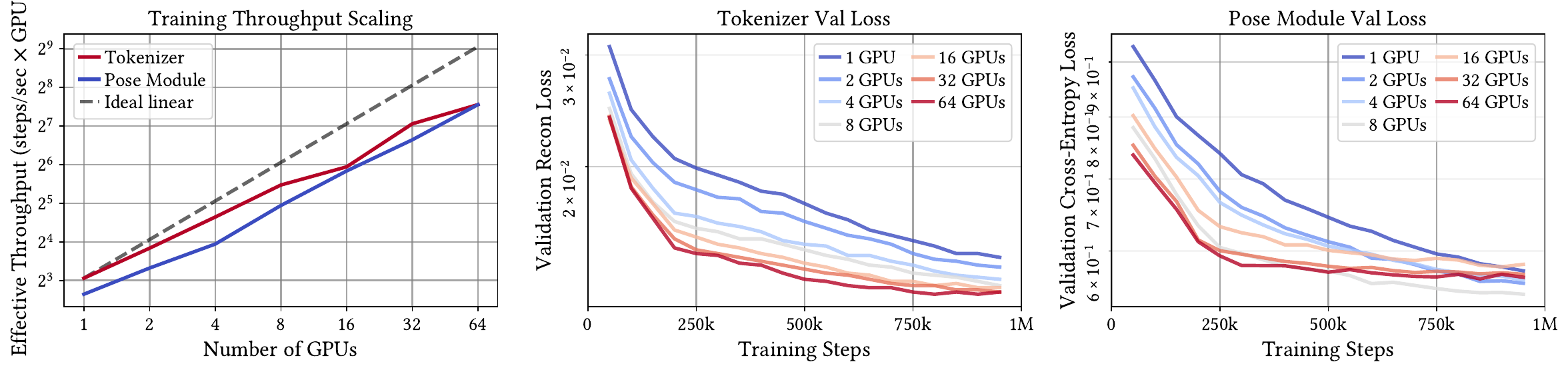}
    \caption{Ablation study on GPU scaling during training.
    Left: ideal vs.\ achieved throughput scaling with the number of GPUs.
    Middle: tokenizer reconstruction loss on the validation set.
    Right: cross-entropy loss for token prediction on the validation set.}
    \label{fig:ablation_gpu_scaling_metrics}
\end{figure*}

Our main experiments utilize a large number of GPUs, which may not be accessible to individual researchers or smaller studios.
To evaluate the impact of compute budget, we train with 1, 2, 4, 8, 16, 32, and 64 GPUs and report results in Figure~\ref{fig:ablation_gpu_scaling_metrics}.
As shown in the left plot, throughput scales near-linearly with GPU count, directly reducing training time.
In general, more GPUs lead to faster convergence and modestly better asymptotic performance, though occasional exceptions arise from random seed variations or distributional differences in the generative model training stage.
The performance gap primarily manifests in tokenizer quality; however, these numerical differences translate to only minor perceptual changes in the generated motion.
We conclude that training with fewer GPUs remains viable without significant degradation in visual quality.

\section{Comparison between FSQ and VQ-VAE}\label{section:appendix_fsq_vqvae}
Our tokenizer and pose module are agnostic to the choice of internal tokenization strategy.
For instance, FSQ~\cite{mentzer2023finite} has been adopted in recent work such as SONIC~\cite{luo2025sonic} and is also applicable to our framework.
To draw a parallel between FSQ and VQ-VAE in our framework, we treat the latent dimensions in FSQ as the number of heads, and the levels per dimension as the codebook size per head.
Similarly, for the pose module, we ask it to predict the corresponding token indices from the FSQ tokenizer.

\begin{figure*}[!t]
    \centering
    \includegraphics[width=1.00\textwidth]{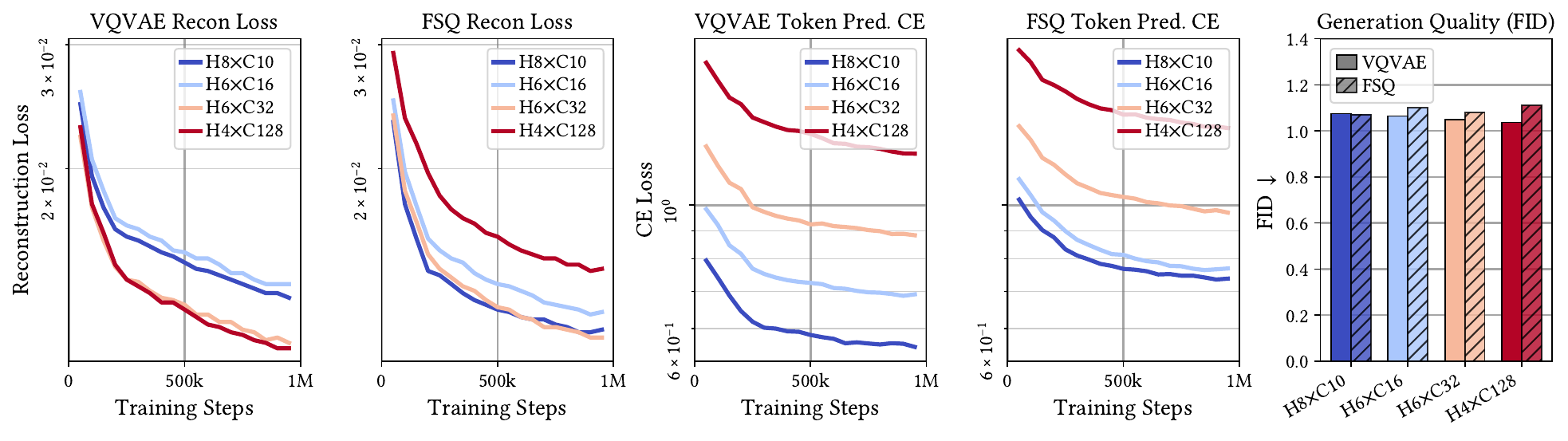}
    \caption{Comparison between FSQ and VQ-VAE with matched token capacity.
    Left two plots: tokenizer reconstruction loss on the validation set during training.
    Middle two plots: cross-entropy loss for token prediction on the validation set.
    Right: FID comparison on the test set.
    ``H'' denotes the number of heads; ``C'' denotes the codebook size per head.
}\label{fig:ablation_fsq_vqvae_metrics}
\end{figure*}

As shown in Figure~\ref{fig:ablation_fsq_vqvae_metrics}, FSQ and VQ-VAE yield similar performance in terms of FID, reconstruction loss, and cross-entropy loss given the same number of tokens and heads.
However, VQ-VAE consistently achieves lower cross-entropy loss and better FID than FSQ, suggesting that VQ-VAE produces a latent space that is slightly easier for the generative model to model.
We observe similar improvement in visual quality and robotics deployment. Based on these results, we use VQ-VAE as the default tokenizer for \ourmethod{}.

\begin{figure}[!t]
    \centering
    \includegraphics[width=0.45\textwidth]{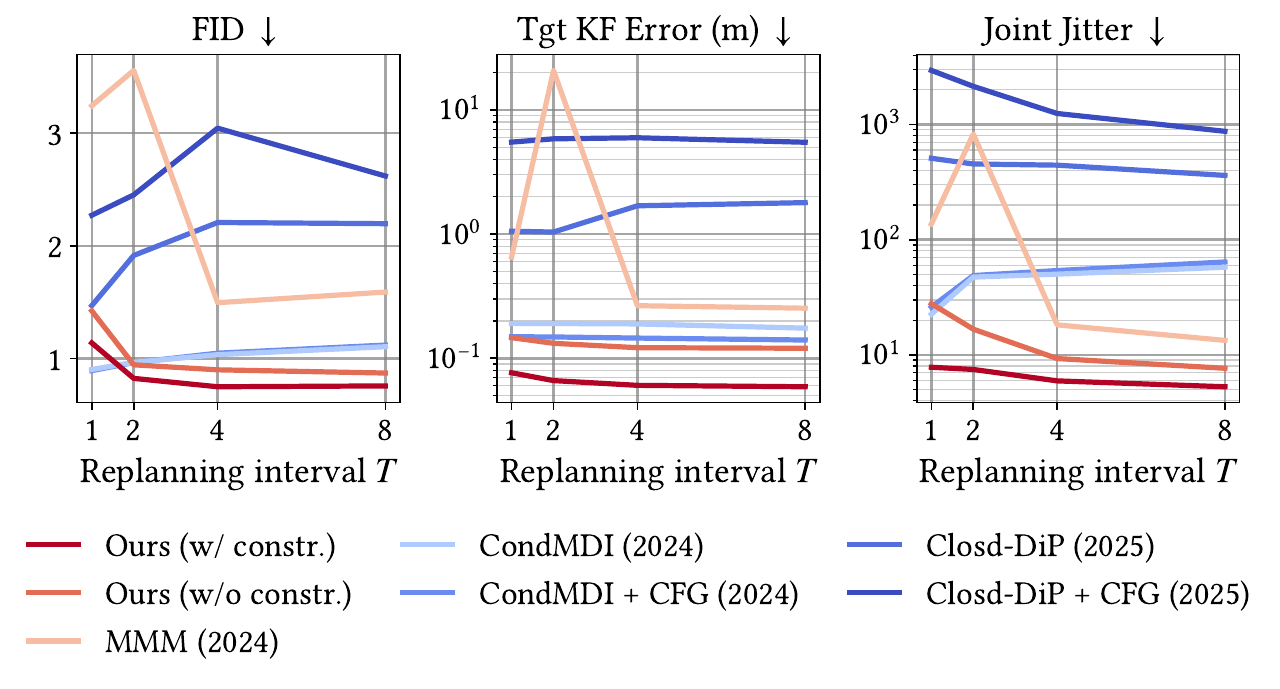}
    \caption{
    Evaluation metrics under different replanning frequencies.
    Left: FID comparison between different replanning frequencies.
    Middle: Keyframe joint position error comparison between different replanning frequencies.
    Right: Joint jittering comparison between different replanning frequencies.
}\label{fig:ablation_replanning_frequency_metrics}
\end{figure}
\section{Replanning Frequency}\label{section:replanning_frequency}
In \ourmethod{}, replanning is governed by the drop-frame attribute $\tau$ introduced in Section~\ref{section:smart_object}: a replan is triggered when fewer than $\tau$ frames remain in the motion buffer.
During deployment, replanning also occurs whenever control commands change to allow for instant control reaction, as described in Section~\ref{section:exp_interactive_animation_and_robotics_applications_and_engineering_details}.

To quantify the impact of replanning frequency, we evaluate \ourmethod{} with replanning intervals ranging from every 1 frame to every 8 frames.
As shown in Figure~\ref{fig:ablation_replanning_frequency_metrics},
discrete latent methods such as \ourmethod{} and MMM exhibit slightly worse FID, keyframe error, and joint jitter at higher replanning frequencies, whereas continuous diffusion methods show improved FID and joint jitter with more frequent replanning.
We hypothesize that discrete latent representations struggle to capture subtle inter-step variations, and overly frequent replanning traps generation in its early phase, delaying the onset of the desired style and degrading fine details.
We observe consistent trends in our visual demos and refer readers to the supplementary videos for further illustration.
Based on these findings, we recommend a default replanning interval of 3 to 9 frames, combined with instant replanning upon command changes, as a practical engineering choice.

\begin{figure*}[!t]
    \centering
    \includegraphics[width=1.0\textwidth]{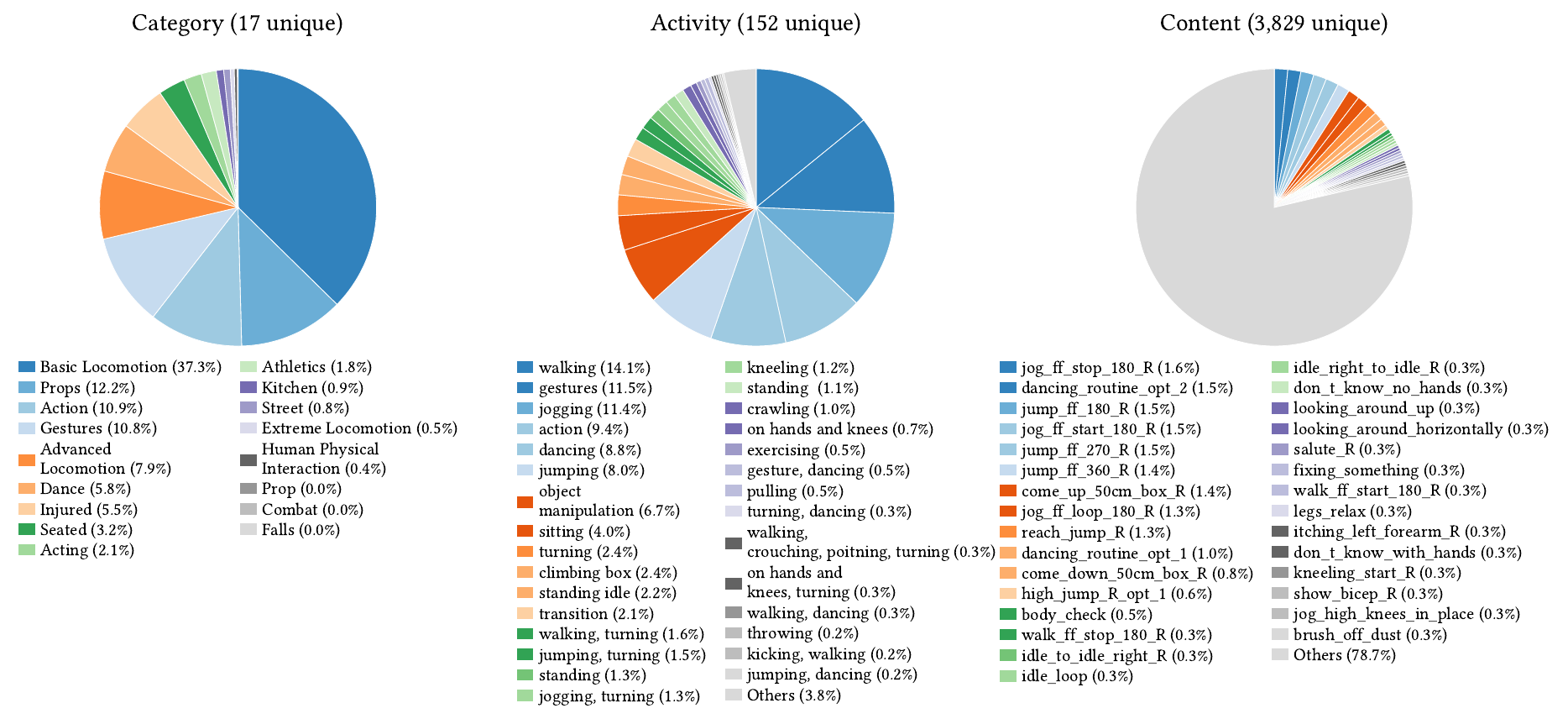}
    \vspace{-19pt}
    \caption{Overall statistics and diversity of the 140k BONES-SEED open-source subset dataset by categories, activities and contents.
    }\label{fig:appendix_dataset_diversity}
\end{figure*}

\section{2D Blendspace and Keyframe Blending in \ourmethod{}}\label{section:appendix_2d_blendspace}
A 2D blendspace is a standard technique in game engines such as Unreal Engine~\cite{epicgames2025blendspaces} that blends between pre-recorded animation clips based on two continuous input axes, typically speed and direction.
The neighboring clips are interpolated at runtime to produce smooth transitions for the given input parameters of speed and direction.
While effective, 2D blendspaces rely heavily on heuristics and operate at the clip level, requiring careful authoring of sample coverage to avoid blending artifacts.

In contrast, \ourmethod{} performs keyframe-level blending: rather than interpolating between entire clips, we directly interpolate individual keyframes in joint space from reference poses and feed them to the neural backbone.
Artifacts such as foot sliding or floating that would typically arise from naive interpolation are handled implicitly by \ourmethod{}.
Direct keyframe interpolation in \ourmethod{} still produces natural, artifact-free motions for two reasons: our model is robust to imperfect keyframes, and the drop-frame replanning mechanism ensures that the blended keyframes are never rigidly enforced during generation.
This frees us from the constraints of traditional 2D blendspaces, offering finer-grained control without the authoring overhead of populating a full blendspace grid.
We refer readers to the supplementary videos for a detailed demonstration.
\begin{figure*}[!h]
    \centering
    \includegraphics[width=1.0\textwidth]{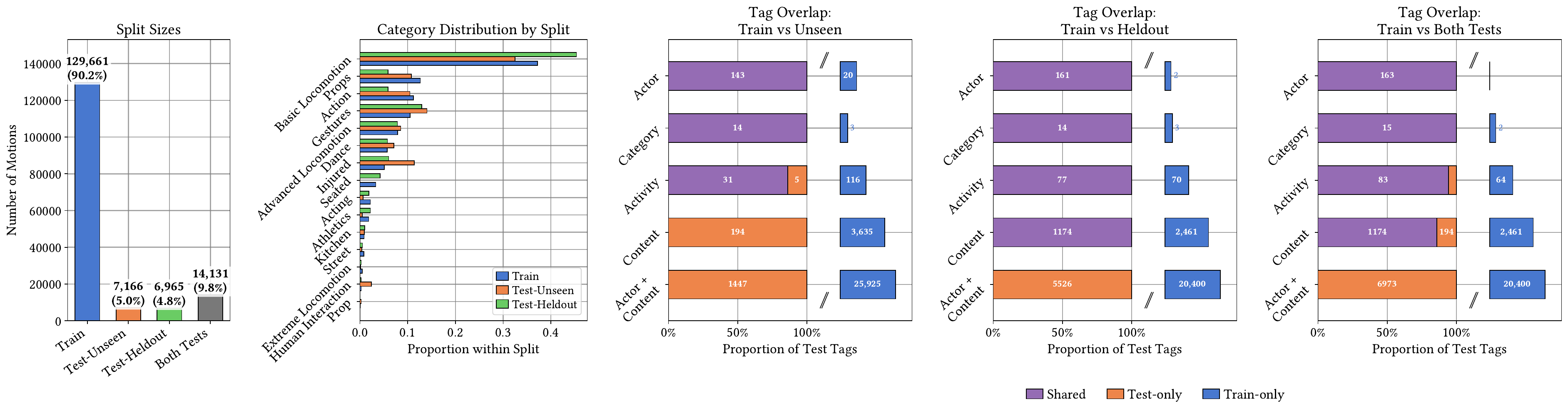}
    \vspace{-23pt}
    \caption{Overview of the two test sets split from the 140k BONES-SEED open-source subset dataset.
    }\label{fig:appendix_dataset_split_overview}
\end{figure*}

\begin{figure*}[!t]
    \centering
    \includegraphics[width=0.333\textwidth]{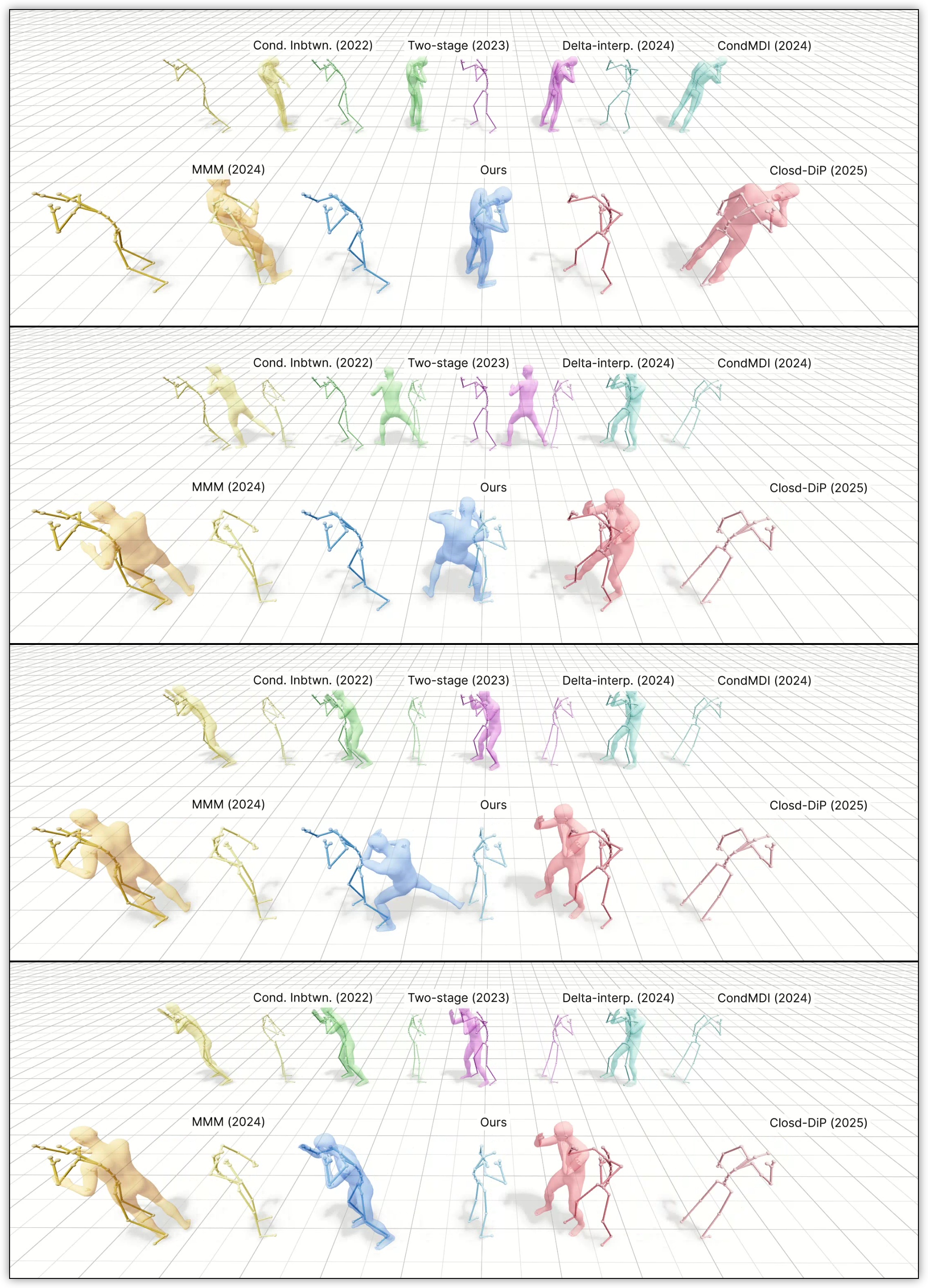}%
    \hfill
    \includegraphics[width=0.333\textwidth]{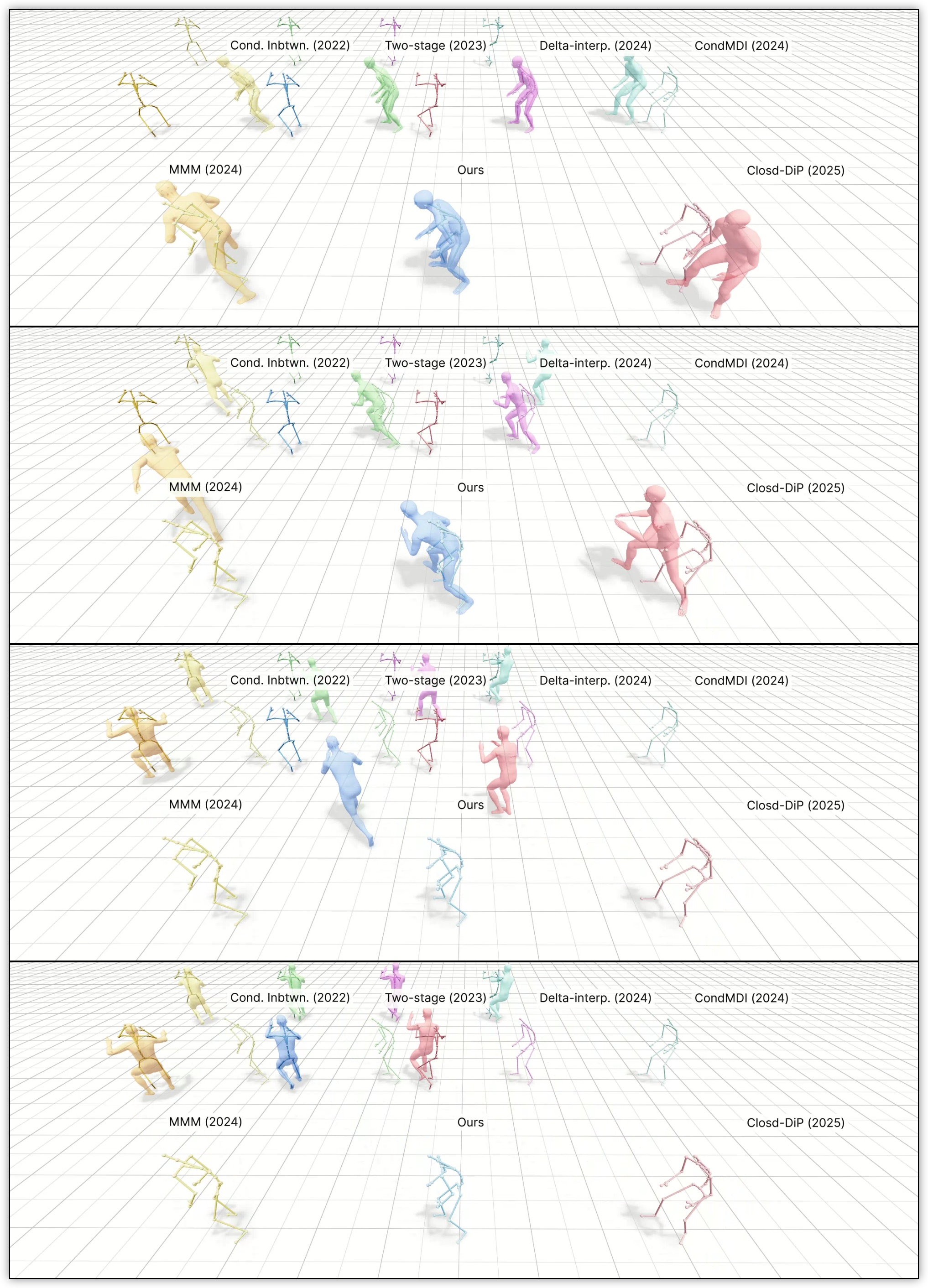}%
    \hfill
    \includegraphics[width=0.333\textwidth]{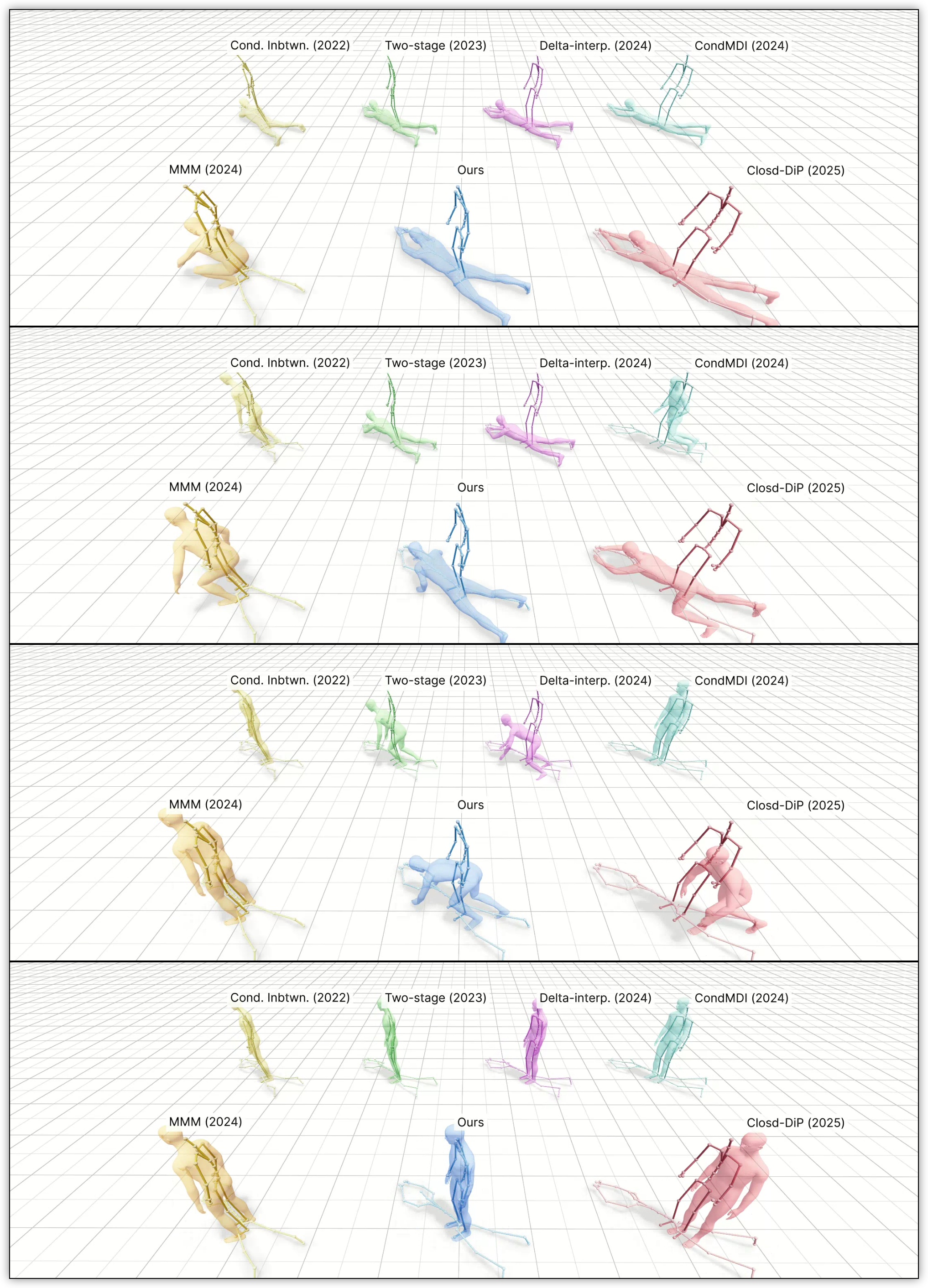}
    \caption{Additional in-betweening visual comparisons across three test cases. Each column shows a different test case; time progresses from top to bottom. Algorithm names are labeled in each subfigure. \ourmethod{} is shown as the blue character.
    }\label{fig:appendix_inbetween_comparisons}
\end{figure*}

\section{Dataset Statistics}\label{section:appendix_dataset_statistics}
Our full dataset contains 350k motion clips spanning 36 categories, 203 unique activities, 9,285 unique content types, and 163 unique performers.
Categories, activities, and content types represent progressively finer levels of motion categorization, as illustrated in Figure~\ref{fig:appendix_dataset_diversity}.

We also release an open-source subset of approximately \textbf{140k motion clips}, publicly available as \textbf{BONES-SEED}~\cite{bones_seed} via Bones Studio, whose statistics are shown in Figure~\ref{fig:appendix_dataset_split_overview}.
This subset retains roughly 40\% of motions, about half of the categories (17/36), 75\% of activities (152/203), 41\% of content types (3,829/9,285), and all 163 performers.
We split this subset (143,792 motions) into three partitions: a training set (129,661 motions, 90.2\%), a \emph{Test-Unseen} set (7,166 motions, 5.0\%), and a \emph{Test-Heldout} set (6,965 motions, 4.8\%).
The two test sets are designed to evaluate different aspects of generalization.
The Test-Unseen set has 0\% content-type overlap with the training set: all 194 content types are novel, and 5 out of 36 activities are unseen during training.
The Test-Heldout set shares 100\% overlap at the category, activity, content, and actor levels, but contains novel actor-content combinations.

As shown in Figure~\ref{fig:appendix_dataset_diversity}, locomotion is by far the most represented category, with diverse variations in speed, direction, and style that enable robust and natural locomotion generation.

The least covered categories are object manipulation and terrain interaction.
For manipulation, although many motions have been captured, the space of possible manipulation tasks is vast, and the dataset currently lacks finger and object motion data.
Additionally, no samples include continuous terrain information, limiting the model's ability to generate realistic motions on uneven surfaces.
While the dataset does contain motions with height variations, coverage remains sparse.
In the most extreme case for example, vaulting over a 1m obstacle has only one or two captured clips, posing severe challenges for learning and generalization.

\begin{figure*}[!t]
    \centering
    \includegraphics[width=0.99\textwidth]{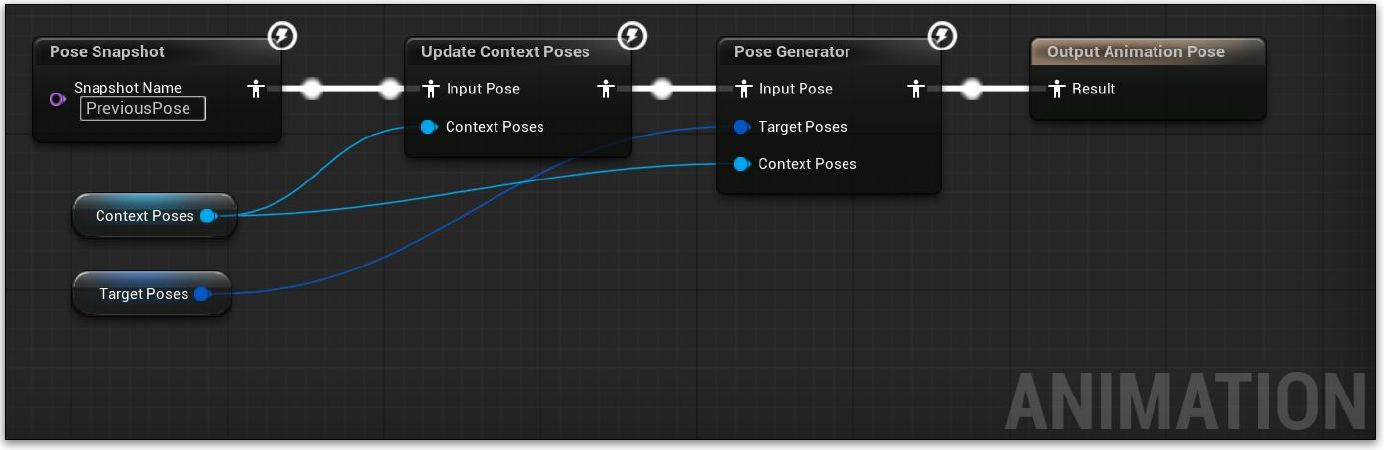}
    \caption{The complete animation graph used in our UE5 demo.
    Context poses are updated each frame and passed alongside target keyframes to the pose generator node, which invokes the {\ourmethod} backbone only when new targets are received.
    }\label{fig:appendix_ue5_anim_graph}
\end{figure*}
\section{In-betweening Visual Comparisons}\label{section:appendix_inbetweening_visual_comparisons}
We provide additional isolated in-betweening visual comparisons in Figure~\ref{fig:appendix_inbetween_comparisons}.
The skeletons denote keyframe constraints, and the characters denote snapshots of generations from each algorithm.
From left to right, the three test cases are: (1) transitioning from yawning to a boxing pose, (2) transitioning from a stealth pose to running, and (3) transitioning from lying on the ground to a standing pose.
\ourmethod{} consistently outperforms all baselines, which exhibit various artifacts such as implausible poses and poor keyframe adherence.
We refer readers to the demo videos for further illustration.

\section{UE5 Demo Implementation}\label{section:appendix_ue5_demo}

We describe the implementation of our UE5 demo, which showcases the full {\ourmethod} pipeline in a real-time interactive setting.

\begin{figure}[!tp]
    \centering
    \includegraphics[width=0.48\textwidth]{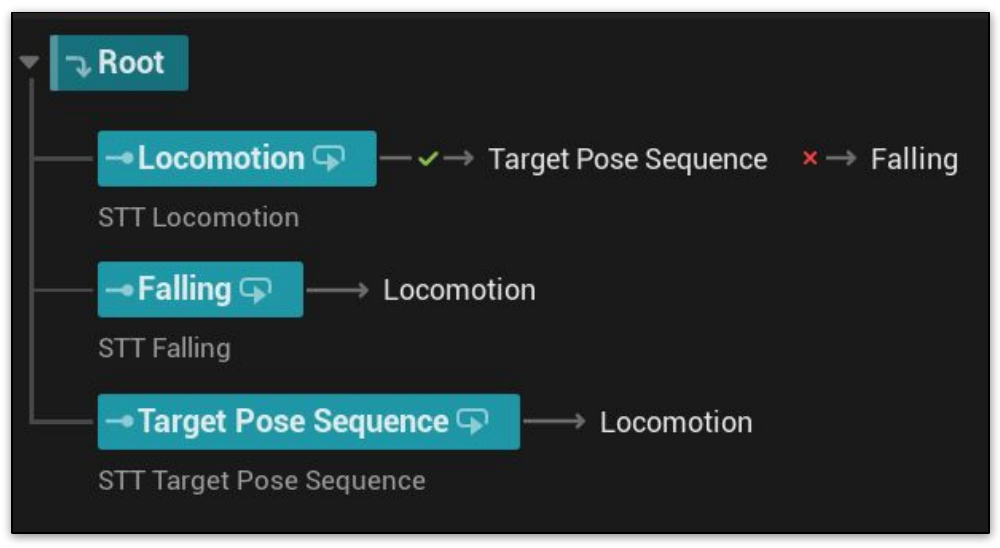}
    \caption{The UE5 StateTree governing character behavior, managing transitions between locomotion, target pose sequence (object interaction), and falling states.
    }\label{fig:appendix_ue5_gamestate}
\end{figure}

\begin{figure}[!tp]
    \centering
    \includegraphics[width=0.99\linewidth]{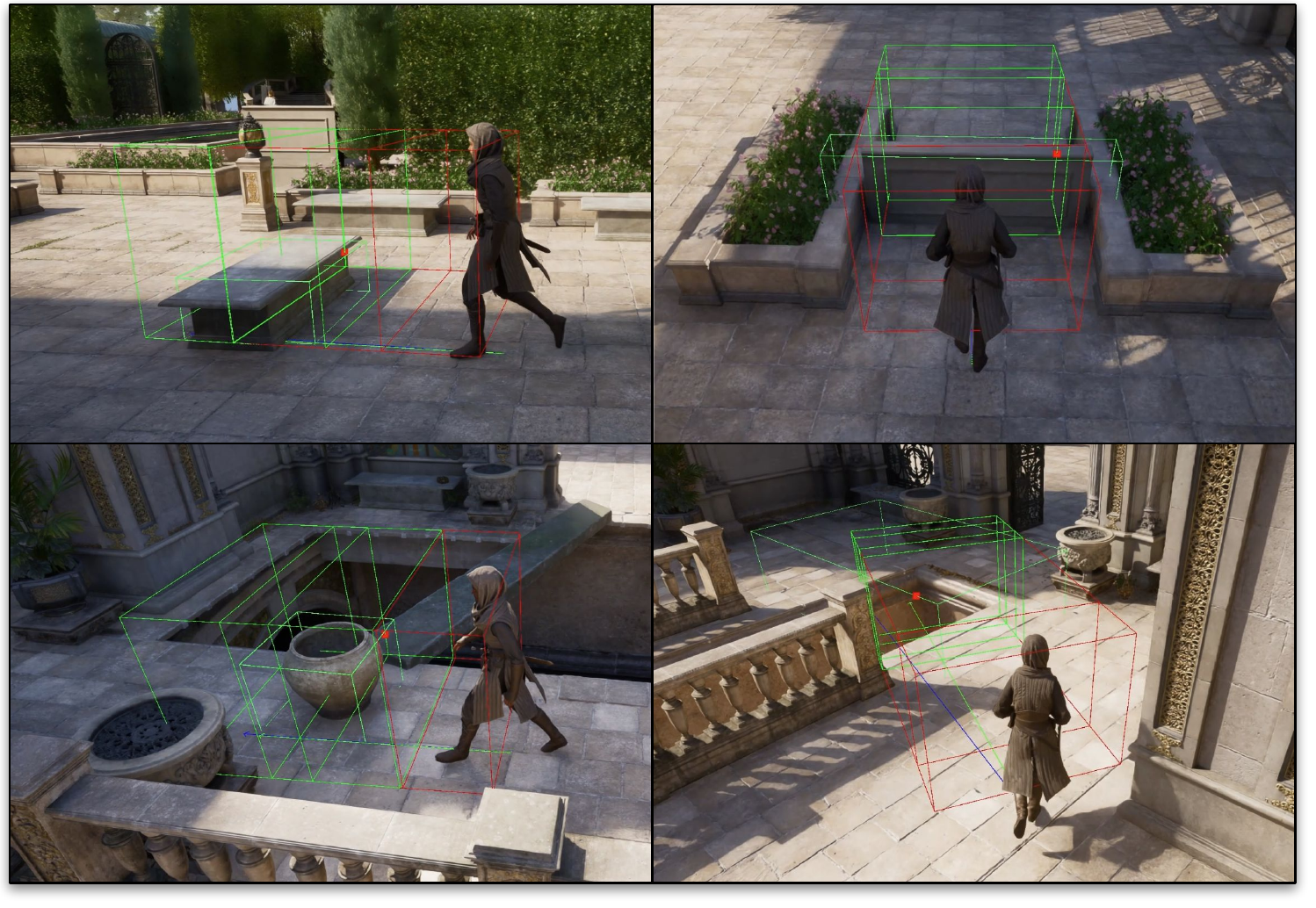}
    \caption{Runtime interaction detection via box traces in our UE5 demo.
    Green and red boxes indicate the detection and interaction trigger volumes, respectively.
    Examples shown include bench sitting, vaulting, object pickup, and platform jumping.
    }\label{fig:appendix_ue5_detection}
\end{figure}
\paragraph{Task Actors: A Unified Runtime Primitive.}
Both smart locomotion and smart object are realized through a shared custom actor blueprint type that we call a \emph{task actor}.
Each task actor encapsulates the runtime logic, assets, and state needed to produce target keyframes for the neural backbone.
At authoring time, a task actor samples and stores sparse target poses from user-provided clip assets, defining either the interaction keyframes for a scene object or the style poses for a parameterized locomotion blendspace.
At runtime, the active task actor provides the target keyframes that {\ourmethod} consumes to generate the next motion sequence.

Because task actors leverage UE5 blueprints, each instance can implement its own gameplay logic independently.
For example, a scene interaction task actor orients its keyframe sequence based on the character's approach angle, using the keyframe anchoring mechanism described in Section~\ref{section:smart_object}.
The key distinction between the two smart primitive types is ownership: the locomotion task actor resides on the character and is always active, while object interaction task actors are placed in the scene as standalone actors.
While target keyframes could, in principle, be supplied by any source, including learned models or procedural methods, for our demo we provide them from a handful of selected clips.

\paragraph{Interaction Detection and State Management.}
During locomotion, the system performs box traces to detect nearby interactable objects, as shown in Figure~\ref{fig:appendix_ue5_detection}.
When a smart object is activated, it becomes the new keyframe provider, interrupting the current motion and triggering generation from the object's target keyframes.
This on-demand interruption relies on the deterministic replanning behavior of {\ourmethod}: because the backbone produces consistent results regardless of when or how frequently replanning is triggered (see Appendix~\ref{section:replanning_frequency}), gameplay actions can seamlessly override the current motion without introducing artifacts.
This property is essential to the smart primitive framework, as it enables responsive, interruptible character behavior driven entirely by game events.

As shown in Figure~\ref{fig:appendix_ue5_gamestate}, the character's high-level behavior is governed by a UE5 StateTree with three states: \emph{locomotion}, where the locomotion task actor continuously supplies keyframes based on velocity, orientation, and replanning frequency; \emph{target pose sequence}, where an object interaction task actor provides keyframes to execute the interaction; and \emph{falling}, which dynamically spawns a task actor at runtime.
For falls, the system measures the current height and runs a projectile simulation to determine the landing position, placing the appropriate landing keyframe based on the fall height.
Upon completion, control returns to the locomotion state.

\paragraph{Animation Graph and Inference Scheduling.}
The animation graph for the entire demo is deliberately minimal, as shown in Figure~\ref{fig:appendix_ue5_anim_graph}.
Each frame, the context pose buffer is updated with the previous frame's pose, and both context and target poses are passed to a \emph{pose generator} node.
This node invokes the {\ourmethod} backbone only when new target keyframes are received; otherwise, it samples directly from the previously generated motion buffer, so model inference does not occur every frame.
Communication between gameplay and animation is bidirectional: gameplay can interrupt the current motion at any time by supplying new keyframes, while the pose generator notifies gameplay when the buffer is nearing its end and a new generation is needed.

\section{Contributors}
\textbf{Project Lead:} Tingwu Wang.\\
\textbf{Research:} Tingwu Wang, Olivier Dionne, Michael De Ruyter, David Minor, Davis Rempe, Kaifeng Zhao, Mathis Petrovich.\\
\textbf{Animation development:} Olivier Dionne, Tingwu Wang, Michael De Ruyter, David Minor, Brian Robison, Xavier Blackwell.\\
\textbf{Robotics development:} Tingwu Wang, Chenran Li, Zhengyi Luo, Ye Yuan.\\
\textbf{Game art, assets and level design:} Brian Robison, Xavier Blackwell, Bernardo Antoniazzi.\\
\textbf{Advising:} Simon Yuen, Xue Bin Peng, Yuke Zhu.\\
\end{document}